\documentclass[11pt]{article}

\PassOptionsToPackage{table}{xcolor}

\usepackage[preprint]{acl}

\usepackage{times}
\usepackage{latexsym}
\usepackage[T1]{fontenc}
\usepackage[utf8]{inputenc}
\usepackage{microtype}
\usepackage{graphicx}
\usepackage{booktabs}
\usepackage{amsmath}
\usepackage{float}

\definecolor{resultblue}{HTML}{EAF2FF}
\definecolor{resultgold}{HTML}{FFF4D6}
\definecolor{resultgreen}{HTML}{EAF7EE}

\title{Same Entity, Different Script: Diagnosing Cross-Script Entity Matching Failures
in Language Models}

\newcommand{\blfootnote}[1]{%
  \begingroup\renewcommand\thefootnote{}\footnote{#1}\addtocounter{footnote}{-1}\endgroup}

\author{
  Athira Gopal$^{*}$ \quad Patil Prathamesh Pravin$^{*}$ \quad Ankush Kadu \\
  \textbf{Arpit Jain} \quad \textbf{Ashwanth Krishnan} \\
  QpiAI India \\
  {\{athira.g, patil.p, ankush.k, arpit.j, ashwanth.krishnan\}@qpiai.tech}
}

\usepackage{multirow}
\usepackage{tabularx}
\usepackage{amssymb}
\usepackage{alphalph}
\usepackage{needspace}
\begin{document}
\maketitle

\blfootnote{$^{*}$\;Equal contribution.}
\begin{abstract}
Multilingual language models can give different identity judgments when the same
people's names are written in different scripts. We study this effect using 601
audited identity pairs with attested Devanagari and Latin forms, holding the
question and relation fixed. Qwen2.5-7B's same-person accuracy falls from 35.1\%
in Latin to 8.5\% in Devanagari under Marathi prompts, while different-person
accuracy remains above 98\%. Evaluation across five models reveals distinct
response biases, and reduced tokenization fragmentation does not reliably close
the gap. Replicating the design on 374 audited Malayalam pairs
built from the same entities reproduces the effect in a second script and
language family: same-person accuracy falls to 5.3\%, about half the matched
Devanagari rate, while the margin still ranks matching above non-matching names
at 0.925 AUROC. Latin-trained probes recover label information from Devanagari states,
exceeding a string-feature baseline under similarity-matched negatives. Scalar
calibration recovers 87\% of the raw model--probe balanced-accuracy gap,
identifying decision criterion placement as a major contributor. Activation
patching localises causal sensitivity to the second entity in Qwen's middle
layers; one shared sensitivity direction transfers about half the full-state
margin correction. A fact-independent intervention at this site raises held-out
Marathi same-person accuracy from 6.5\% to 33.5\%, with 98.5\% different-person
accuracy. Together, these results distinguish script-dependent discrimination
from criterion placement and identify a causal site for improving identity judgments.
\end{abstract}

\section{Introduction}
\begin{figure*}[t]
\centering
\includegraphics[width=\textwidth]{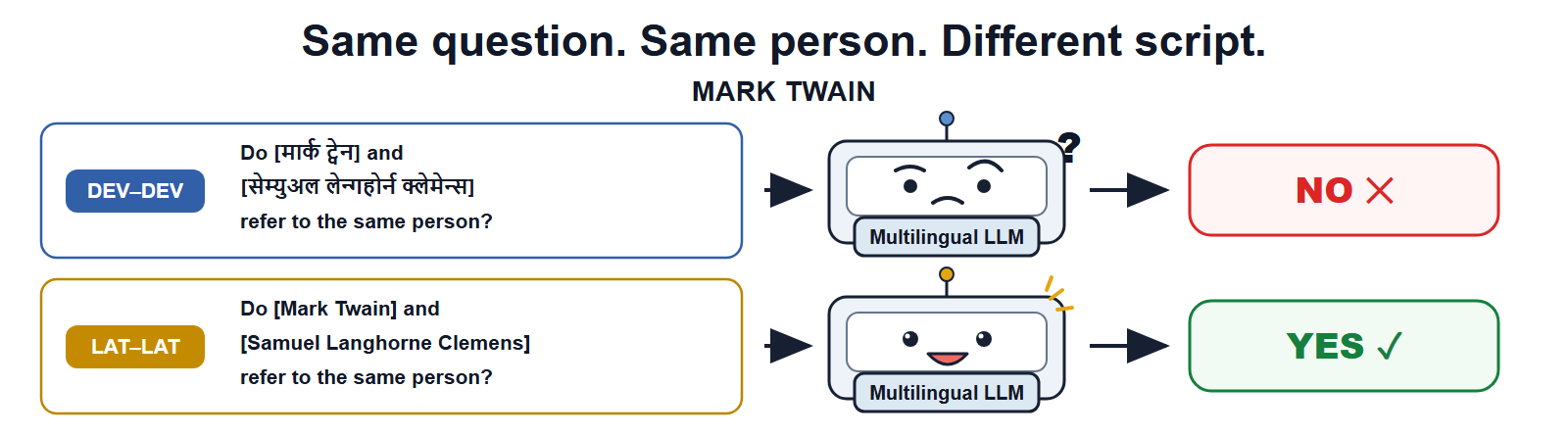}
\caption{Controlled identity verification. Both names and the English question
are supplied; only the attested script-associated name forms change. Qwen and
Llama reject this illustrative Dev--Dev pair and accept its Lat--Lat counterpart.}
\label{fig:marktwain}
\end{figure*}

An identity judgment should remain consistent when the same people are presented
through corresponding names in different scripts. Yet multilingual models can
reject a matching pair in Devanagari and accept its attested Latin forms
(Figure~\ref{fig:marktwain}). This matters for multilingual entity matching:
a system must connect aliases, birth names and pseudonyms even when a user's
writing system differs from the model's most familiar surface forms.

We study this problem through controlled identity verification. Both candidate
names are supplied; only their script-associated surface forms change. The
question language, people and relation remain fixed. Across 601 audited pairs,
Qwen2.5-7B's Marathi same-person accuracy drops from 35.1\% in Lat--Lat to 8.5\%
in Dev--Dev, while it continues to reject almost all different-person pairs.
The contrast isolates a failure to affirm identity within a task the model can
partly solve through other surface forms. A second corpus in Malayalam, a different script and a
Dravidian rather than Indo-Aryan language, shows the effect is not specific to
Devanagari or to Indo-Aryan languages.

Our analysis connects three levels of evidence. First, we establish the
behavioural gap across scripts and prompt languages, separating discrimination
from response bias. Five-model evaluation and romanization controls show why
tokenization efficiency alone is insufficient. Second, Latin-trained probes and
output-score calibration establish how much label information remains available
when the model answers incorrectly. Third, activation patching and subspace
interventions identify where that information can influence the decision.

The resulting account is actionable. In Qwen, the second entity's middle-layer
state is a strong intervention site, and a shared sensitivity direction transfers
roughly half the full-state correction. A vector learned from training entities
raises held-out Marathi same-person accuracy more than fivefold. Controls on
ranking, script condition and position explain this gain: much of it comes from
moving the decision criterion at a causally sensitive site. The script contrast
also changes discrimination, so calibration and script robustness are distinct
evaluation targets. Our contribution is a controlled benchmark and a diagnosis
that connects these targets to measurable internal computations.

\section{Related work}
Multilingual models develop shared, English-aligned representations
\citep{wendler2024llamas,wu2024semantic}, yet factual judgments can vary across
languages. \citet{qi2023crosslingual} examine consistency across translated
queries; \citet{ai2025knowledge} study reference in code-mixed coreferential
statements across language families and scripts, including activation patching.
Our design fixes the question language and varies the two supplied names,
following one identity relation across behavioural and internal measurements.

\citet{wang2025lostmultilinguality} localise cross-lingual factual inconsistency
near the transition to language-specific final-layer representations and propose
a linear shortcut. We examine entity surface forms and find strong middle-layer
sensitivity at the second entity. Testing a related linear intervention against
output calibration and other script conditions clarifies what its accuracy gain
changes. Romanization can improve non-Latin processing
\citep{husain2024romansetu,liu2024transliterations,saji2025romanlens}; our comparison
of automatic and attested forms separates shorter token sequences from reliable
behavioural recovery.

We combine activation patching
\citep{meng2022locating,dumas2025patching,nanda2022transformerlens} with linear
probes \citep{indictunedlens2026}. The controlled task makes it possible to compare
label decodability, output calibration and causal transfer on corresponding inputs.

\section{Task, data and protocol}
\label{sec:task}
\textbf{Identity pairs.} The corpus contains 601 manually audited Wikidata
pairs \citep{vrandecic2014wikidata}, each linking a canonical name to an attested
alias, birth name, pseudonym or nickname.\footnote{Code and corpora:
\url{https://github.com/Multilingual-entity/Entity_Identification}} Both names have observed Devanagari and
Latin forms. Independently varying the two spans yields Dev--Dev, Dev--Lat,
Lat--Dev and Lat--Lat. These are conventional name forms, so the manipulation
includes spelling and familiarity differences associated with script.
Each positive has a different-person hard negative, selected by occupation,
country and relation tier; 99.8\% share occupation. A separate analysis rebuilds
negatives for name similarity (\S\ref{sec:probe}).

The audit excludes identical Devanagari fields, joined aliases and parenthetical
metadata, as well as pairs whose two scripts encode different name relationships.
The retained set comprises 404 observed-alias relations and 197 relations involving
explicit name properties. Names are attested rather than generated by a
transliterator, making the Latin condition a comparison with conventional forms
of the same identity relation. This design supplies both candidates directly,
avoiding the additional requirement to generate an unfamiliar name.

\textbf{A second script.} A Malayalam corpus covers 380 of the same Wikidata
entities, with both names given in Malayalam and in the original Latin forms.
A Malayalam speaker audited all 380 items and all 380 distractors, rejecting
six whose fields carried a generic honorific or a second alias, which leaves
374. Of the 760 Malayalam strings, 171 are exactly attested in Wikidata
Malayalam labels or aliases, 18 are spelling variants of an attested form, and
571 are machine-transliterated and then audited; none of the transliterations
was rejected. The same speaker checked the three Malayalam question templates.
Splits of 222/76/76 are inherited from a seed-reproduced split of the 601-item
corpus rather than recomputed, so no probe, vector or threshold from the
Devanagari experiments is transferred here. We evaluate Qwen2.5-7B
behaviourally on this corpus (Appendix~\ref{app:malayalam}).

\textbf{Languages and scoring.} Marathi and Hindi corpora cover the same
identities and have byte-identical Latin forms; 528/601 Devanagari pairs differ.
Each corpus is queried in its own language and English, with three paraphrases
and both yes/no letter assignments. We score the final-position logit margin
$m=z_{\mathrm{yes}}-z_{\mathrm{no}}$ without sampling. Behavioural accuracy pools
the six prompts per item and reports positive and negative pairs separately.
For entity-level probe and calibration comparisons, margins are averaged across
the six prompts before classification. AUROC measures ranking independently of
the decision threshold. Confidence intervals use 2,000 item-clustered bootstrap
resamples. Appendix~\ref{app:protocol} specifies aggregation for each experiment.

Both Marathi and Hindi templates have been verified by native speakers.
Counterbalancing is also used in the knowledge and causal gates: an item counts
as known only when both answer-key assignments are correct. A fixed preference
for A or B therefore cannot qualify an item for causal analysis. Reporting both
pair polarities further distinguishes improved identity recognition from an
increase in acceptance that also admits unrelated people.

\textbf{Models and selection.} We evaluate Qwen2.5-7B-Instruct
\citep{qwen2025qwen25}, Gemma-2-2b-it \citep{gemma2024}, Llama-3.1-8B-Instruct
\citep{llama2024}, Mistral-Small-24B and Sarvam-M-24B. The 24B models use the
Marathi corpus. Qwen provides the detailed causal analysis. A fixed split of
361/120/120 target items supplies training, validation and test data; distractors
come from a shared pool and can cross splits. Probe layers and intervention
settings are chosen on validation. Vector scales use 60 validation items and
one template, with evaluation on all three templates. Frozen controls reuse the
test set without further fitting. Construction, prompts, gates and compute
details appear in Appendices~\ref{app:corpus}--\ref{app:gates}
and~\ref{app:compute}.

\section{The script gap}
\label{sec:gap}
\subsection{Behaviour across models and languages}
\label{sec:behaviour}
\begin{table*}[t]
\centering
\small
\caption{Accuracy (\%) over 601 items, three paraphrases and both answer keys. Pos. and neg. denote same-person and different-person pairs. English rows use the Marathi corpus; the 24B models were evaluated on that corpus only. Textual gaps use unrounded values.}
\label{tab:behaviour}
\begin{tabular}{llcccccccc}
\toprule
& & \multicolumn{2}{c}{Dev--Dev} & \multicolumn{2}{c}{Dev--Lat} & \multicolumn{2}{c}{Lat--Dev} & \multicolumn{2}{c}{Lat--Lat}\\
\cmidrule(lr){3-4}\cmidrule(lr){5-6}\cmidrule(lr){7-8}\cmidrule(lr){9-10}
Model & Prompt & pos. & neg. & pos. & neg. & pos. & neg. & pos. & neg.\\
\midrule
\rowcolor{resultblue}
Qwen2.5-7B  & Marathi & \textbf{8.5} & 99.8 & 29.4 & 99.1 & 35.0 & 99.8 & \textbf{35.1} & 99.7\\
\rowcolor{resultblue}
            & Hindi & \textbf{16.4} & 99.9 & 45.2 & 98.6 & 52.2 & 99.5 & \textbf{41.1} & 99.6\\
\rowcolor{resultblue}
            & English & \textbf{28.2} & 100.0 & 61.3 & 98.9 & \textbf{66.0} & 98.7 & 56.3 & 99.6\\
Gemma-2-2b  & Marathi & 91.8 & 25.2 & \textbf{96.7} & 27.6 & 92.2 & 29.0 & 96.6 & 34.1\\
            & Hindi & 94.1 & 44.6 & \textbf{97.5} & 38.4 & 91.0 & 52.2 & 97.2 & 49.9\\
            & English & 97.3 & 21.7 & \textbf{99.0} & 21.2 & 97.1 & 32.8 & 98.1 & 41.5\\
Llama-3.1-8B & Marathi & 34.2 & 66.5 & \textbf{35.7} & 66.7 & 34.9 & 66.8 & 35.5 & 68.3\\
            & Hindi & 19.3 & 83.5 & 29.8 & 83.7 & \textbf{33.7} & 84.0 & 28.0 & 86.2\\
            & English & 47.9 & 85.3 & 66.2 & 80.3 & \textbf{73.8} & 87.7 & 67.8 & 93.7\\
\midrule
Mistral-Small 24B & Marathi & 14.3 & 99.2 & 40.0 & 97.2 & \textbf{46.1} & 98.8 & 40.1 & 97.9\\
                  & English & 27.7 & 99.9 & 64.6 & 98.4 & \textbf{69.6} & 98.9 & 68.1 & 99.3\\
Sarvam-M 24B & Marathi & 75.2 & 65.6 & \textbf{83.9} & 64.9 & \textbf{83.9} & 71.1 & 82.0 & 73.9\\
            & English & 82.7 & 97.2 & 87.5 & 96.2 & \textbf{91.3} & 94.4 & 87.6 & 97.2\\
\bottomrule
\end{tabular}
\end{table*}
\textbf{Qwen's deficit is concentrated on matching names.}
Under Marathi prompts, Dev--Dev positive accuracy is 8.5\%, compared with
35.1\% in Lat--Lat; negative accuracy remains at least 98.6\% across the evaluated
conditions (Table~\ref{tab:behaviour}). The gap persists under Hindi and English
prompts. It includes a discrimination loss: Lat--Lat exceeds Dev--Dev in
margin AUROC by 0.037 $[0.025,0.049]$ in Marathi, 0.038 $[0.026,0.051]$ in Hindi
and 0.023 $[0.014,0.033]$ in English over all 601 items.

The effect spans all three paraphrases and substantive relation tiers.
Even among 154 pairs with over 80\% Latin-form similarity, Marathi accuracy is
10.8\% in Dev--Dev and 32.1\% in Lat--Lat. The gap is larger among the 58
least-similar pairs: 31.3 points versus 21.3 in the most-similar band
(Appendix~\ref{app:behaviour}; Table~\ref{tab:overlap}). Thus, near-duplicate
spellings do not account for the phenomenon.

\textbf{The behavioural polarity varies by model.}
Gemma accepts most positives and many negatives, whereas Qwen strongly favours
rejection. Using the signal-detection measure $d'$ to separate these biases,
the Lat--Lat advantage excludes zero for Qwen in Marathi, Gemma in all three
prompt languages and Llama in English (Appendix~\ref{app:dprime}). Llama's
Marathi and Hindi outputs exhibit fixed-letter instruction failures and are
excluded from this interpretation. Mistral-Small-24B has a 25.8-point Marathi
positive-accuracy gap, versus 6.9 for the Indic-adapted Sarvam-M-24B. Their
negative accuracies and checkpoint versions differ, making this a descriptive
family-and-size comparison (Appendix~\ref{app:models}).

Mixed scripts reveal a further distinction. Lat--Dev exceeds Lat--Lat positive
accuracy by 9.7 points in English and 11.1 in Hindi without better per-template
ranking (Appendix~\ref{app:latdev}). Script configuration therefore affects
criterion placement as well as discrimination, motivating the calibration tests below.

\begin{table*}[t]
\centering\small
\caption{Qwen2.5-7B on 374 Malayalam items and the same 374 entities in
Devanagari, each corpus queried in its own language. Script configurations are
Mlym--Mlym, Mlym--Lat, Lat--Mlym and Lat--Lat for Malayalam, and Dev--Dev,
Dev--Lat, Lat--Dev and Lat--Lat for Devanagari. Pos.\ and neg.\ are
same-person and different-person accuracy; AUROC is entity-averaged over six
prompt margins. The gap rows are Lat--Lat minus the both-native condition,
paired by entity, with 95\% item-clustered intervals over 2{,}000 resamples.
The two accuracy gaps differ by 12.9 points, 5.1 of which come from the
both-native cell and 7.8 from the differing Lat--Lat baselines. The Devanagari
column is a new run on this 374-item subset with its own distractor map; it is
not the 601-item corpus of Table~\ref{tab:behaviour} and the two are not
interchangeable. English-prompt results and the held-out split appear in
Table~\ref{tab:malayalamfull}.}
\label{tab:malayalam}
\begin{tabular}{lcccccc}
\toprule
& \multicolumn{3}{c}{Malayalam ($n=374$)} & \multicolumn{3}{c}{Devanagari ($n=374$)}\\
\cmidrule(lr){2-4}\cmidrule(lr){5-7}
Name scripts & pos. & neg. & AUROC & pos. & neg. & AUROC\\
\midrule
Both native & \textbf{5.3} & 100.0 & 0.925 & \textbf{10.4} & 100.0 & 0.935\\
Native, Latin & 33.5 & 98.6 & 0.941 & 33.2 & 98.8 & 0.960\\
Latin, native & 38.1 & 99.6 & 0.964 & 37.8 & 99.8 & 0.970\\
Both Latin & \textbf{48.8} & 99.9 & 0.990 & \textbf{41.0} & 99.6 & 0.981\\
\midrule
Gap, points & \multicolumn{3}{c}{$+43.5\ [40.0, 47.1]$} & \multicolumn{3}{c}{$+30.6\ [26.6, 34.5]$}\\
Gap, AUROC & \multicolumn{3}{c}{$+0.065\ [0.050, 0.082]$} & \multicolumn{3}{c}{$+0.045\ [0.031, 0.061]$}\\
\bottomrule
\end{tabular}
\end{table*}

\textbf{The gap is not specific to Devanagari.} On 374 Malayalam items and the
same 374 entities in Devanagari, Qwen's native-script same-person accuracy is
5.3\% under Malayalam prompts against 10.4\% under Marathi prompts, while
negative accuracy is at least 98.6\% in every cell of both corpora
(Table~\ref{tab:malayalam}). The script gap is correspondingly larger, 43.5
points $[40.0, 47.1]$ against 30.6 $[26.6, 34.5]$, though the Lat--Lat
baselines also differ: of the 12.9-point difference between the two gaps, 5.1
points come from the native cell and 7.8 from the baseline. The margin AUROC gap
is larger as well, 0.065 $[0.050, 0.082]$ against 0.045 $[0.031, 0.061]$, so
discrimination moves and not only the criterion. The Mlym--Mlym cell is
nonetheless this paper's clearest dissociation: in the same cell, decisions sit
near the floor at 5.3\% while ranking sits near the ceiling at 0.925. The
held-out 76 entities agree, at 39.0 and 28.5 points. Under English prompts,
Lat--Mlym exceeds Lat--Lat, 66.4\% against 62.9\%, reproducing the mixed-script
ordering of Table~\ref{tab:behaviour} in a second script
(Table~\ref{tab:malayalamfull}).

\subsection{Tokenization and familiar forms}
\label{sec:tokenizer}
Devanagari names are tokenized roughly three times as finely as Latin names.
Fragmentation correlates weakly with the Latin-script benefit, and longer names
are harder to retrieve in isolation (Appendix~\ref{app:frag}). Nevertheless,
surface features predict script-sensitive items at only 0.56--0.60 out-of-fold
AUC, versus 0.86--0.96 after adding relation states (Table~\ref{tab:auc}).
Five automatic romanizations reduce Qwen's mean name length from 16.5 tokens to
7.7--10.4, but replacement yields no reliable recovery. Substitution of the
attested Latin form reproduces Lat--Lat within 0.004 in all nine primary
model--language cells. The tested reductions in fragmentation are therefore
insufficient; spelling fidelity and familiarity remain relevant
(Appendix~\ref{app:translit}).

The automatic schemes reproduce the attested Latin spelling exactly for only
0.8--4.5\% of names. Thus, moving into Latin characters and presenting a familiar
Latin name are different manipulations. The contrast also appears across
models: Gemma and Llama use substantially fewer Devanagari tokens per character
than Qwen, yet their behaviour still depends on script. Together, the prediction
and substitution tests motivate examining relation-level states beyond surface
token counts.

\section{Available information and decision calibration}
\label{sec:probe}
We fit ridge classifiers ($\alpha=10$) on L2-normalised Lat--Lat hidden states
and evaluate on Dev--Dev, so no Devanagari state enters probe training.
Layers are selected on validation and accuracy is measured on the 120 test
items. Cross-language transfer also succeeds across Qwen, Gemma and Llama
(Appendix~\ref{app:probe}).

\textbf{Label information transfers across scripts.}
Qwen's question-mark probe reaches 0.925 balanced accuracy and 0.867
same-person recall on Marathi Dev--Dev inputs. Averaged over layers 10--18,
it recovers 78.9\% of exact Marathi failures. At the second entity, the
validation-selected probe reaches 0.933 balanced accuracy
(Table~\ref{tab:calvsprobe}). These measurements establish accessible
label-relevant structure in states associated with incorrect decisions.

\textbf{Similarity-matched negatives strengthen the comparison.}
On the original negative map, a five-feature string classifier reaches 0.933,
matching the probe. Its features are sequence similarity, token and character
Jaccard overlap, relative length difference and final-token agreement. We rebuild distractors to approximate each positive pair's
name similarity; matching within 0.05 is possible for 144/601 items, including
25 test items (Table~\ref{tab:matched-main}). Across all 120 test items under this partially matched map, the
question-mark probe exceeds the string classifier: 0.908 versus 0.796 in
Marathi, a difference of $+0.112$ $[+0.067,+0.163]$, and $+0.083$
$[+0.025,+0.142]$ in English. On the tightly matched subset, model separation
remains within 0.01 of its original value. The signal extends beyond these
five string features; richer orthographic signals and familiarity are not
separated from identity retrieval by this test (Appendix~\ref{app:surface}).

Both classifiers are refitted on the rebuilt training negatives, and the probe
layer is reselected on validation. This compares their predictive information
under the same modified task. Where the string classifier fails, the probe is
correct on 65.3\% of Marathi pairs and 75.5\% of English pairs. The comparison
therefore tests complementary information as well as aggregate accuracy.

\begin{table*}[t]
\centering\small
\caption{Question-mark probe and string baseline on the same 120 held-out
entities under the original and partially similarity-matched negative maps.
Values are balanced accuracy; paired differences include 95\% item-clustered
intervals. The matched map permits tight matching on 25 test items.
Full controls appear in Appendix~\ref{app:surface}.}
\label{tab:matched-main}
\begin{tabular}{llccc}
\toprule
Prompt & Negative map & Probe & String baseline & Difference\\
\midrule
Marathi & Original & 0.925 & 0.933 & $-0.008$ [$-0.033$, $+0.017$]\\
        & Matched & \textbf{0.908} & 0.796 & $+0.112$ [$+0.067$, $+0.163$]\\
English & Original & 0.950 & 0.933 & $+0.017$ [$-0.004$, $+0.042$]\\
        & Matched & \textbf{0.879} & 0.796 & $+0.083$ [$+0.025$, $+0.142$]\\
\bottomrule
\end{tabular}
\end{table*}

\begin{table*}[t]
\centering\small
\caption{Entity-level decisions on 120 held-out Dev--Dev items, averaging six prompt margins before classification. BA is balanced accuracy; TPR/FPR are true-/false-positive rates. All settings are validation-selected. The original negative map is used here; similarity-matched results appear in Table~\ref{tab:probematched}.}
\label{tab:calvsprobe}
\begin{tabular}{lccc|ccc}
\toprule
& \multicolumn{3}{c|}{Marathi} & \multicolumn{3}{c}{English}\\
\cmidrule(lr){2-4}\cmidrule(lr){5-7}
Rule & BA & TPR & FPR & BA & TPR & FPR\\
\midrule
Raw model & 0.525 & 0.058 & 0.008 & 0.633 & 0.267 & 0.000\\
Margin + constant & 0.879 & 0.808 & 0.050 & 0.904 & 0.875 & 0.067\\
E2 probe & \textbf{0.933} & 0.883 & 0.017 & \textbf{0.954} & 0.942 & 0.033\\
String features & \textbf{0.933} & 0.883 & 0.017 & 0.933 & 0.883 & 0.017\\
\bottomrule
\end{tabular}
\end{table*}
\Needspace{3\baselineskip}
\textbf{Calibration explains much of the behavioural shortfall.}
A single constant added to the model's margin, selected for balanced accuracy
on validation, raises Marathi entity-level balanced accuracy from 0.525 to
0.879. This recovers 87\% of the distance to the E2 probe's 0.933.
The probe retains an advantage of $+0.054$ $[+0.017,+0.092]$ in Marathi and
$+0.050$ $[+0.021,+0.079]$ in English. These are comparisons at their selected
operating points, with false-positive rates reported in Table~\ref{tab:calvsprobe}.
Substantial discriminative information reaches the model's own output score,
and criterion placement accounts for much of its failure to use that
information in the final answer.

\section{Localising causal sensitivity}
\label{sec:patching}
We replace a Dev--Dev residual state with the corresponding state from a
Latin-bearing donor run and measure the change in $m$. The causal gate requires
failure in Dev--Dev and success in the donor condition across paraphrases and
answer keys. Held-out gates contain 12 Marathi, 15 Hindi and 24 English items.
The layer band 10--18 was fixed from a pilot; intervals resample items, and flip
rates are descriptive at these sample sizes (Appendix~\ref{app:patch}).

\begin{table*}[t]
\centering\small
\caption{Held-out Qwen patching: mean margin change with 95\% item-clustered
intervals. Site effects average layers 10--18 and three Latin-bearing donors.
The last row compares correct and wrong-entity Lat--Lat donors on matched
layers and items. Full profiles appear in Appendix~\ref{app:patch}.}
\label{tab:patch-main}
\begin{tabular}{lccc}
\toprule
Site / contrast & Marathi ($n=12$) & Hindi ($n=15$) & English ($n=24$)\\
\midrule
First entity (E1) & 0.12 [$-0.13$, 0.38] & $-0.03$ [$-0.25$, 0.16] & 0.27 [0.02, 0.53]\\
Second entity (E2) & \textbf{2.66} [1.67, 3.71] & \textbf{3.16} [2.47, 3.90] & \textbf{2.85} [2.28, 3.45]\\
Question mark & 0.61 [0.33, 0.97] & 0.33 [0.16, 0.52] & 0.79 [0.59, 1.00]\\
Readout & 1.26 [0.96, 1.58] & 1.08 [0.87, 1.30] & 1.33 [1.12, 1.57]\\
E2: correct $-$ wrong entity & 2.09 [0.41, 3.84] & $-0.51$ [$-2.06$, 1.08] & 0.70 [$-0.21$, 1.54]\\
\bottomrule
\end{tabular}
\end{table*}

\textbf{The second entity is the strongest site.}
E2 patching raises margins by 2.66--3.16 across languages, while E1 has little
effect (Table~\ref{tab:patch-main}). Reverse patching damages the working run.
The matched correct-versus-wrong-entity contrast establishes an entity-specific
component in Marathi; the Hindi and English intervals include zero.
Joint E1/E2/question-mark patching across layers 12--17 flips 50\%, 60\% and
71\% of held-out Marathi, Hindi and English items. Matched-negative accuracy
remains 100\%. At layers 18--20 the joint effect falls below one margin point,
while sensitivity shifts toward the question mark.

The donor-language control further separates name form from question language.
At the peak over layers 10--18, an English Lat--Lat donor changes the Marathi
margin by 6.26, comparable to 5.97 from a Marathi Lat--Lat donor; an English
Dev--Dev donor gives 0.11. These peak values use a different aggregation from
Table~\ref{tab:patch-main}, so they support the donor contrast rather than an
additional site comparison (Table~\ref{tab:donor}). The effective transfer can
therefore cross prompt language while retaining the Latin-bearing name context.

\subsection{Position, name and computation}
\label{sec:orderswap}
Swapping the canonical name and alias preserves E2 dominance: Marathi E2
patching changes the margin by 4.79 in the original order and 3.78 after swapping,
while E1 stays within 0.42 of zero across all six language--order combinations.
The effect follows the second slot, the first position that can attend to both
names (Appendix~\ref{app:orderswap}). Nonce names also show E2-over-E1 sensitivity,
placing the finding in relational comparison more broadly
(Appendix~\ref{app:nonce}).

Cumulative attention patches transfer more correction than MLP patches:
5.41 versus 0.51 in Marathi, with the same ordering in Hindi and English.
These interventions are not an additive decomposition of residual effects.
Multi-head patches exceed individual-head patches, consistent with distributed
transfer (Appendix~\ref{app:components}). Exploratory Llama and Mistral-Small
sweeps also show E2-over-E1 sensitivity at different depths and smaller
magnitudes (Appendices~\ref{app:llama} and~\ref{app:mistral}).

\subsection{A shared direction transfers correction}
\label{sec:jacobian}
\label{sec:mediator}
To identify effective directions at E2, we differentiate $m$ with respect to
the residual state. Dev--Dev and Lat--Lat local gradients have cosine
similarity 0.37--0.61 across layers 12--17 in Marathi, against at most 0.004
in magnitude for random directions. Their partial alignment motivates testing
which shared subspaces carry the donor correction (Appendix~\ref{app:jacobian}).

For an orthonormal basis $U$, we intervene with
\begin{equation}
  h_D' = h_D + UU^{\top}(h_L-h_D).
\end{equation}
Directions are fitted on training items. The sensitivity direction is the
normalised mean Dev--Dev gradient over gated training items; alternatives use
the mean donor difference, its principal component, a ridge-probe direction or
a random direction. Layer 16 and rank one are validation-selected; the frozen
test fold evaluates the held-out paraphrase P2 (Appendix~\ref{app:mediator}).

\begin{table*}[t]
\centering\small
\begin{tabular}{lcccccc}
\toprule
& \multicolumn{3}{c}{Marathi ($n=12$)} & \multicolumn{3}{c}{English ($n=24$)} \\
\cmidrule(lr){2-4}\cmidrule(lr){5-7}
Direction & repair & signed & $|\Delta|$ & repair & signed & $|\Delta|$ \\
\midrule
Whole patch & 2.54 & 0.48 & 1.17 & 4.90 & 1.03 & 1.98 \\
\midrule
Local sensitivity & 1.30 & 0.23 & 0.47 & 2.65 & 0.79 & 1.01 \\
Mean difference & 3.50 & 1.81 & 1.82 & 4.33 & 2.67 & 2.67 \\
Donor PCA & 0.09 & $-0.05$ & 0.16 & 0.43 & 0.19 & 0.22 \\
Probe & $-0.01$ & 0.00 & 0.13 & $-0.01$ & 0.02 & 0.08 \\
Random & $-0.01$ & 0.01 & 0.11 & 0.05 & $-0.01$ & 0.09 \\
\bottomrule
\end{tabular}
\caption{Rank-one donor-projected patching at E2, layer 16, held-out fold t01e2. Repair is the gated margin change; signed and $|\Delta|$ measure changes on never-correct same-person pairs outside the gate. Negative accuracy is 1.00 throughout. The whole-state patch is the reference.}
\label{tab:mediator}
\end{table*}
One sensitivity direction transfers a Marathi margin gain of 1.30
$[0.73,1.95]$, versus 2.54 for a full-state patch, and 2.65 $[2.06,3.34]$
versus 4.90 in English (Table~\ref{tab:mediator}). The probe direction transfers
approximately zero under the same donor projection. Linear decodability and
donor-projected causal transfer therefore differ at this site. This comparison
measures their combination of donor projection and downstream sensitivity;
it does not establish that the probe direction is unused.

The mean-difference direction produces larger but broader changes, also moving
same-person pairs outside the gate. Negative accuracy remains 1.00 in these
subspace tests. The rank-one result concerns margin correction: the substantial
answer flips above require multi-layer intervention. The next experiment tests
whether a shared vector can achieve gains without a test item's donor.

The distinction between margin transfer and answer correction connects the two
experiments. A state intervention may move the decision score substantially
without crossing its threshold. Sustained patching across the middle-layer band
produces answer flips that the single-layer tests rarely achieve. This motivates
applying a shared correction over the same band, and measuring both its accuracy
and the ranking of positive against negative pairs.

\section{Improving judgments through a shared intervention}
\label{sec:repair}
At each layer, we estimate a vector from 361 training items by summing the mean
E2 changes associated with replacing either name's script:
\begin{align}
 a_\ell &= \mathbb{E}_{i\in\mathcal{T}}\big[
 h^{LD}_{i,\ell}+h^{DL}_{i,\ell}-2h^{DD}_{i,\ell}\big],\\
 h'_{\ell,E2} &= h_{\ell,E2}+\alpha a_\ell/\|a_\ell\|_2.
\end{align}
Here LD, DL and DD denote Lat--Dev, Dev--Lat and Dev--Dev. Vectors are
layer-specific and shared across items. Validation selects the layers 12--17
window and scale by balanced accuracy, with negative accuracy as a tiebreak;
no negative-accuracy floor is imposed (Appendix~\ref{app:window}).

\textbf{The gain generalises to held-out entities.}
Marathi same-person accuracy rises from 0.065 to 0.335, with negative accuracy
0.985; English rises from 0.283 to 0.590, with negative accuracy 0.976.
Marathi gains occur on all three templates despite selection on one.
The Hindi rendering also improves, with a larger negative-accuracy tradeoff
(Table~\ref{tab:winreplicate}). Applying the frozen Marathi intervention at E1
instead yields only 0.071 positive accuracy. The effective position agrees
with the patching localisation (Table~\ref{tab:site}).

The multi-layer setting matters. The earlier single-layer intervention raises
Marathi positive accuracy only to 0.083 on test. Holding the scale at 6 while
widening to layers 12--17 raises validation accuracy from 0.083 to 0.233;
extending the scale search then reaches 0.392. Balanced-accuracy selection
returns scales of 12 for Marathi and 24 for English on the Marathi corpus.
The held-out gains thus follow a validation-selected combination of depth and
magnitude (Appendices~\ref{app:recovery} and~\ref{app:window}).

\begin{table*}[t]
\centering\small
\caption{Held-out intervention and calibration. AUROC averages six margins per entity; change includes a 95\% item-clustered interval. TPR (FPR) pools prompts. The offset is validation-selected under a negative-accuracy floor; threshold only is an oracle test-curve diagnostic. Appendix~\ref{app:protocol} gives selection rules.}
\label{tab:calibration}
\begin{tabular}{lcc|ccc}
\toprule
 & \multicolumn{2}{c|}{Entity-averaged AUROC} & \multicolumn{3}{c}{TPR (FPR)}\\
\cmidrule(lr){2-3}\cmidrule(lr){4-6}
Arm & Dev--Dev & change & vector & offset & threshold only\\
\midrule
Marathi & 0.952 & $+0.003$ $[-0.008, +0.015]$ & 0.335 (0.015) & 0.312 (0.012) & \textbf{0.336} (0.015)\\
English (mr corpus) & 0.961 & $-0.003$ $[-0.016, +0.012]$ & 0.590 (0.024) & \textbf{0.714} (0.037) & 0.644 (0.024)\\
Hindi & 0.935 & $-0.016$ $[-0.038, +0.005]$ & 0.546 (0.131) & \textbf{0.656} (0.178) & 0.594 (0.131)\\
\bottomrule
\end{tabular}
\end{table*}
\textbf{Controls identify what improves.}
Entity-averaged AUROC changes are small and their intervals include zero
(Table~\ref{tab:calibration}). A validation-selected scalar offset reaches a
nearby Marathi operating point, 0.312 true-positive rate at 0.012 false-positive
rate, versus the vector's 0.335 at 0.015. An oracle threshold on the test curve
matches the vector's Marathi rate; this is a diagnostic, not a deployable rule.
Per-template analysis also reveals a ranking component: on the tuning template
the vector reaches 0.425 true-positive rate versus 0.308 for threshold adjustment
at a matched false-positive rate, and ranking improves on templates 0 and 2
(Table~\ref{tab:pertemplate}).

The frozen vector improves Lat--Lat too, raising Marathi positive accuracy
from 0.340 to 0.629 and English from 0.529 to 0.756. Treating both script
conditions narrows the disparity in three of four corpus--language arms and
widens it by 1.9 points in Marathi (Appendix~\ref{app:control}). These controls
identify a broadly useful criterion shift with template-dependent ranking
effects. The intervention improves identity judgments at the identified site;
script-specific parity remains a separate objective.

These measurements distinguish three useful outcomes for multilingual identity
verification. Decoding tests whether states support the label; AUROC tests
whether the model's scores separate matching from non-matching names; and
accuracy at a reported false-positive rate tests the decisions a user receives.
The probe, patch and steering experiments connect these outcomes without
assuming that improvement in one guarantees improvement in the others.
In particular, evaluating the same frozen intervention in both script conditions
reveals whether an accuracy gain also reduces their disparity. This makes the
second-entity site useful for controlled improvement and gives subsequent
interventions a concrete evaluation protocol.

\section{Conclusion}
Controlled name-form substitutions expose substantial cross-script identity
matching failures. Cross-script probes, calibration and causal interventions
show that a large behavioural deficit can coexist with accessible label
information. In Qwen, the second entity's middle-layer states provide an
effective intervention site: a shared sensitivity direction transfers about
half the full-state margin correction, and a fact-independent window
intervention raises held-out Marathi same-person accuracy from 6.5\% to 33.5\%
while retaining 98.5\% negative accuracy. Evaluating discrimination, decision
criterion and both script conditions together makes these gains interpretable
and provides a basis for improving multilingual identity verification.
\label{page:main-end}

\clearpage
\typeout{ACL-MAIN-PAGES: \number\numexpr\value{page}-1\relax}
\section*{Limitations}
\textbf{Coverage.} The corpus covers 601 person-name relations and two scripts,
with Marathi, Hindi and English prompts. Hindi reuses the Marathi identities
with different Devanagari renderings, testing transfer across renderings
rather than an independent entity sample. A second corpus covers 374 of the same relations in Malayalam and
Latin, with Malayalam and English prompts, evaluated behaviourally on Qwen only;
no probe, patching or intervention result extends to it. Other writing systems, entity types and factual relations require further study.

\textbf{Scope of the diagnosis.} Detailed mechanistic results concern Qwen.
Other models provide behavioural and probe comparisons and exploratory causal
analyses, with differing response biases and effective layer ranges. The 24B
comparison matches family and parameter count but uses Mistral-Small 3.0,
whereas Sarvam-M derives from 3.1; it does not isolate Indic post-training.
Sarvam-M's causal gate contains only one Marathi item.

\textbf{Information and causal interpretation.} Probes establish label
decodability, and the matched-negative analysis exceeds a five-feature surface
baseline. Neither uniquely identifies retrieval of an identity association:
learned orthographic regularities and familiarity may contribute. Tight
similarity matching is available for 144/601 items. Nonce names alter response
bias as well as entity content, limiting conclusions about entity specificity.
The E2 result identifies an intervention site rather than the location of a
stored fact; local gradients describe sensitivity near the measured state.

\textbf{Evaluation dependence.} Target items are split into training,
validation and test sets, but globally selected distractors can cross those
boundaries (Appendix~\ref{app:probe}). The stable causal gate yields only 12--24
held-out items per language, motivating item-level intervals and descriptive
flip rates. The same held-out split was evaluated for the single-layer vector,
window vector, condition control and site control. Settings were selected on
validation and the latter controls were frozen, but these analyses share test
items and are not independent replications. The sensitivity-vector steering
sweep is validation-only and is reported separately from held-out results.

\clearpage
\bibliography{references_acl}

@inproceedings{wendler2024llamas,
  title     = {Do Llamas Work in {E}nglish? On the Latent Language of Multilingual Transformers},
  author    = {Wendler, Chris and Veselovsky, Veniamin and Monea, Giovanni and West, Robert},
  booktitle = {Proceedings of the 62nd Annual Meeting of the Association for Computational Linguistics (ACL)},
  year      = {2024},
  url       = {https://arxiv.org/abs/2402.10588}
}

@inproceedings{wu2024semantic,
  title     = {The Semantic Hub Hypothesis: Language Models Share Semantic Representations Across Languages and Modalities},
  author    = {Wu, Zhaofeng and Yu, Xinyan Velocity and Yogatama, Dani and Lu, Jiasen and Kim, Yoon},
  booktitle = {International Conference on Learning Representations (ICLR)},
  year      = {2025},
  url       = {https://arxiv.org/abs/2411.04986}
}

@misc{saji2025romanlens,
  title  = {{RomanLens}: The Role of Latent Romanization in Multilinguality in {LLMs}},
  author = {Saji, Alan and Husain, Jaavid Aktar and Jayakumar, Thanmay and Dabre, Raj and Kunchukuttan, Anoop and Puduppully, Ratish},
  year   = {2025},
  note   = {arXiv:2502.07424}
}

@inproceedings{liu2024transliterations,
  title     = {How Transliterations Improve Crosslingual Alignment},
  author    = {Liu, Yihong and Wang, Mingyang and Kargaran, Amir Hossein and Imani, Ayyoob and Xhelili, Orgest and Ye, Haotian and Ma, Chunlan and Yvon, Fran{\c{c}}ois and Sch{\"u}tze, Hinrich},
  booktitle = {COLING 2025},
  year      = {2025},
  note      = {arXiv:2409.17326}
}

@inproceedings{husain2024romansetu,
  title     = {{RomanSetu}: Efficiently Unlocking Multilingual Capabilities of Large Language Models via Romanization},
  author    = {Husain, Jaavid Aktar and Dabre, Raj and Kumar, Aswanth and Gala, Jay and Jayakumar, Thanmay and Puduppully, Ratish and Kunchukuttan, Anoop},
  booktitle = {Proceedings of the 62nd Annual Meeting of the Association for Computational Linguistics (ACL)},
  year      = {2024},
  url       = {https://arxiv.org/abs/2401.14280}
}

@inproceedings{dumas2025patching,
  title     = {Separating Tongue from Thought: Activation Patching Reveals Language-Agnostic Concept Representations in Transformers},
  author    = {Dumas, Cl{\'e}ment and Wendler, Chris and Veselovsky, Veniamin and Monea, Giovanni and West, Robert},
  booktitle = {Proceedings of the 63rd Annual Meeting of the Association for Computational Linguistics (Volume 1: Long Papers)},
  pages     = {31822--31841},
  address   = {Vienna, Austria},
  year      = {2025},
  url       = {https://aclanthology.org/2025.acl-long.1536/}
}

@misc{nanda2022transformerlens,
  title  = {{TransformerLens}: A Library for Mechanistic Interpretability of Generative Language Models},
  author = {Nanda, Neel and Bloom, Joseph},
  year   = {2022},
  note   = {\url{https://github.com/TransformerLensOrg/TransformerLens}}
}

@misc{indictunedlens2026,
  title  = {{Indic-TunedLens}: Interpreting Multilingual Models in {I}ndian Languages},
  author = {Panchal, Mihir and Varshney, Deeksha and {Mamta} and Ekbal, Asif},
  year   = {2026},
  note   = {arXiv:2602.15038}
}

@inproceedings{qi2023crosslingual,
  title     = {Cross-Lingual Consistency of Factual Knowledge in Multilingual Language Models},
  author    = {Qi, Jirui and Fern{\'a}ndez, Raquel and Bisazza, Arianna},
  booktitle = {Proceedings of the 2023 Conference on Empirical Methods in Natural Language Processing (EMNLP)},
  year      = {2023},
  note      = {arXiv:2310.10378. Outstanding Paper Award}
}

@inproceedings{ai2025knowledge,
  title     = {Are Knowledge and Reference in Multilingual Language Models Cross-Lingually Consistent?},
  author    = {Ai, Xi and Ihsani, Mahardika Krisna and Kan, Min-Yen},
  booktitle = {Findings of the Association for Computational Linguistics: EMNLP 2025},
  pages     = {4975--5011},
  year      = {2025},
  doi       = {10.18653/v1/2025.findings-emnlp.267},
  url       = {https://aclanthology.org/2025.findings-emnlp.267/}
}

@article{vrandecic2014wikidata,
  title   = {Wikidata: A Free Collaborative Knowledgebase},
  author  = {Vrande{\v{c}}i{\'c}, Denny and Kr{\"o}tzsch, Markus},
  journal = {Communications of the ACM},
  volume  = {57},
  number  = {10},
  year    = {2014}
}

@inproceedings{meng2022locating,
  title     = {Locating and Editing Factual Associations in {GPT}},
  author    = {Meng, Kevin and Bau, David and Andonian, Alex and Belinkov, Yonatan},
  booktitle = {Advances in Neural Information Processing Systems (NeurIPS)},
  year      = {2022},
  note      = {arXiv:2202.05262}
}

@misc{qwen2025qwen25,
  title  = {Qwen2.5 Technical Report},
  author = {{Qwen Team}},
  year   = {2025},
  note   = {arXiv:2412.15115}
}

@techreport{llama2024,
  title        = {The Llama 3 Herd of Models},
  author       = {{Llama Team, AI @ Meta}},
  institution  = {Meta},
  year         = {2024},
  note         = {arXiv:2407.21783. Weights obtained from the NousResearch mirror of
                  Llama-3.1-8B-Instruct},
  url          = {https://arxiv.org/abs/2407.21783}
}

@techreport{gemma2024,
  title        = {Gemma 2: Improving Open Language Models at a Practical Size},
  author       = {{Gemma Team}},
  institution  = {Google DeepMind},
  year         = {2024},
  note         = {arXiv:2408.00118},
  url          = {https://arxiv.org/abs/2408.00118}
}

@inproceedings{wang2025lostmultilinguality,
  title     = {Lost in Multilinguality: Dissecting Cross-lingual Factual Inconsistency in Transformer Language Models},
  author    = {Wang, Mingyang and Adel, Heike and Lange, Lukas and Liu, Yihong and Nie, Ercong and Str{\"o}tgen, Jannik and Sch{\"u}tze, Hinrich},
  booktitle = {Proceedings of the 63rd Annual Meeting of the Association for Computational Linguistics (Volume 1: Long Papers)},
  address   = {Vienna, Austria},
  year      = {2025},
  url       = {https://aclanthology.org/2025.acl-long.253/}
}
\clearpage
\appendix
\renewcommand{\thesection}{\AlphAlph{\value{section}}}

\section{Corpus construction}
\label{app:corpus}

Items are identity relations retrieved from Wikidata \citep{vrandecic2014wikidata}:
one entity, two attested surface names. Both a Devanagari and a Latin form of each
name must be observed in the knowledge base; no Latin form is machine-generated
(\texttt{generated\_lat} is false for all 601 retained items, and the Devanagari
language field is Marathi throughout).

The Hindi corpus used in the replication covers the same 601 Wikidata entities with
Hindi renderings of the Devanagari forms; the Latin forms are byte-identical to the
Marathi corpus in all 601 items, while the Devanagari forms differ in 528 of the 601
items and in 902 of the 1{,}202 individual name forms. Because the
entity set and the Wikidata relations are unchanged, the tier distribution in
Table~\ref{tab:tiers} applies to both corpora.

\begin{table*}[tp]
\centering\small
\caption{Relation tiers, 601 retained items.}
\label{tab:tiers}
\begin{tabular}{lr}
\toprule
Tier & Items\\
\midrule
Alias observed only                               & 404\\
Explicit relation plus attested alias, birth name & 171\\
Explicit plus alias, pseudonym                    & 13\\
Explicit plus alias, nickname                     & 7\\
Explicit relation observed, birth name            & 5\\
Explicit relation observed, nickname              & 1\\
\midrule
Total & 601\\
\bottomrule
\end{tabular}
\end{table*}

\textbf{Audit and exclusion.} Every candidate was audited automatically for script
purity of each field and manually for whether the two fields hold the same name rather
than two different names of the same person, which is the failure mode the earlier
version of this corpus did not catch. Items whose two Devanagari fields were identical,
whose fields carried a parenthetical disambiguator or several aliases joined by a
separator, or whose Devanagari was a character mapping of the Latin rather than a
Marathi rendering, were removed. The audit record is in \texttt{data\_audit.csv}, in which all 601 retained items
pass with no issues recorded. Exclusion happened during corpus construction rather
than at pipeline time, so \texttt{excluded\_rows.csv} is empty in every run; the
items that were removed are listed in Appendix~\ref{app:errors}.

\textbf{Hard negatives.} Each item is scored against all 600 others and paired with the
highest scorer, where the score awards 100 for a shared occupation QID, 20 for a
shared country QID, 5 for a matching relation tier and 2 for a matching Devanagari
language, plus a hash-based tiebreak below 0.01. Lady Gaga, for example, is paired with Reese Witherspoon rather than an arbitrary
entity, so that rejecting the negative takes more than noticing that the two names are
unrelated in occupation or country. The pairs are not matched on name similarity, so
this does not establish that rejection requires entity knowledge; Appendix~\ref{app:surface}
measures what a classifier reading only the strings can do.

Of 601 items, 600 (99.8 percent) are paired with a distractor sharing an occupation
and 545 (90.7 percent) with one sharing a recorded country. Five further pairs have
no country recorded on either side, which the matcher credits as agreement, so the
matcher's own count of country matches is 550. All share at least one of these
attributes; the sole item without an occupation match shares a country. The negatives were constructed
once and are shared unchanged across all models, which aids comparability but means
this component has not been independently rebuilt.

\textbf{Splits.} Items are partitioned 361 train / 120 validation / 120 test by
\texttt{fact\_id}, fixed by seed 20260813 and reused unchanged by every downstream
stage.

\section{Prompts}
\label{app:prompts}

\textbf{Template review.} Both Marathi and Hindi question templates have been
verified by native speakers.

Prompt count per model per corpus is $601 \text{ items} \times 4 \text{ script
conditions} \times 2 \text{ polarities} \times 2 \text{ answer keys} \times 3
\text{ paraphrases} \times 2 \text{ prompt languages} = 57{,}696$, and 115{,}392 across
the two corpora.

\textbf{Answer-key counterbalancing.} Each question is asked twice, once with A denoting
yes and once with B denoting yes. A model that ignores the question and prefers a letter
scores exactly one of the two, so an item counts as known only if both orders are
answered correctly. This discards letter preference as evidence of knowledge, and is why
every accuracy in this paper is pooled over both answer keys.

\textbf{The question templates.} Table~\ref{tab:templatestrings} gives three paraphrases
per prompt language, with {\tt [a]} and {\tt [b]} the two supplied names. The source of this paper builds with pdf\LaTeX{} and
carries no Devanagari font, so the Devanagari strings are transliterated here; the exact
forms are in {\tt question\_templates.json} in the
released code. Paraphrase P2 asks whether the canonical name {\tt [a]} is another name of the alias
{\tt [b]}, rather than whether the two denote one person, and it is the hardest paraphrase
in every condition, including Lat--Lat (Appendix~\ref{app:behaviour}). The answer
instruction is appended in English for every prompt language,
unchanged, so the response format never varies with the script contrast:
{\tt Answer only A or B. \{yes\} = yes. \{no\} = no.}

\begin{table*}[tp]
\centering\small
\caption{The three question paraphrases per prompt language. Both Marathi and Hindi
templates have been verified by native speakers. The paraphrases are matched in function across languages: the same person, names for one
person, and whether one is another name for the other.}
\label{tab:templatestrings}
\begin{tabular}{llp{0.56\linewidth}}
\toprule
Language & \# & Template\\
\midrule
\multirow{3}{*}{Marathi}
 & 0 & {\em [a] ani [b] he ekac vyakti ahet ka?}\newline Are [a] and [b] one and the same person?\\
 & 1 & {\em [a] ani [b] hi nave ekac vyaktila darshavatat ka?}\newline Do the names [a] and [b] denote the same person?\\
 & 2 & {\em [a] he [b] yance dusre nav ahe ka?}\newline Is [a] another name of [b]?\\
\midrule
\multirow{3}{*}{Hindi}
 & 0 & {\em kya [a] aur [b] ek hi vyakti hain?}\newline Are [a] and [b] the same person?\\
 & 1 & {\em kya [a] aur [b] ek hi vyakti ke nam hain?}\newline Are [a] and [b] names of the same person?\\
 & 2 & {\em kya [a] [b] ka dusra nam hai?}\newline Is [a] another name for [b]?\\
\midrule
\multirow{3}{*}{English}
 & 0 & Do [a] and [b] refer to the same person?\\
 & 1 & Are [a] and [b] names for the same person?\\
 & 2 & Is [a] another name for [b]?\\
\bottomrule
\end{tabular}
\end{table*}

Scoring reads the logits at the final position and takes the difference between the
token for the yes-letter and the token for the no-letter. No text is generated, so scoring involves no
sampling. It is not bitwise reproducible across batch compositions: bfloat16 reductions
depend on batch shape, which moves margins by up to 1.5 points near the decision
boundary and flips 126 of 28{,}848 positive prompts, 0.4 percent, without changing
any reported comparison.

\section{Knowledge and causal gates}
\label{app:gates}

An item counts as known in a script condition only if both answer-key orders are
answered correctly, which excludes letter preference from counting as knowledge. In
this run, under Marathi prompts, 385 of 601 items are known in at
least one condition, 201 correct in Lat--Lat and wrong in Dev--Dev, and 11 in the
reverse direction. Under English prompts the counts are 542 known, 180 and 32. The
201 to 11 asymmetry is the first evidence that the effect is directional rather than
noise.

The causal gate additionally requires that an item fail in Dev--Dev and succeed in the
donor condition, on every paraphrase and both answer keys. In the
present run 1202 candidate rows (601 items $\times$ 2 prompt languages) yield 169
stable items in Qwen, 53 Marathi and 116 English, and 160 in Gemma under the margin
gate read on negative pairs, 33 Marathi and 127 English. On Llama the Marathi arm
yields none. Table~\ref{tab:gatecounts} collects the counts.

\begin{table*}[tp]
\centering\small
\caption{Causal sample sizes. ``Gate'' counts all qualifying items, ``run'' counts
executed items after the per-language cap, and ``test'' counts unique held-out items used
for test estimates. Qwen runs its complete gate in every cell; the Gemma and Llama
rows retain the cap of 60 they were produced with. Qwen and Llama are gated on positive pairs;
Gemma is gated on negative pairs.}
\label{tab:gatecounts}
\begin{tabular}{llrrr}
\toprule
Model & Prompt & Gate & Run & Test\\
\midrule
Qwen2.5-7B & Marathi & 53 & 53 & 12\\
           & Hindi & 50 & 50 & 15\\
           & English (Marathi run) & 116 & 116 & 24\\
           & English (Hindi run) & 107 & 107 & 24\\
\midrule
Gemma-2-2b & Marathi & 33 & 33 & 10\\
           & Hindi & 65 & 60 & 12\\
           & English (Marathi run) & 127 & 60 & 24\\
           & English (Hindi run) & 125 & 60 & 26\\
\midrule
Llama-3.1-8B & Marathi & 0 & 0 & 0\\
           & Hindi & 0 & 0 & 0\\
           & English (Marathi run) & 15 & 15 & 3\\
           & English (Hindi run) & 17 & 17 & 3\\
\bottomrule
\end{tabular}
\end{table*}

\begin{table*}[tp]
\centering\small
\caption{Distribution of Dev--Dev positive rate among candidate items. The causal gate
requires exactly 0.0.}
\label{tab:gate}
\begin{tabular}{lrrrrrrr}
\toprule
Dev--Dev rate & 0.00 & 0.17 & 0.33 & 0.50 & 0.67 & 0.83 & 1.00\\
\midrule
Qwen  & \textbf{497} & 25 & 21 & 8  & 41 & 3   & 6\\
Gemma & \textbf{0}   & 0  & 0  & 1  & 81 & 131 & 388\\
\bottomrule
\end{tabular}
\end{table*}

Table~\ref{tab:gate} explains why no Gemma item passes the binary gate. Gemma accepts nearly
every positive pair, so Dev--Dev failures essentially do not occur, and a gate defined
on binary correctness cannot fire. This is a property of the measuring instrument under a strong acceptance bias, not
evidence that the effect is absent in Gemma. The margin gate read on negative pairs,
used for Gemma throughout this paper, is the remedy we adopt; this table therefore
describes a diagnostic that Gemma fails, not the gate its causal results use.

\section{Full behavioural results}
\label{app:behaviour}

\begin{figure*}[tp]
  \centering
  \includegraphics[width=\linewidth]{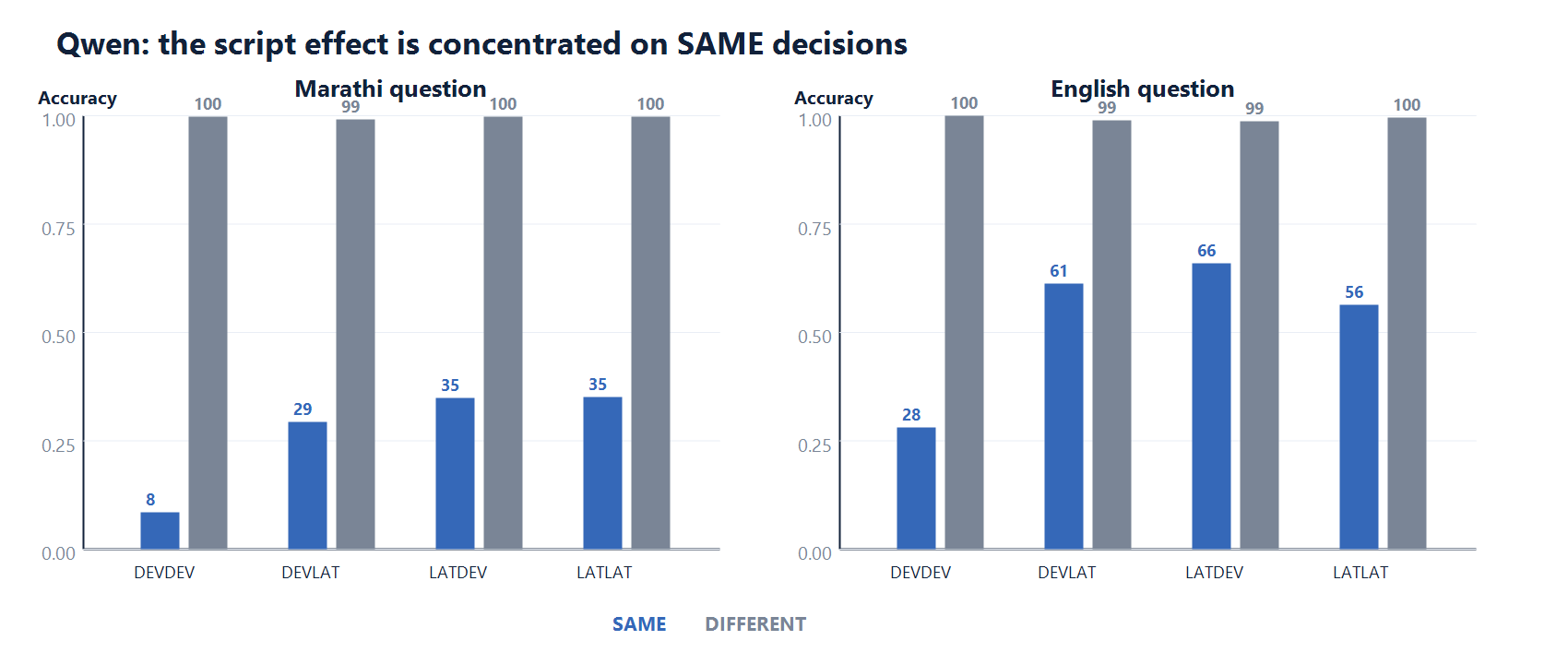}
  \caption{Qwen accuracy across the four ways of writing the two names.
  The large change is concentrated on SAME pairs; DIFFERENT accuracy remains near
  ceiling. Values pool all 601 items, three paraphrases and both answer-key orders.}
  \label{fig:behavior}
\end{figure*}

Figure~\ref{fig:behavior} shows the shape of the effect in Qwen: the change falls
almost entirely on SAME pairs, while DIFFERENT accuracy stays near ceiling in every
script combination.

Table~\ref{tab:para} gives Qwen's positive-pair accuracy by paraphrase, and
Table~\ref{tab:tierbeh} gives it by relation tier. The gap is present in every paraphrase and every tier with more than one item,
so it spans multiple wordings and relation types.

\begin{table*}[tp]
\centering\small
\caption{Qwen positive-pair accuracy (\%) by paraphrase. Negatives, omitted, remain
between 97.2 and 100.0 in every cell.}
\label{tab:para}
\begin{tabular}{llcccc}
\toprule
Prompt & Paraphrase & Dev--Dev & Dev--Lat & Lat--Dev & Lat--Lat\\
\midrule
Marathi & P0 & 10.6 & 37.7 & 51.0 & 43.9\\
        & P1 & 11.6 & 46.9 & 51.2 & 51.7\\
        & P2 & 3.2  & 3.5  & 2.7  & 9.8\\
Hindi   & P0 & 14.8 & 61.8 & 73.4 & 50.0\\
        & P1 & 28.2 & 71.6 & 79.7 & 59.2\\
        & P2 & 6.2  & 2.2  & 3.5  & 14.1\\
English & P0 & 35.0 & 69.0 & 84.1 & 60.3\\
        & P1 & 34.6 & 71.1 & 87.1 & 65.4\\
        & P2 & 14.9 & 43.8 & 26.9 & 43.2\\
\bottomrule
\end{tabular}
\end{table*}

Paraphrase P2, which asks whether one name is another name of the second rather than
whether the two denote one person, is markedly harder in every condition. The script
gap survives in all three, but absolute accuracies differ enough that reporting a
single paraphrase would misstate the level. All figures in the main text are pooled
over the three.

\begin{table*}[tp]
\centering\small
\caption{Qwen positive accuracy (\%) by relation tier: Marathi prompts on the Marathi
corpus in the upper block, Hindi prompts on the Hindi corpus in the lower.}
\label{tab:tierbeh}
\begin{tabular}{llccccr}
\toprule
Prompt & Tier & Dev--Dev & Dev--Lat & Lat--Dev & Lat--Lat & $n$\\
\midrule
Marathi & Alias observed                  & 10.4 & 27.7 & 36.3 & 32.5 & 404\\
        & Explicit plus alias, birth name & 4.3  & 34.8 & 33.0 & 43.0 & 171\\
        & Explicit plus alias, pseudonym  & 0.0  & 9.0  & 16.7 & 17.9 & 13\\
        & Explicit plus alias, nickname   & 14.3 & 33.3 & 42.9 & 45.2 & 7\\
        & Explicit observed, birth name   & 13.3 & 33.3 & 33.3 & 16.7 & 5\\
        & Explicit observed, nickname     & 0.0  & 0.0  & 0.0  & 0.0  & 1\\
\midrule
Hindi   & Alias observed                  & 19.6 & 45.8 & 56.2 & 39.7 & 404\\
        & Explicit plus alias, birth name & 9.6  & 46.4 & 46.3 & 45.6 & 171\\
        & Explicit plus alias, pseudonym  & 10.3 & 20.5 & 19.2 & 24.4 & 13\\
        & Explicit plus alias, nickname   & 14.3 & 38.1 & 47.6 & 57.1 & 7\\
        & Explicit observed, birth name   & 13.3 & 40.0 & 33.3 & 30.0 & 5\\
        & Explicit observed, nickname     & 0.0  & 0.0  & 0.0  & 0.0  & 1\\
\bottomrule
\end{tabular}
\end{table*}

Pseudonyms are hardest and plain observed aliases easiest, and Dev--Dev is the lowest
condition in every tier of appreciable size. The ordering of the other three conditions
is not preserved across tiers: Lat--Dev exceeds Lat--Lat for observed aliases, 36.3
against 32.5 over 404 items, and falls below it for birth names, 33.0 against 43.0 over
171.

\textbf{Origin split.} Items are labelled by whether the entity is of Indic or foreign
origin. Under Marathi prompts both start at essentially the same Devanagari accuracy,
8.3 percent for the 489 foreign-origin entities and 9.4 percent for the 112
Indic-origin ones, and both
gain from Latin script, to 35.9 and 32.0 percent respectively. Indic names, which
should be best attested in Devanagari, receive no advantage there. We rest the argument on the
equality at Devanagari rather than on foreign entities gaining more, since the
differential gain, 27.6 against 22.6 points, is small. This is difficult
to reconcile with an explanation based on the volume of Devanagari text seen during
training.

\section{Sensitivity measure}
\label{app:dprime}

Because the three primary models differ in response bias, raw accuracy on either polarity
confounds knowledge with decision threshold. We therefore report
$d' = z(\text{hit rate}) - z(\text{false alarm rate})$. For each model, prompt
language and script condition, responses are pooled across items, paraphrases and
counterbalanced answer keys before computing the hit and false-alarm rates. Every pooled hit and false-alarm rate uses the log-linear estimator
$(x + 0.5)/(n + 1)$ before the $z$ transform, not only rates of zero or one.
Confidence intervals come from 2000 cluster-bootstrap resamples: items are sampled
with replacement and the pooled rates and $d'$ are recomputed within each resample.
Items, rather than individual prompts, are resampled because prompts derived from one
item are not independent. Table~\ref{tab:dprime} reports the resulting gaps.

\begin{table*}[tp]
\centering\small
\caption{Lat--Lat minus Dev--Dev $d'$ gap, 95\% bootstrap CI over the 601 items of
each corpus. Marathi and English rows are from the Marathi run, Hindi from the Hindi
run. Llama's Marathi and Hindi arms ($\dagger$) are excluded from the analysis as
instruction-following failures (Appendix~\ref{app:llama}); their intervals exclude zero
but they are not among the cells interpreted in \S\ref{sec:behaviour}.}
\label{tab:dprime}
\begin{tabular}{llccc}
\toprule
Model & Prompt & Dev--Dev & Lat--Lat & Gap (CI)\\
\midrule
Qwen  & Marathi & 1.494 & 2.376 & $+0.88\ [0.14, 1.55]$\\
      & Hindi & 2.122 & 2.425 & $+0.30\ [-0.22, 0.76]$\\
      & English & 2.765 & 2.833 & $+0.07\ [-0.45, 0.74]$\\
Gemma & Marathi & 0.724 & 1.410 & $+0.69\ [0.57, 0.80]$\\
      & Hindi & 1.428 & 1.910 & $+0.48\ [0.33, 0.65]$\\
      & English & 1.136 & 1.866 & $+0.73\ [0.56, 0.91]$\\
Llama & Marathi$^{\dagger}$ & 0.020 & 0.103 & $+0.08\ [0.06, 0.10]$\\
      & Hindi$^{\dagger}$ & 0.108 & 0.506 & $+0.40\ [0.35, 0.44]$\\
      & English & 0.994 & 1.993 & $+1.00\ [0.90, 1.11]$\\
\bottomrule
\end{tabular}
\end{table*}

Gemma additionally shows a monotone dose response in English as the number of
Latin-script entities increases: 1.14 with neither, 1.44 and 1.53 with one, 1.87
with both, the extremes being the Dev--Dev and Lat--Lat entries above. Qwen's ordering is not monotone.

The two interpretable cells that do not exclude zero are Qwen's English and Hindi arms.
The reason is a floor on its false-alarm rate rather than an absence of sensitivity: one
false alarm in 3606 Dev--Dev English trials, against fourteen in the same number of
Lat--Lat trials, leaves the estimate dominated by the $(x + 0.5)/(n + 1)$ correction
applied to such rates rather than by the rates themselves. The AUC of the yes-minus-no
margin, which needs no such correction, separates the same conditions in every prompt
language (\S\ref{sec:behaviour}).

\section{Fragmentation analyses}
\label{app:frag}

We report every fragmentation analysis we ran, including the two that partially favour
the tokenizer account. Except where a corpus is named, all figures in this appendix are
from the Qwen run on the Marathi corpus.

\textbf{Segmentation by model and corpus.} Devanagari names are cut into 16.5 tokens for Qwen, 10.1 for Llama and 9.3 for Gemma on the Marathi corpus, or 0.98, 0.61 and 0.56 tokens per character. The Hindi corpus gives 16.4, 9.8 and 9.0, or 0.99, 0.60 and 0.55, so the ordering and the near two-fold spread are properties of the tokenizers rather
than of one orthography.

\textbf{Recognition does depend on length.} Feeding each Devanagari name in isolation
and asking whether its Latin counterpart is the nearest of the 601 candidates at
layer 8:

\begin{table*}[tp]
\centering\small
\caption{Isolated-name retrieval by Devanagari token count, all 1{,}202 names, Qwen at
layer 8.}
\label{tab:reclen}
\begin{tabular}{llrrr}
\toprule
Corpus & Token count & Names & Mean tokens & Retrieved at rank 1\\
\midrule
Marathi & $\leq 10$ & 217 & 8.4  & \textbf{41.0\%}\\
        & 11--15    & 396 & 13.0 & 22.0\%\\
        & 16--20    & 303 & 17.8 & 17.5\%\\
        & $\geq 21$ & 286 & 26.2 & \textbf{16.8\%}\\
\midrule
Hindi   & $\leq 10$ & 220 & 8.4  & \textbf{42.7\%}\\
        & 11--15    & 400 & 13.0 & 24.0\%\\
        & 16--20    & 308 & 17.9 & 22.4\%\\
        & $\geq 21$ & 274 & 26.3 & \textbf{21.5\%}\\
\bottomrule
\end{tabular}
\end{table*}

Length matters for recognition. It does not determine it: among 159 names of 11 to 12
tokens, 36 are retrieved and 123 are not.

\textbf{Fertility on its own predicts nothing.} The Devanagari-to-Latin fertility gap
does not predict the behavioural gap at the item level: Spearman $\rho = 0.024$ ($p = 0.56$) under English prompts and $0.062$ ($p = 0.12$) under Marathi prompts. Holding fertility fixed also barely moves the surprisal correlations, which change
by less than 0.05 in every cell. Items the model mostly answers correctly are only
slightly less surprising than items it mostly fails, 2.74 against 3.16 per character under Marathi prompts.

\textbf{Fragmentation predicts the benefit of Latin script, weakly.}
Table~\ref{tab:fragcorr} correlates the Devanagari-to-Latin token ratio against the
margin gained by switching that item to Latin. The pooled relationship is significant
and small, $\rho = +0.146$ over 601 items. It is not merely a between-group artefact of
foreign names being both more fragmented and more helped: within the 489 foreign-origin
names it is $+0.168$ and significant, and within the 112 native-origin names $+0.091$
and not, although that subgroup is small enough that the two are not distinguishable
from one another. We therefore do not claim that the within-group relationship is
absent. It is present and weak.

\begin{table*}[tp]
\centering\small
\caption{Fragmentation correlations, Spearman $\rho$. The last row of each block uses
the first layer at which the correct counterpart reaches rank 1, and is necessarily
restricted to names that ever do so. The origin split is a property of the entity, taken
from the Marathi corpus and applied to the Hindi items by \texttt{fact\_id}; the two
corpora cover the same 601 entities.}
\label{tab:fragcorr}
\begin{tabular}{lccc}
\toprule
Relationship & $\rho$ & $p$ & $n$\\
\midrule
\multicolumn{4}{l}{\emph{Marathi}}\\
Token ratio vs Latin-script benefit, pooled & $+0.146$ & $3.3\times10^{-4}$ & 601\\
\quad foreign-origin names                  & $+0.168$ & $1.9\times10^{-4}$ & 489\\
\quad native-origin names                   & $+0.091$ & $0.34$             & 112\\
Token ratio vs layer of first recognition   & $+0.220$ & $4.6\times10^{-8}$ & 604 of 1202\\
\midrule
\multicolumn{4}{l}{\emph{Hindi}}\\
Token ratio vs Latin-script benefit, pooled & $+0.122$ & $2.7\times10^{-3}$ & 601\\
\quad foreign-origin names                  & $+0.116$ & $1.0\times10^{-2}$ & 489\\
\quad native-origin names                   & $+0.077$ & $0.42$             & 112\\
Token ratio vs layer of first recognition   & $+0.292$ & $8.0\times10^{-15}$ & 680 of 1202\\
\bottomrule
\end{tabular}
\end{table*}

\begin{table*}[tp]
\centering\small
\caption{Mean Latin-script benefit in margin points by quartile of fragmentation,
Qwen, 601 items. Each corpus is read under its own prompt language.}
\label{tab:fragquart}
\begin{tabular}{llcc}
\toprule
Corpus & Quartile & Mean token ratio & Mean benefit\\
\midrule
Marathi & Q1, least fragmented & 2.12 & 3.78\\
        & Q2 & 2.60 & 3.82\\
        & Q3 & 3.16 & 5.74\\
        & Q4, most fragmented & 4.49 & \textbf{6.34}\\
\midrule
Hindi   & Q1, least fragmented & 2.12 & 3.29\\
        & Q2 & 2.60 & 3.43\\
        & Q3 & 3.12 & 4.48\\
        & Q4, most fragmented & 4.41 & \textbf{5.50}\\
\bottomrule
\end{tabular}
\end{table*}

Table~\ref{tab:fragquart} shows the same relationship as a dose-response, and on this
corpus it is monotone: mean benefit rises from 3.78 to 6.34 across quartiles spanning a
token ratio of 2.12 to 4.49. Fragmentation is a genuine covariate of the script effect,
not a spurious one, and we say so rather than resting the argument on a null that this
corpus does not support.

What fragmentation is not is a sufficient explanation, and the three results the main
text relies on do not depend on these correlations. Reducing fragmentation by automatic
romanization gives no reliable recovery and lowers performance in most cells, while the
human Latin form of
the same name makes it better (\S\ref{sec:tokenizer}). Two models with roughly half the
Devanagari fragmentation of the third show gaps of comparable size. And out-of-fold
prediction of which items are script-sensitive reaches 0.56 to 0.60 AUC from surface
features against 0.86 to 0.96 from mid-network relation states
(Table~\ref{tab:auc}): a per-item predictor built from token ratios is close to useless
even though the aggregate correlation is real. A covariate that explains this little per item does not account for the effect on
its own. Token ratios are an imperfect proxy and downstream representations can
carry tokenizer effects a surface count cannot see, so the defensible claim is the
narrower one: the reductions in fragmentation we tested did not restore performance.

\begin{table*}[tp]
\centering
\small
\caption{Out-of-fold AUC for predicting which items are script-sensitive. The two
Hindi columns, right of the rule, are the same measurement on Hindi renderings of the
same entities. The columns rest on unequal evidence: the number of script-sensitive
items is 201, 180, 63, 199 and 33 respectively, so Gemma's two columns are the least
well supported.}
\label{tab:auc}
\begin{tabular}{lccc|cc}
\toprule
Feature set & Qwen (mr) & Qwen (en) & Gemma (mr) & Qwen (hi) & Gemma (hi)\\
\midrule
Surface features only        & 0.602 & 0.564 & 0.588 & 0.572 & 0.561\\
{}+ early-layer retrieval    & 0.590 & 0.568 & 0.557 & 0.574 & 0.626\\
{}+ retrieval across layers  & 0.694 & 0.728 & 0.672 & 0.722 & 0.653\\
{}+ relation states          & \textbf{0.954} & \textbf{0.962} & \textbf{0.864} & \textbf{0.937} & \textbf{0.915}\\
\bottomrule
\end{tabular}
\end{table*}

\section{Transliteration}
\label{app:translit}

\textbf{How close the automatic schemes get.} Across the five schemes, automatic
romanizations reproduce the human Latin field exactly for 0.8 to 4.5 percent of names and differ from it by a mean normalised edit distance of 0.30 to 0.36
(Table~\ref{tab:translit}). Merely moving a name into Latin characters is therefore not the same as
supplying the attested Latin form.

Five romanization schemes are run: \texttt{itrans}, \texttt{iast}, \texttt{iast\_ascii}
and \texttt{hk} from \texttt{indic\_transliteration}, and \texttt{translit\_ascii} from
\texttt{unidecode}. Each is applied in two arms. The \emph{replaced} arm substitutes the
romanization for the native name, which is the fragmentation test; the \emph{anchored}
arm writes it in brackets beside the native name, which is the deployable fix. The
corpus's own Latin field is carried through both arms as a control: substituting it
reproduces the Lat--Lat condition to within 0.004 in all nine model-by-language
cells,
which is what establishes that the substitution mechanism works and that the automatic
schemes fail on their content rather than on their construction.

\begin{table*}[tp]
\centering\small
\caption{Romanization on held-out test items, against the Dev--Dev baseline and the Lat--Lat reference. Ranges span the five automatic schemes. Replacing the native name with
any of them does not reliably exceed the baseline; anchoring beside it helps but does not reach the reference. English rows are from the Marathi run; the Hindi corpus is a Hindi rendering of the same 601 entities.}
\label{tab:romanization}
\begin{tabular}{llcccc}
\toprule
Model & Prompt & Dev--Dev & replaced & anchored & Lat--Lat\\
\midrule
Qwen2.5-7B & Marathi & 0.537 & 0.523--0.548 & 0.535--0.554 & 0.706\\
           & Hindi & 0.546 & 0.508--0.525 & 0.550--0.562 & 0.748\\
           & English & 0.667 & 0.550--0.579 & 0.671--0.733 & 0.781\\
Gemma-2-2b & Marathi & 0.754 & 0.621--0.683 & 0.675--0.731 & 0.871\\
           & Hindi & 0.829 & 0.646--0.702 & 0.681--0.721 & 0.827\\
           & English & 0.529 & 0.510--0.548 & 0.533--0.615 & 0.646\\
Llama-3.1-8B & Marathi & 0.515 & 0.502--0.510 & 0.521--0.544 & 0.531\\
             & Hindi & 0.510 & 0.500--0.502 & 0.535--0.575 & 0.562\\
             & English & 0.702 & 0.535--0.617 & 0.610--0.662 & 0.850\\
\bottomrule
\end{tabular}
\end{table*}

Fragmentation is reduced but not removed by the substitution. The native names average
16.5 tokens; the romanizations average between 7.7 and 10.4, against 5.7 for the human
Latin form. So the schemes recover between half and four fifths of the token-count gap and none
of the performance, while the human form recovers all of both. Reduced fragmentation is
therefore not sufficient, and we report the manipulation as changing surface familiarity
as well as segmentation rather than as isolating segmentation alone.

\begin{table*}[tp]
\centering\small
\caption{Transliteration outputs against canonical Latin spellings.}
\label{tab:translit}
\begin{tabular}{lll}
\toprule
Version 1 & Version 2 & Canonical\\
\midrule
\texttt{imdira gamdhi}   & \texttt{imdira gamdhi}   & Indira Gandhi\\
\texttt{jaoNrja iliyata} & \texttt{jorja iliyata}   & George Eliot\\
\texttt{meri eNna ivhansa} & \texttt{meri aena ivhansa} & Mary Anne Evans\\
\texttt{naremdra modi}   & \texttt{naremdra modi}   & Narendra Modi\\
\bottomrule
\end{tabular}
\end{table*}

Version 2 repairs diacritic artefacts but not the two substantive defects, both
inherited from IAST: the inherent vowel is written even where it is unpronounced
(\emph{iliyata} for \emph{Eliot}), and the anusvara is rendered \emph{m} rather than
\emph{n} (\emph{gamdhi} for \emph{Gandhi}). Behavioural accuracy is accordingly unchanged between versions, 3.8 against 3.4
percent for foreign-origin names.

We note explicitly that these strings are close to, not far from, their canonical
forms: mean string similarity is 0.82 for Indic-origin names, with individual values
such as 0.93 for \emph{mahatma gamdhi} and 0.92 for \emph{naremdra modi}. Strings
roughly ninety percent identical to a working form score four times worse than it.
This strengthens rather than weakens the canonical-form reading, but it does mean the
transliteration condition varies spelling fidelity as well as script. A condition
using correct but non-canonical Latin spellings would isolate the two, and we did not
run one.

\section{Probing}
\label{app:probe}

\textbf{Grouping.} Grouping is five-fold by target item, so no target fact appears in
both fitting and evaluation. The split is not entity-disjoint with respect to the globally
selected distractors, which are drawn from one pool across folds, and the extent is worth
stating rather than only flagging. Of the 120 test entities, 100 have a distractor taken
from another split and 76 of those from the training split; the corresponding count for
the validation split is 94 of 120. A distractor's names are therefore often names the
probe and the correction vector have seen, as part of a different item.

What that can affect is bounded by where the distractor appears. It is present only in the
DIFFERENT-pair prompt of an item, never in the SAME-pair prompt, so prior exposure could
only help the model or a probe reject negatives. It does not: on test items whose
distractor came from the training split, Dev--Dev negative-pair accuracy is 99.1 percent
under Marathi prompts and 100.0 under English, against 100.0 and 100.0 on items whose
distractor was also held out. Positive-pair accuracy does differ between those groups, 8.3
against 3.4 percent in Marathi, but the distractor is absent from positive prompts
altogether, so that difference is a property of which items the similarity-based matching
gave a cross-split distractor to and not an effect of exposure. We therefore report the
non-disjointness as a corpus-construction fact with no measurable consequence for the
class it could touch, rather than as a resolved threat.

\textbf{Failure recall under English prompts.} Restricted to items the model answers
incorrectly and averaged over layers 10 to 18, the probe recovers the correct decision
on 86.4 percent of English failures at the question mark and 83.3 percent at the final readout. The corresponding Marathi and Hindi figures are in Table~\ref{tab:probepos}.

\textbf{Gemma's delayed direction.} Fitting on Marathi states and evaluating on English
states is at 0.500 at layer 10, but the validation score rises to 0.867 at layer 20 and the held-out test score there is 0.838. Cross-aligned probe information therefore generalises in every direction we
tested, although its most readable layer can shift.

\textbf{Isolated-name retrieval.} Ranking each Devanagari name against all 601 Latin
candidates without any surrounding context, Qwen's two name roles reach 0.511 and 0.456 top-1 retrieval at layer 3, the peak recognition layer. This is the cleaner identity measurement than the contextual one, because no
unchanged second name is available to support retrieval; the layer differs from the
layer 8 used in Table~\ref{tab:reclen}, which reports the same measurement by token
count.

Probes are ridge classifiers, $\alpha = 10$, fitted on L2-normalised hidden states.
Fitting always uses Lat--Lat states and evaluation always uses Dev--Dev states, so
every probe is cross-script by construction; no Devanagari state is seen during
fitting. Cross-validation is five-fold with grouping by \texttt{fact\_id}, so no target
fact appears in both fitting and evaluation. Two protocols appear in this paper and
produce different tables. The five-fold scheme described here yields the layer profiles
and the per-group breakdowns, which are descriptive. The headline held-out figures
instead use the fixed 361/120/120 split: the layer is chosen on validation and the
accuracy read on test at that layer, so no test item participates in the choice. Where a
number is called held-out it comes from the second protocol. The prompt language of the fitting and
evaluation states is crossed independently, giving four combinations, and probes are
run at both the question-mark and readout positions at every layer.

Both accuracy and per-class recall are reported because early layers are weak and
class-unbalanced rather than discriminating. At layer 0 the Marathi question-mark probe
reaches 0.525 balanced accuracy in Qwen and 0.491 in Gemma, with recall split unevenly
between the classes, 0.411 same against 0.639 different in Qwen and 0.691 against 0.291
in Gemma, so the direction of the imbalance is not the same in the two models. Only from
layer 4 onward is the probe discriminating in both.

\begin{table*}[tp]
\centering\small
\caption{Held-out balanced accuracy, 120 test items. The layer is chosen on the
validation split alone and the accuracy reported on the test split at that layer, so
no test item participates in the choice. The selected layer is given in parentheses.}
\label{tab:probeheld}
\begin{tabular}{llccc}
\toprule
Fit on & Evaluate on & Qwen & Gemma & Llama\\
\midrule
English & English & 0.950 (L18) & 0.921 (L11) & 0.942 (L11)\\
English & Marathi & 0.929 (L15) & 0.883 (L10) & 0.954 (L9)\\
\rowcolor{resultgreen}
Marathi & English & 0.942 (L21) & \textbf{0.838 (L20)} & 0.925 (L21)\\
Marathi & Marathi & 0.925 (L16) & 0.925 (L14) & 0.950 (L19)\\
\bottomrule
\end{tabular}
\end{table*}

\textbf{Layer profile.} Table~\ref{tab:probelayer} tracks the probe down the network.
Both models are weak and class-unbalanced through the early layers, in
opposite directions, then rise sharply and plateau through the middle. Both decline near the output,
and the decline is informative: in each model it appears almost entirely in the
\emph{same} class, as the next paragraph quantifies.

\begin{table*}[tp]
\centering\small
\caption{Cross-script probe by layer, Marathi prompts, question-mark position. Bal.\ is
balanced accuracy, S and D are recall on same-person and different-person items.}
\label{tab:probelayer}
\begin{tabular}{llccc|ccc}
\toprule
& & \multicolumn{3}{c|}{Qwen (28 layers)} & \multicolumn{3}{c}{Gemma (26 layers)}\\
\cmidrule(lr){3-5}\cmidrule(lr){6-8}
Prompt & Layer & Bal. & S & D & Bal. & S & D\\
\midrule
Marathi & 0    & 0.525 & 0.411 & 0.639 & 0.491 & 0.691 & 0.291\\
        & 4    & 0.832 & 0.915 & 0.749 & 0.665 & 0.927 & 0.403\\
        & 8    & 0.902 & 0.820 & 0.983 & \textbf{0.913} & 0.857 & 0.970\\
        & 12   & 0.871 & 0.742 & 1.000 & 0.910 & 0.834 & 0.987\\
        & 16   & \textbf{0.913} & 0.832 & 0.993 & 0.901 & 0.807 & 0.995\\
        & 20   & 0.899 & 0.804 & 0.993 & 0.859 & 0.724 & 0.995\\
        & 24   & 0.834 & 0.671 & 0.997 & 0.833 & 0.669 & 0.997\\
        & last & 0.809 & \textbf{0.621} & 0.997 & 0.712 & \textbf{0.428} & 0.997\\
\midrule
Hindi   & 0    & 0.507 & 0.121 & 0.894 & 0.507 & 0.622 & 0.391\\
        & 4    & 0.781 & 0.952 & 0.611 & 0.820 & 0.764 & 0.877\\
        & 8    & 0.932 & 0.887 & 0.977 & 0.622 & 0.246 & 0.998\\
        & 12   & 0.918 & 0.835 & 1.000 & 0.891 & 0.784 & 0.998\\
        & 16   & 0.930 & 0.865 & 0.995 & 0.911 & 0.834 & 0.988\\
        & 20   & 0.924 & 0.864 & 0.985 & 0.908 & 0.832 & 0.985\\
        & 24   & 0.880 & 0.860 & 0.900 & 0.874 & 0.772 & 0.977\\
        & last & 0.854 & 0.903 & 0.805 & 0.682 & 0.373 & 0.992\\
\bottomrule
\end{tabular}
\end{table*}

Both models' late-layer representations drift toward rejection: \emph{different} recall
reaches 0.997 in each while \emph{same} recall falls to 0.621 in Qwen and 0.428 in
Gemma. The drift is sharper in Gemma, whose behavioural bias runs the other way, so the
late-layer representation is not simply tracking the model's output preference. This is further evidence that the probe is not merely reading off the
model's impending decision, since a probe doing that would degrade in whichever
direction the model errs, which is what we observe only at the very end of the network
and not in the middle where the reported figures are taken.

\textbf{Group breakdown.} Restricting to failures and splitting by behavioural group
gives 0.789 \emph{same} recall on script-sensitive items, 0.944 on items robust across
scripts, 0.913 on other known items, and 0.662 on items the model never learned. The
last figure is above chance, so the probe is not a pure readout of stored knowledge.
What it does show is a monotone ordering, from items answered correctly everywhere down
to items never answered correctly, spanning 28 points. A probe with a blanket
\emph{same} bias would score high on all four groups alike.

\textbf{Probe by position.} No single position is privileged. On Marathi failures,
question-mark recall is 0.789 versus 0.570 at final readout, with the question mark ahead
throughout layers 10--18; in Hindi, readout leads in eight of nine layers. The
dissociation of \S\ref{sec:probe} therefore rests on probe-versus-model performance, not
on a positional pattern that holds across prompt languages.

\textbf{Contextual alignment by changed span.} We also change one entity's
script-associated surface form and retrieve its matching fact-level state among 601
candidates. At E2, alignment is higher when the first rather than second span changes:
0.97 versus 0.49 in Qwen; this ordering holds in all nine cells of the three primary
models, including Llama's two excluded arms, whose behavioural failure does not prevent
the alignment measurement, and narrows or reverses by the question mark (Table~\ref{tab:inprompt}).
Because the unchanged name can itself identify the fact after E2, this measures contextual
cross-script alignment, not definitive preservation or loss of the switched entity.
Isolated-name retrieval is the cleaner identity test; the replicated asymmetry is part of
what motivates the E2 causal tests of \S\ref{sec:patching}.

\begin{figure*}[tp]
  \centering
  \includegraphics[width=\linewidth]{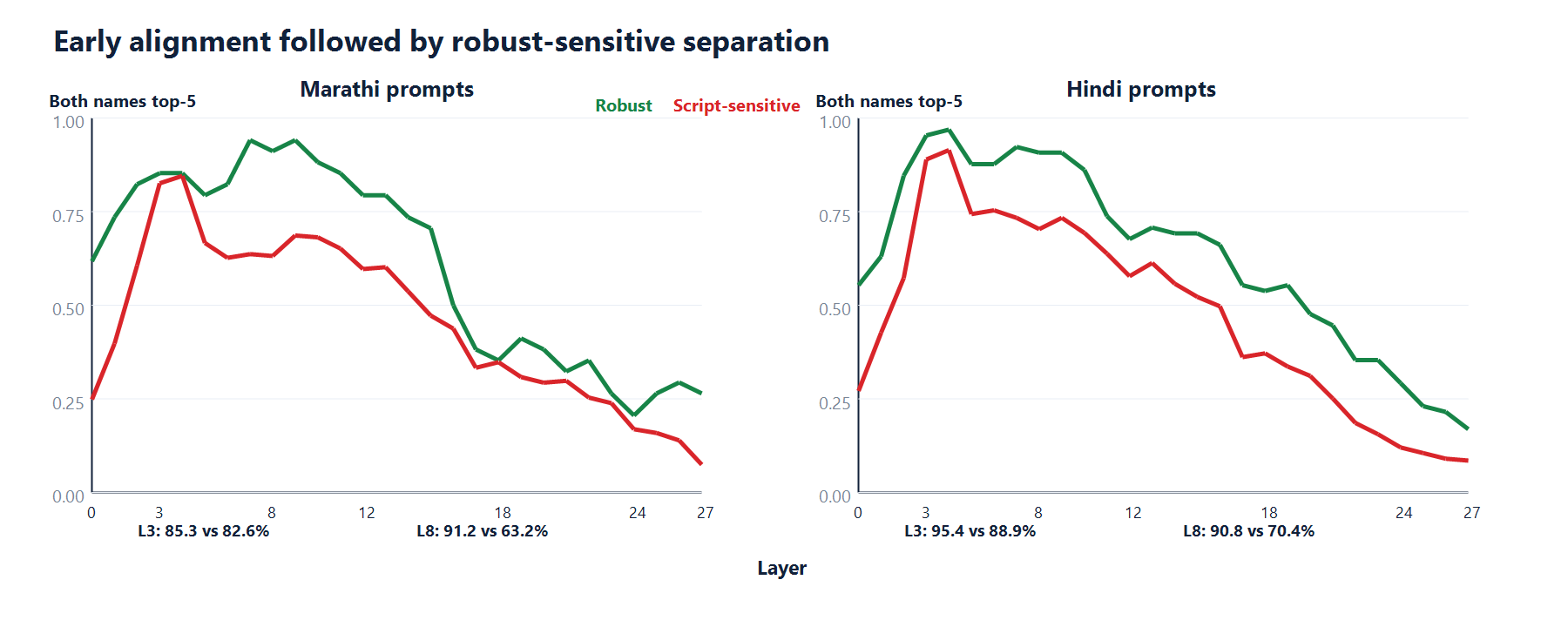}
  \caption{Qwen contextual alignment at E2 under Marathi prompts. Robust and
  script-sensitive items are close at layer 3 and separate by layer 8. Because E2
  contains the preceding name, this is a fact-context alignment measure rather than an
  isolated-name identity measure.}
  \label{fig:context-alignment}
\end{figure*}

\textbf{Contextual alignment trajectory.} At E2, items that are robust across scripts
and items that are script-sensitive show similar contextual cross-script alignment
early and diverge in the early-middle layers. The probability that both names reach
the top five differs by 2.7 points at layer 3 (85.3 versus 82.6 percent) and by 28.0
points at layer 8 (91.2 versus 63.2 percent). Because the E2 state also contains the
preceding name, this measures contextual cross-script alignment rather than either
name in isolation. The isolated-name analysis independently establishes early identity
alignment; together the measurements support a formation-then-contextual-loss
trajectory without assuming they are the same representation
(Figure~\ref{fig:context-alignment}).

\begin{table*}[tp]
\centering\small
\caption{Probe by position and by behavioural group, Qwen, Marathi and Hindi prompts.
Position figures are held-out test items at the layer chosen on validation alone; ties in
validation are broken toward the lower layer, and two of the four selections are exact
ties, Marathi's question mark with layer 17 and Hindi's with layer 15. Failure figures
are \emph{same} recall on exact failures, averaged over layers 10--18, and do not depend
on a single-layer selection. Group figures are out-of-fold over the same layers and are
Marathi only. All are from this run.}
\label{tab:probepos}
\begin{tabular}{lcc|cc}
\toprule
& \multicolumn{2}{c|}{Marathi} & \multicolumn{2}{c}{Hindi}\\
\cmidrule(lr){2-3}\cmidrule(lr){4-5}
& Question mark & Readout & Question mark & Readout\\
\midrule
Balanced accuracy, held-out test & \textbf{0.925} & 0.892 & \textbf{0.942} & 0.925\\
Selected layer & 16 & 6 & 8 & 17\\
\emph{Same} recall on failures & \textbf{0.789} & 0.570 & 0.829 & 0.927\\
\midrule
\multicolumn{5}{l}{\emph{Failure recall by behavioural group, Marathi, question-mark position}}\\
Robust across scripts & \multicolumn{4}{c}{0.944}\\
Other known items & \multicolumn{4}{c}{0.913}\\
Script-sensitive & \multicolumn{4}{c}{\textbf{0.789}}\\
Never learned & \multicolumn{4}{c}{\textbf{0.662}}\\
\bottomrule
\end{tabular}
\end{table*}

\begin{table*}[tp]
\centering\small
\caption{Top-1 contextual cross-script alignment when the first or second entity span is
switched, layer 8. The first-entity position cannot encode a later change to the second
entity under causal masking, so that cell is degenerate and is omitted. Because the
unchanged name can identify the fact at later positions, these are fact-context rather
than isolated-name identity scores. Llama's Marathi row is reported despite its behavioural arm being excluded. English
rows are from the Marathi run, as in Table~\ref{tab:behaviour}.}
\label{tab:inprompt}
\begin{tabular}{llcccccc}
\toprule
& & \multicolumn{2}{c}{Qwen} & \multicolumn{2}{c}{Gemma} & \multicolumn{2}{c}{Llama}\\
\cmidrule(lr){3-4}\cmidrule(lr){5-6}\cmidrule(lr){7-8}
Prompt & Switched span & at E2 & at Q & at E2 & at Q & at E2 & at Q\\
\midrule
Marathi & first  & \textbf{0.967} & 0.918 & \textbf{0.859} & 0.839 & \textbf{0.937} & 0.928\\
        & second & \textbf{0.491} & 0.899 & \textbf{0.394} & 0.615 & \textbf{0.547} & 0.842\\
Hindi   & first  & \textbf{0.938} & 0.708 & 0.788 & 0.793 & \textbf{0.932} & 0.923\\
        & second & \textbf{0.627} & 0.899 & \textbf{0.404} & 0.683 & \textbf{0.616} & 0.913\\
English & first  & \textbf{0.906} & 0.760 & \textbf{0.797} & 0.687 & \textbf{0.936} & 0.903\\
        & second & \textbf{0.474} & 0.851 & \textbf{0.343} & 0.685 & \textbf{0.527} & 0.875\\
\bottomrule
\end{tabular}
\end{table*}

\section{Full patching profiles}
\label{app:patch}

\begin{table*}[t]
\centering
\small
\caption{Mean margin change from patching, held-out Qwen test items; Hindi from the Hindi run.
E1 and E2 are the entities, Q the question mark, R the readout, and the window column
patches E1, E2 and Q jointly over layers 12--17. Site columns average layers 10--18 and the
three Latin-bearing donors, control and reverse columns all of 8--20, so the two are not
comparable; Appendix~\ref{app:patch} gives the matched contrast and the gate sizes.}
\label{tab:patch}
\begin{tabular}{lcccccccc}
\toprule
& \multicolumn{4}{c}{Site} & \multicolumn{3}{c}{Control} & Window\\
\cmidrule(lr){2-5}\cmidrule(lr){6-8}\cmidrule(lr){9-9}
Prompt & E1 & E2 & Q & R & wrong ent. & wrong pos. & reverse & 12--17\\
\midrule
\rowcolor{resultblue}
Marathi ($n_{\rm test}{=}12$) & 0.12 & \textbf{2.66} & 0.61 & 1.26 & 0.70 & 0.57 & $-2.79$ & \textbf{12.95}\\
\rowcolor{resultblue}
Hindi ($n_{\rm test}{=}15$) & $-0.03$ & \textbf{3.16} & 0.33 & 1.08 & 2.08 & 0.40 & $-1.44$ & \textbf{12.35}\\
\rowcolor{resultgold}
English ($n_{\rm test}{=}24$) & 0.27 & \textbf{2.85} & 0.79 & 1.33 & 0.86 & 0.37 & $-0.72$ & \textbf{11.65}\\
\bottomrule
\end{tabular}
\end{table*}

\begin{table*}[t]
\centering\small
\caption{Item-clustered intervals for the patching effects of Table~\ref{tab:patch}, 2000
resamples of entities. The upper block is the held-out split and is the estimate we rely
on; the lower adds the train and validation items patched under the same gate, which is
better powered but not independent of the layer choice. Appendix~\ref{app:patch}
gives the gate sizes, the entity-specificity contrast and why both blocks are reported.}
\label{tab:patchci}
\begin{tabular}{lccc}
\toprule
Quantity & Marathi & Hindi & English \\
\midrule
\multicolumn{4}{l}{\emph{Held-out test items} ($n=12$, $15$, $24$)} \\
First entity (E1) & 0.12 [$-$0.13, 0.38] & $-$0.03 [$-$0.25, 0.16] & 0.27 [0.02, 0.53] \\
Second entity (E2) & \textbf{2.66} [1.67, 3.71] & \textbf{3.16} [2.47, 3.90] & \textbf{2.85} [2.28, 3.45] \\
Question mark (Q) & 0.61 [0.33, 0.97] & 0.33 [0.16, 0.52] & 0.79 [0.59, 1.00] \\
Readout (R) & 1.26 [0.96, 1.58] & 1.08 [0.87, 1.30] & 1.33 [1.12, 1.57] \\
E2 minus wrong-entity donor & 2.09 [0.41, 3.84] & $-$0.51 [$-$2.06, 1.08] & 0.70 [$-$0.21, 1.54] \\
\midrule
\multicolumn{4}{l}{\emph{Every gated item patched} ($n=53$, $50$, $116$)} \\
First entity (E1) & 0.02 [$-$0.07, 0.10] & 0.08 [$-$0.02, 0.19] & 0.21 [0.12, 0.32] \\
Second entity (E2) & \textbf{2.42} [2.01, 2.84] & \textbf{2.89} [2.45, 3.31] & \textbf{2.53} [2.25, 2.82] \\
Question mark (Q) & 0.41 [0.30, 0.53] & 0.41 [0.32, 0.51] & 0.70 [0.62, 0.79] \\
Readout (R) & 0.96 [0.82, 1.10] & 1.14 [1.01, 1.26] & 1.23 [1.15, 1.32] \\
E2 minus wrong-entity donor & 1.23 [0.55, 1.94] & 0.15 [$-$0.56, 0.85] & 0.76 [0.33, 1.22] \\
\bottomrule
\end{tabular}
\end{table*}
\textbf{How to read Tables~\ref{tab:patch} and~\ref{tab:patchci}.} The site columns of
Table~\ref{tab:patch} average layers 10--18 and the three Latin-bearing donor conditions,
while the control and reverse columns average all tested layers, 8--20, over a different
donor set. The two are therefore not directly comparable, and the matched contrast is the
paired entity-specificity row of Table~\ref{tab:patchci}, restricted on both sides to the
Lat--Lat donor, the same layers and the entities present in both arms, because the
wrong-entity control exists only for that donor. Total gate sizes are 53 Marathi, 50 Hindi
and 116 English items, of which 12, 15 and 24 fall in the held-out split.

Table~\ref{tab:patchci} reports two blocks and we rely on the upper one. It is the held-out
test split and is independent of every analysis choice. The lower block adds the train and
validation items that were also patched: better powered, but not independent, because the
layer band was fixed on a pilot and those items informed it. We report both rather than
choosing, and read agreement between them as reassurance about the small held-out sample
rather than as a second result. All three lower blocks are complete gates. Capping the
English column at 60 items leaves the Marathi and Hindi
cells unchanged to the reported precision and widens the English intervals; the held-out
block is unaffected either way, because no such cap binds on the test split.

\begin{table*}[tp]
\centering\small
\caption{Mean margin change by layer and site, Lat--Lat source, pooled over splits.
E1 and E2 are the first and second entity, Q the question mark, R the readout
position. Gate sizes are $n{=}53$ Marathi and $n{=}116$ English for the
Marathi corpus, $n{=}50$ Hindi and $n{=}107$ English for the Hindi corpus.}
\label{tab:profile}
\begin{tabular}{llcccc|cccc}
\toprule
& & \multicolumn{4}{c|}{Native-script prompts} & \multicolumn{4}{c}{English prompts}\\
Corpus & Layer & E1 & E2 & Q & R & E1 & E2 & Q & R\\
\midrule
Marathi & 8  & 0.28 & 0.31 & 0.42 & 0.02 & 0.95 & $-0.31$ & 0.01    & 0.11\\
        & 9  & 0.14 & 0.53 & 0.47 & 0.05 & 0.93 & $-0.02$ & $-0.08$ & 0.10\\
        & 10 & 0.14 & 0.59 & 0.35 & 0.12 & 0.78 & $-0.25$ & 0.01    & 0.20\\
        & 11 & 0.12 & 0.79 & 0.38 & 0.09 & 0.60 & 0.52 & 0.08 & 0.19\\
        & 12 & 0.17 & 1.80 & 0.23 & 0.08 & 0.45 & 1.94 & 0.16 & 0.25\\
        & 13 & 0.09 & 3.19 & 0.19 & 0.20 & 0.47 & 2.11 & 0.36 & 0.31\\
        & 14 & $-0.07$ & 3.02 & 0.19 & 0.25 & 0.14 & 1.93 & 0.39 & 0.35\\
        & 15 & $-0.23$ & 4.82 & 0.38 & 0.23 & 0.14 & 3.17 & 0.91 & 0.48\\
        & 16 & $-0.21$ & \textbf{6.21} & 0.37 & 0.26 & 0.22 & \textbf{4.53} & 0.94 & 0.57\\
        & 17 & $-0.08$ & 5.84 & 0.53 & 0.23 & 0.27 & 3.90 & 1.40 & 0.84\\
        & 18 & $-0.01$ & 0.61 & \textbf{2.72} & 11.10 & $-0.01$ & 0.42 & \textbf{4.12} & 10.84\\
        & 19 & 0.06 & 0.17 & 2.07 & 15.25 & $-0.10$ & $-0.21$ & 3.00 & 12.78\\
        & 20 & 0.04 & 0.07 & 0.22 & 18.39 & $-0.08$ & $-0.09$ & 0.01 & 15.65\\
\midrule
Hindi   & 8  & 0.39 & $-0.11$ & 0.03 & $-0.03$ & 0.73 & $-0.19$ & $-0.07$ & 0.06\\
        & 9  & 0.25 & 0.23 & 0.02 & 0.02 & 0.78 & 0.31 & $-0.14$ & 0.04\\
        & 10 & 0.29 & 0.16 & 0.04 & 0.06 & 0.67 & $-0.04$ & $-0.00$ & 0.14\\
        & 11 & 0.14 & 0.35 & 0.15 & 0.01 & 0.44 & 0.82 & 0.04    & 0.12\\
        & 12 & 0.07 & 1.41 & 0.06 & $-0.01$ & 0.28 & 2.14 & 0.10 & 0.17\\
        & 13 & 0.07 & 2.24 & 0.17 & 0.04 & 0.39 & 2.14 & 0.32 & 0.18\\
        & 14 & 0.00 & 2.03 & 0.15 & 0.06 & 0.00    & 1.95 & 0.31 & 0.22\\
        & 15 & 0.15 & 4.12 & 0.17 & 0.09 & $-0.06$ & 3.15 & 0.92 & 0.33\\
        & 16 & 0.13 & 6.29 & 0.34 & 0.22 & 0.05 & \textbf{4.54} & 0.95 & 0.46\\
        & 17 & 0.19 & \textbf{6.41} & 0.41 & 0.44 & 0.12 & 3.90 & 1.35 & 0.75\\
        & 18 & 0.03 & 0.99 & \textbf{2.40} & 11.98 & $-0.04$ & 0.46 & \textbf{4.02} & 11.00\\
        & 19 & 0.01 & 0.08 & 2.05 & 14.99 & $-0.11$ & $-0.24$ & 2.89 & 13.05\\
        & 20 & 0.02 & $-0.02$ & $-0.01$ & 16.83 & $-0.10$ & $-0.12$ & $-0.02$ & 15.93\\\bottomrule
\end{tabular}
\end{table*}

Table~\ref{tab:profile} shows the handoff described in the main text. The second
entity peaks at layer 16 and collapses at 18; the question mark is negligible through
17 and rises exactly where the entity site fails. The readout column is reported for
completeness but is uninformative: patching the final position substitutes the
healthy run's decision directly and is close to an upper bound by construction.

Patching the first-entity site has little effect at every tested layer, including the
earliest. This rules out that site as a useful residual-stream intervention point in the
tested range; it does not prove that the first name was read correctly or that no
first-name error was carried downstream. Table~\ref{tab:windows} reports the
multi-layer windows and the joint-site combinations.

\begin{table*}[tp]
\centering\small
\caption{Multi-layer and joint-site patching at the second entity, Lat--Lat source,
pooled over splits. Values are mean margin change with flip rate in parentheses.}
\label{tab:windows}
\begin{tabular}{lcc}
\toprule
Patch & Marathi & English\\
\midrule
Layers 9--11   & 0.77 (0.00) & 0.04 (0.07)   \\
Layers 12--14  & 4.04 (0.04) & 3.10 (0.22)\\
Layers 15--17  & 9.08 (0.30) & 6.91 (0.50)\\
Layers 12--17  & \textbf{11.70 (0.43)} & \textbf{9.05 (0.64)}\\
Layers 18--20  & 0.56 (0.00) & 0.28 (0.00)\\
\midrule
E1+E2, 12--17    & 12.23 (0.51) & 9.99 (0.69) \\
E2+Q, 12--17     & 12.29 (0.49) & 10.96 (0.77)\\
E1+E2+Q, 12--17  & 12.80 (0.57) & 11.85 (0.77)\\\bottomrule
\end{tabular}
\end{table*}

Adding the first entity to a joint patch of the second contributes little over the
second alone, consistent with the single-site profile. Single-layer patches move the margin substantially but flip the decision rarely;
only sustained multi-layer patches flip at an appreciable rate, reaching 57 percent
in Marathi and 77 percent in English when all three sites are patched together. These
rates pool every gated item; the held-out rates of \S\ref{sec:patching}, 50 and 71
percent, are the same quantity on the test split alone.

\begin{table*}[tp]
\centering\small
\caption{Patching variants at E2 in Qwen, held-out test items, peak over layers 10--18.
Marathi and English columns are from the Marathi run, Hindi from the Hindi run.
Cross-language rows take the donor state from the other prompt language. The final block
applies the 12--17 window at E1 and E2 jointly to the matched negative items.
The spread across the three same-language donors is 1.94 margin points in Marathi,
2.43 in English and 2.61 in Hindi.}
\label{tab:donor}
\begin{tabular}{llccc}
\toprule
& Variant & Marathi & English & Hindi\\
\midrule
Donor, same language & Lat--Lat (used throughout) & \textbf{5.97} & 4.45 & \textbf{7.22}\\
                     & Lat--Dev & 5.88 & \textbf{6.88} & 4.61\\
                     & Dev--Lat & 4.03 & 5.15 & 5.66\\
\midrule
Donor, other language & Lat--Lat & 6.26 & 4.21 & 4.30\\
                      & Dev--Dev & 0.11 & 0.53 & 1.09\\
\midrule
Negative pairs & margin change, joint 12--17 & $-0.76$ & $-1.10$ & $-2.73$\\
               & accuracy after patching & 100\% & 100\% & 100\%\\
\bottomrule
\end{tabular}
\end{table*}

\begin{figure*}[tp]
  \centering
  \includegraphics[width=\linewidth]{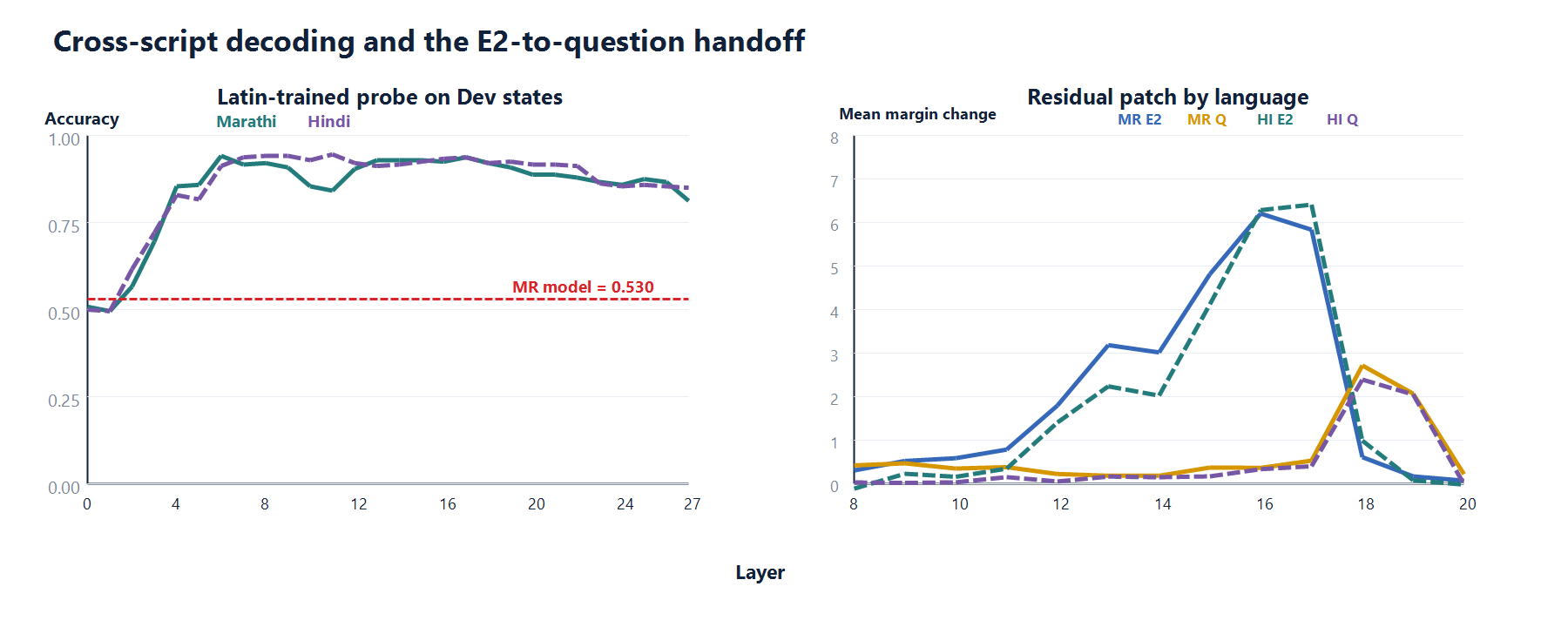}
  \caption{Two stages of the dissociation in Qwen. Left: a probe fitted on Lat--Lat
  states reads Dev--Dev states far above the model's own uncalibrated balanced accuracy, a gap Table~\ref{tab:calvsprobe} shows to be 87 percent a calibration difference. Right:
  patching at E2 becomes effective before the corrective effect moves to the question
  position. The patching profile is pooled over splits for localisation; held-out
  effect sizes and sample sizes are reported in Table~\ref{tab:patch}.}
  \label{fig:probe-patch}
\end{figure*}

\section{Components and heads}
\label{app:components}

At the second entity across layers 8 to 20, on held-out test items, patching attention
output alone gives 0.52 in Marathi and 0.29 in English and MLP output alone gives 0.07
and 0.48, against 2.16 and 1.31 for a single-layer residual patch. A single-layer
component patch imports only that layer's write, whereas a residual patch imports the
accumulated state, so those two are not on the same scale and the comparison bounds
rather than measures each component's share.

Patching a component cumulatively at every layer up to the current one removes that
asymmetry, and the ordering it gives is one-sided. Attention reaches 5.41 in Marathi
against 0.51 for the MLP, and 4.51 against 1.10 in English (Table~\ref{tab:components}). Gemma, read on the polarity
its gate uses, gives 1.68 against 1.13 and 1.60 against 0.80. Attention-component
patches therefore transfer more corrective signal in both models, strongly so in Qwen
and with weaker separation in Gemma. The transfer is distributed across layers rather
than localised in one: the single-layer figures above are near zero where the cumulative
ones are large.

Head screening on training items followed by confirmation on held-out items gives 1.70
for the single best head, 3.03 for the top three, 3.56 for the top five and 5.60 for the
top ten in Marathi, against 0.30 for the mean of the twelve selected heads patched
individually. No sparse mechanism is identified: the best head recovers about a third of
what the top ten recover jointly, so the effect is spread over several heads without
being uniform across them.

\begin{table*}[tp]
\centering\small
\caption{Component and head patching at E2 in Qwen. Marathi and English columns are
from the Marathi run, Hindi from the Hindi run. Single-layer components are
averaged over layers 8--20; cumulative components patch the same component at every layer up to
and including the current one. Heads are screened on training items over layers 8--20
and confirmed on validation and test items. ``Mean of 12'' averages the twelve selected
heads patched individually.}
\label{tab:components}
\begin{tabular}{lccc}
\toprule
Patched & Marathi & English & Hindi\\
\midrule
Full residual stream, single layer & 2.16 & 1.31 & 1.58\\
Attention output only, single layer & 0.52 & 0.29 & 0.41\\
MLP output only, single layer & 0.07 & 0.48 & 0.26\\
\midrule
Attention output, cumulative & \textbf{5.41} & \textbf{4.51} & \textbf{4.16}\\
MLP output, cumulative & 0.51 & 1.10 & 0.56\\
\midrule
Best single head & 1.70 & 1.44 & 1.66\\
Top 3 heads jointly & 3.03 & 2.53 & 2.99\\
Top 5 heads jointly & 3.56 & 3.01 & 4.22\\
Top 10 heads jointly & \textbf{5.60} & \textbf{4.56} & \textbf{6.94}\\
Mean of 12 selected heads, individually & 0.30 & 0.33 & 0.44\\
\bottomrule
\end{tabular}
\end{table*}

\section{Name-order swap}
\label{app:orderswap}

Qwen, both names exchanged between the two slots. The causal columns are the mean margin
change from patching that site over layers 12 to 17 on positive pairs, 20 gated facts per
prompt language; the behavioural columns are accuracy over all 601 facts. Marathi and
English are from the Marathi run, Hindi from the Hindi run. Table~\ref{tab:orderswap} reports
the causal and behavioural columns side by side.

\begin{table*}[tp]
\centering\small
\caption{Name-order swap in Qwen. Left: causal effect of patching each entity site,
canonical against swapped order. Right: behavioural accuracy, canonical against swapped.}
\label{tab:orderswap}
\begin{tabular}{lcccccccc}
\toprule
& \multicolumn{2}{c}{E2 patch} & \multicolumn{2}{c}{E1 patch}
& \multicolumn{2}{c}{Dev--Dev acc.} & \multicolumn{2}{c}{Lat--Lat acc.}\\
\cmidrule(lr){2-3}\cmidrule(lr){4-5}\cmidrule(lr){6-7}\cmidrule(lr){8-9}
Prompt & canon. & swap. & canon. & swap. & canon. & swap. & canon. & swap.\\
\midrule
Marathi & 4.79 & 3.78 & -0.11 & 0.03 & 0.551 & 0.572 & 0.719 & 0.751\\
Hindi & 5.78 & 4.45 & -0.08 & 0.41 & 0.572 & 0.632 & 0.746 & 0.814\\
English & 3.72 & 3.36 & -0.42 & 0.09 & 0.675 & 0.759 & 0.799 & 0.856\\
\bottomrule
\end{tabular}
\end{table*}

The causal effect tracks the slot rather than the string that occupies it, and the E1
site stays near zero in all six combinations. The behavioural gain from swapping appears in
Lat--Lat as well as in Dev--Dev, which is why we read it as an ordering effect and not as
evidence about script. The two are not ordered the same way in every arm: Lat--Lat gains
more under Marathi prompts, 0.032 against 0.021, and under Hindi, 0.068 against 0.060,
while under English prompts Dev--Dev gains more, 0.084 against 0.057. In $d'$ the same holds: the swap raises $d'$ in every
condition, by 0.39 and 0.16 under Marathi prompts and 0.43 and 0.60 under Hindi.

\section{Jacobian geometry}
\label{app:jacobian}

\begin{figure*}[tp]
  \centering
  \includegraphics[width=\linewidth]{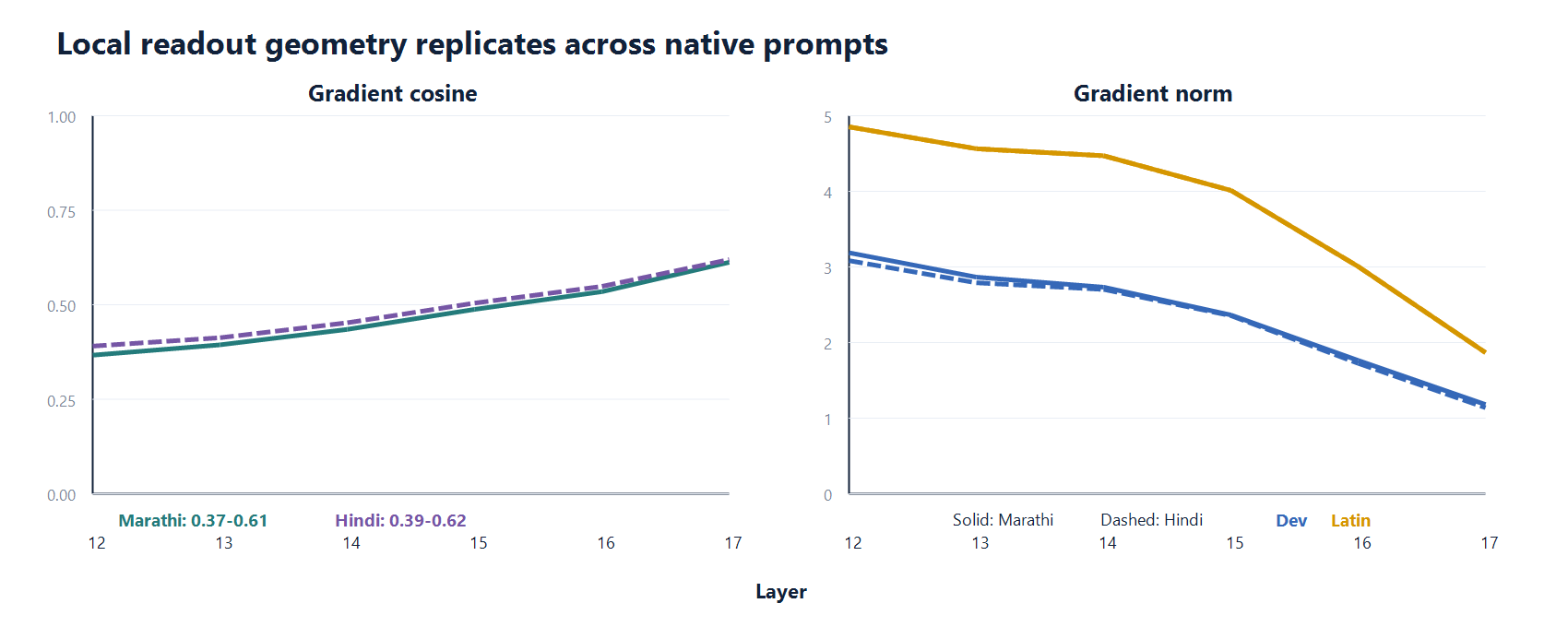}
  \caption{Local sensitivity at E2 in Qwen under Marathi prompts. Dev--Dev and
  Lat--Lat gradient directions overlap but do not coincide (left), and the Dev--Dev
  output is less sensitive at the same position (right). Values are means over 60
  pairs per layer.}
  \label{fig:jacobian}
\end{figure*}

Qwen at E2, 60 positive pairs per layer for the paired-frame columns and 80 for the
projection deficit. \emph{Deficit} is the Lat--Lat minus Dev--Dev projection onto the
Devanagari run's gradient. $\cos(g_{\mathrm{dev}},g_{\mathrm{lat}})$ compares the two runs'
gradient directions and $\cos(g_{\mathrm{dev}},r)$ a random direction in the same space.
The last two columns recompute the deficit with both states read in the Latin run's frame
and with each state read in its own run's frame.
Figure~\ref{fig:jacobian} shows the same two facts geometrically.

\begin{table*}[tp]
\centering\small
\caption{Local sensitivity at E2 in Qwen, layers 12 to 17. The deficit is positive and
grows with depth when both states are read through the Devanagari run's direction, and
neither grows nor stays positive once each state is read in its own frame.}
\label{tab:jacobian}
\begin{tabular}{lccccccc}
\toprule
Layer & deficit & $\cos(g_{\mathrm{dev}},g_{\mathrm{lat}})$ & $\cos(g_{\mathrm{dev}},r)$
& $\|g_{\mathrm{dev}}\|$ & $\|g_{\mathrm{lat}}\|$ & Latin frame & own frame\\
\midrule
\multicolumn{8}{l}{\emph{Marathi prompts}}\\
12 & 0.33 & 0.37 & +0.001 & 3.19 & 4.85 & $-$0.01 & $-$0.40\\
13 & 0.53 & 0.39 & +0.002 & 2.87 & 4.57 & $-$0.09 & $-$0.22\\
14 & 0.63 & 0.44 & +0.000 & 2.73 & 4.47 & 0.07 & 0.19\\
15 & 0.89 & 0.49 & $-$0.002 & 2.37 & 4.01 & 0.34 & $-$0.18\\
16 & 1.32 & 0.53 & +0.001 & 1.76 & 3.01 & 0.58 & $-$0.36\\
17 & 1.66 & 0.61 & +0.001 & 1.18 & 1.87 & 0.84 & $-$0.08\\
\midrule
\multicolumn{8}{l}{\emph{English prompts}}\\
12 & $-$0.02 & 0.47 & $-$0.003 & 2.95 & 4.03 & 0.00 & $-$0.76\\
13 & $-$0.12 & 0.46 & +0.004 & 2.53 & 3.59 & $-$0.11 & $-$0.73\\
14 & 0.18 & 0.48 & $-$0.001 & 2.36 & 3.36 & 0.15 & $-$0.46\\
15 & 0.46 & 0.48 & $-$0.003 & 1.97 & 2.82 & 0.48 & $-$0.77\\
16 & 0.71 & 0.52 & +0.000 & 1.57 & 2.06 & 0.96 & $-$0.58\\
17 & 0.72 & 0.59 & $-$0.001 & 1.06 & 1.25 & 1.07 & $-$0.45\\
\midrule
\multicolumn{8}{l}{\emph{Hindi prompts}}\\
12 & 0.00 & 0.43 & $-$0.001 & 2.86 & 4.12 & $-$0.03 & $-$0.10\\
13 & 0.21 & 0.46 & +0.001 & 2.70 & 4.02 & $-$0.15 & 0.02\\
14 & 0.45 & 0.47 & +0.000 & 2.63 & 4.00 & $-$0.03 & 0.06\\
15 & 0.99 & 0.51 & +0.004 & 2.33 & 3.48 & 0.05 & $-$0.18\\
16 & 1.19 & 0.57 & +0.003 & 1.85 & 2.69 & 0.45 & $-$0.55\\
17 & 1.28 & 0.63 & $-$0.002 & 1.32 & 1.78 & 0.87 & $-$0.33\\
\bottomrule
\end{tabular}
\end{table*}

The Devanagari gradient is smaller than the Latin one at every layer in all three
prompt languages, and the two directions grow more alike with depth while never approaching
each other closely. The random-direction column is the scale against which the observed
cosines should be read.

\section{Recovery details}
\label{app:recovery}

\begin{figure*}[tp]
  \centering
  \includegraphics[width=\linewidth]{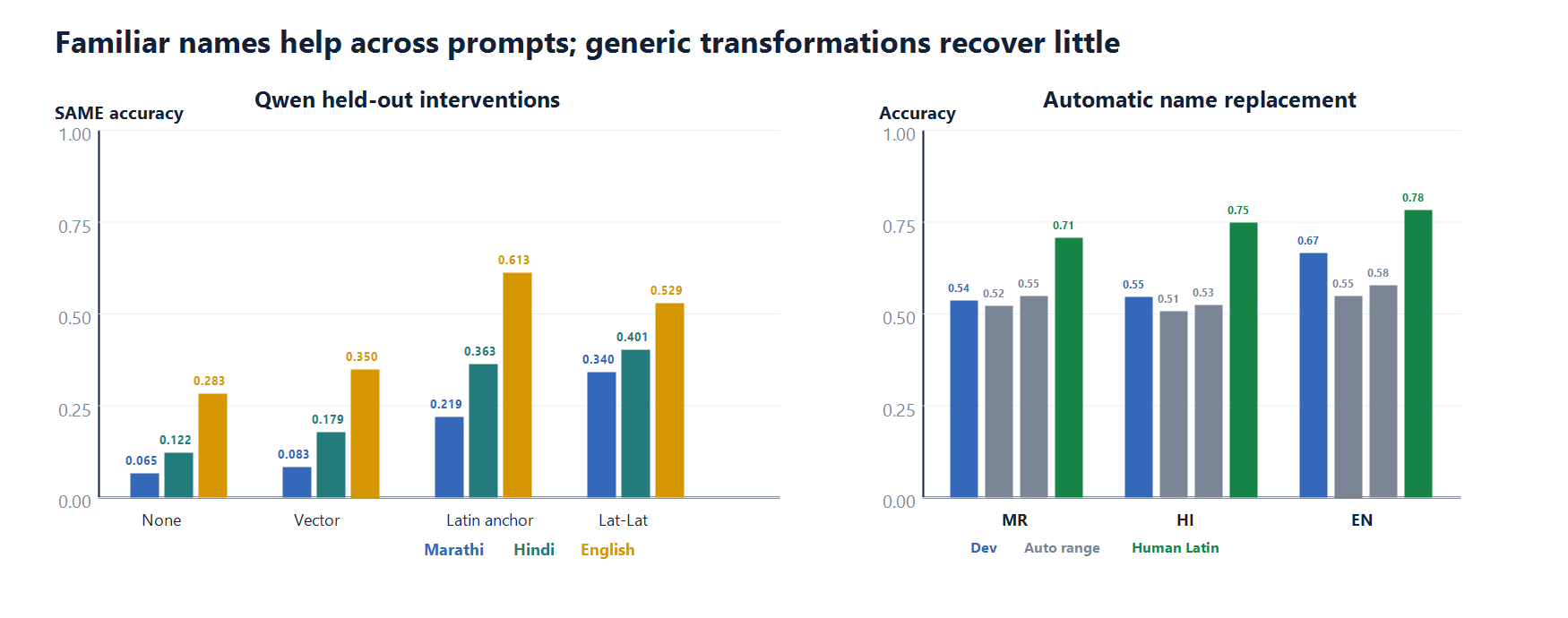}
  \caption{Held-out Qwen recovery controls, all of them single-layer. Added at one
  layer a universal additive vector yields a small gain, while supplying familiar Latin
  names helps substantially (left). Automatic romanizations shorten the strings but
  remain below the familiar human Latin form (right). The small gain here is a property
  of the single-layer setting: Appendix~\ref{app:window} gives the same vector across
  layers 12--17, where it moves accuracy substantially, and \S\ref{sec:repair} says
  what that movement is.}
  \label{fig:recovery}
\end{figure*}

Figure~\ref{fig:recovery} summarises the controls: added at a single layer, a universal
additive vector gives a small gain where supplying the familiar Latin form gives a large
one. Appendix~\ref{app:window} shows what the same vector does across the 12--17 band.

The correction vector is a mean difference of hidden states at the second entity over the
361 training items. Writing $h^{c}_i$ for the state of item $i$ at that position under
condition $c$, and $\mathcal{T}$ for the training items, the four variants are

\begin{align*}
v_{\text{full}} &= \tfrac{1}{|\mathcal{T}|}\textstyle\sum_{i \in \mathcal{T}}
  \left(h^{\text{LL}}_i - h^{\text{DD}}_i\right),\\
v_{\text{first}} &= \tfrac{1}{|\mathcal{T}|}\textstyle\sum_{i \in \mathcal{T}}
  \left(h^{\text{LD}}_i - h^{\text{DD}}_i\right),\\
v_{\text{second}} &= \tfrac{1}{|\mathcal{T}|}\textstyle\sum_{i \in \mathcal{T}}
  \left(h^{\text{DL}}_i - h^{\text{DD}}_i\right),\\
v_{\text{additive}} &= v_{\text{first}} + v_{\text{second}},
\end{align*}

where LL, DD, LD and DL are Lat--Lat, Dev--Dev, Lat--Dev and Dev--Lat. The normalisation
order matters and is stated here because it is not recoverable from the numbers: all four
are constructed first and each is then rescaled to unit length, so
$v_{\text{additive}}$ is the normalised sum of the two raw single-swap vectors and not the
sum of the two normalised ones. Scale therefore carries all magnitude in the sweep, and a
comparison across variants is a comparison of directions. One vector is built per layer,
shared across entities, and the window arm applies the vector of each layer at that layer.

Layer and scale were selected on validation items over layers $\{10,12,14,16,18\}$ and
scales $\{0.5,1.0,1.5,2.0,3.0,4.0,6.0,8.0\}$, maximising balanced accuracy subject to
negative-pair accuracy used only to break ties, and the smaller scale preferred on a
further tie. There is no floor on negative-pair accuracy in the rule, which is why the
selected settings are free to lower it and do.
Selection returned layer 16 at $\alpha{=}6.0$ under Marathi prompts and layer 12 at
$\alpha{=}8.0$ under English prompts. Table~\ref{tab:grid} reports four of the eight
scales; the omitted ones lie between their neighbours in every cell.

\begin{table*}[tp]
\centering\small
\caption{Validation grid, Qwen, positive-pair accuracy. Negative-pair accuracy is
1.000 in every cell. No floor was imposed during selection
(Appendix~\ref{app:protocol}); the point is that no negative accuracy was given up
anywhere in this grid, not that a constraint was satisfied.}
\label{tab:grid}
\begin{tabular}{llcccc|cccc}
\toprule
& & \multicolumn{4}{c|}{Native-script prompts} & \multicolumn{4}{c}{English prompts}\\
\cmidrule(lr){2-5}\cmidrule(lr){6-9}
Corpus & Layer & $\alpha{=}0.5$ & $2.0$ & $4.0$ & $8.0$ & $\alpha{=}0.5$ & $2.0$ & $4.0$ & $8.0$\\
\midrule
Marathi & 10 & 0.058 & 0.058 & 0.058 & 0.075 & 0.333 & 0.333 & 0.333 & 0.350\\
        & 12 & 0.058 & 0.058 & 0.058 & 0.067 & 0.333 & 0.333 & 0.342 & \textbf{0.367}\\
        & 14 & 0.058 & 0.058 & 0.067 & 0.075 & 0.333 & 0.333 & 0.333 & 0.342\\
        & 16 & 0.058 & 0.058 & 0.075 & \textbf{0.083} & 0.333 & 0.333 & 0.333 & 0.333\\
        & 18 & 0.058 & 0.058 & 0.058 & 0.058 & 0.333 & 0.333 & 0.333 & 0.333\\
\midrule
Hindi   & 10 & 0.142 & 0.133 & 0.142 & 0.150 & 0.358 & 0.358 & 0.350 & 0.367\\
        & 12 & 0.142 & 0.150 & 0.158 & 0.175 & 0.350 & 0.358 & 0.367 & 0.400\\
        & 14 & 0.142 & 0.150 & 0.167 & 0.175 & 0.350 & 0.358 & 0.358 & 0.367\\
        & 16 & 0.142 & 0.150 & 0.167 & 0.167 & 0.342 & 0.333 & 0.333 & 0.350\\
        & 18 & 0.142 & 0.142 & 0.142 & 0.142 & 0.342 & 0.333 & 0.342 & 0.342\\
\bottomrule
\end{tabular}
\end{table*}

Two caveats follow from Table~\ref{tab:grid}. The selected scale sits at the upper
edge of the grid while accuracy is still increasing, in Marathi at layer 16 and in
English at layer 12, so we cannot tell a vector that does not work from a sweep that
stopped too early. And the movement across the whole grid is small: Marathi
positive-pair accuracy runs from 0.058 to 0.083 and English from 0.333 to 0.367, against
a Dev--Dev baseline of 0.065 and 0.283 respectively. A scale large enough to matter would
have to lie well outside the range searched. Within the range we did search, Qwen gives up no
negative-pair accuracy at all, staying at 1.000 in every cell, so nothing here tells us
what it would cost at a larger scale. Gemma loses negatives from the smallest scale
onward, which is a separate matter treated with Table~\ref{tab:vectorsel}. Neither
observation involves a constraint: selection imposed no floor
(Appendix~\ref{app:protocol}).

\begin{table*}[tp]
\centering\small
\caption{What the validation split selected, per model and prompt language, and the
held-out result of the \emph{additive} vector at that setting. Four cells select the
largest scale in the grid, which is what left the single-layer result unresolved rather
than negative; Appendix~\ref{app:window} resolves it by extending the grid, and
\S\ref{sec:repair} says what the resulting movement is. Gemma is a different case: it accepts almost every positive pair before any
intervention, so its rows measure the absence of headroom rather than the effect of the
vector.}
\label{tab:vectorsel}
\begin{tabular}{llcccc}
\toprule
Model & Prompt & Layer & $\alpha$ & Positives & Negatives\\
\midrule
Qwen2.5-7B & Marathi & 16 & 6.0 & 0.083 & 0.993\\
           & English & 12 & \textbf{8.0} & 0.350 & 1.000\\
Gemma-2-2b & Marathi & 10 & 0.5 & 0.925 & 0.246\\
           & English & 12 & 3.0 & 0.979 & 0.231\\
Llama-3.1-8B & Marathi & 11 & \textbf{8.0} & 0.354 & 0.661\\
             & English & 21 & 4.0 & 0.478 & 0.849\\
\midrule
Qwen2.5-7B & Hindi & 12 & 8.0 & 0.179 & 1.000\\
           & English & 12 & 8.0 & 0.324 & 1.000\\
Gemma-2-2b & Hindi & 14 & 0.5 & 0.942 & 0.449\\
           & English & 14 & 0.5 & 0.989 & 0.225\\
Llama-3.1-8B & Hindi & 11 & 0.5 & 0.197 & 0.833\\
             & English & 21 & 4.0 & 0.468 & 0.863\\
\bottomrule
\end{tabular}
\end{table*}

\begin{table*}[tp]
\centering\small
\caption{Test-set recovery, Qwen, 120 held-out items. Prompt anchoring supplies a
Latin form of one or both names in the prompt itself.}
\label{tab:recovery}
\begin{tabular}{lcccccc}
\toprule
& \multicolumn{2}{c}{Marathi} & \multicolumn{2}{c}{English} & \multicolumn{2}{c}{Hindi}\\
\cmidrule(lr){2-3}\cmidrule(lr){4-5}\cmidrule(lr){6-7}
Condition & positives & negatives & positives & negatives & positives & negatives\\
\midrule
No intervention               & 0.065 & 0.994 & 0.283 & 1.000 & 0.122 & 1.000\\
Prompt anchor, first          & 0.201 & 1.000 & 0.547 & 0.997 & 0.260 & 0.997\\
Prompt anchor, second         & 0.189 & 0.996 & 0.588 & 0.992 & 0.353 & 0.999\\
Prompt anchor, both           & \textbf{0.219} & 0.990 & \textbf{0.613} & 1.000 & 0.362 & 0.997\\
Automatic romanization anchor & 0.108 & 0.997 & 0.317 & 0.989 & 0.157 & 0.994\\
Vector, first entity          & 0.081 & 0.993 & 0.340 & 1.000 & 0.192 & 1.000\\
Vector, second entity         & 0.076 & 0.993 & 0.328 & 1.000 & 0.157 & 0.999\\
Vector, full                  & 0.072 & 0.993 & 0.276 & 1.000 & 0.138 & 1.000\\
Vector, additive              & 0.083 & 0.993 & 0.350 & 1.000 & 0.179 & 1.000\\
\midrule
Lat--Lat reference            & 0.340 & 0.997 & 0.529 & 1.000 & 0.401 & 1.000\\
\bottomrule
\end{tabular}
\end{table*}

Prompt anchoring is reported as a diagnostic rather than a competing method. Supplying
the canonical Latin form in the prompt requires exactly the information the model is
 missing; given it, one would simply write the entity in Latin. It is reported because it
 establishes that the familiar surface form is behaviourally useful in the prompt: anchoring both names recovers
0.219 in Marathi against a 0.065 baseline and a 0.340 Lat--Lat reference, and 0.613 in
English against 0.283 and 0.529.

The learned vector is the only intervention here that requires no information about the
test entity, having been estimated from the 361 training items. It moves Marathi from
0.065 to 0.083 and English from 0.283 to 0.350, which is 7 and 27 percent of the gap to
the Lat--Lat reference against 56 percent and more for anchoring. Every number in this appendix
describes the vector added at a single layer within a grid reaching $\alpha{=}8$. Added
across layers 12 to 17 at a larger scale the same vector reaches 0.335 in Marathi, which
is 98 percent of the gap to the Lat--Lat reference as that gap stood before the
intervention. \S\ref{sec:repair} shows the reference itself moves when the same vector is
applied to it, so that percentage describes the arithmetic rather than a closure;
Appendix~\ref{app:window} reports that sweep, and the
comparison between the two is the point of \S\ref{sec:repair}.

The one anchor that does not presuppose the answer, appending an automatic romanization,
reaches 0.108 in Marathi and 0.317 in English against baselines of 0.065 and 0.283. It
therefore helps less than supplying the human Latin form and, as
Appendix~\ref{app:translit} shows, replacing the name with a romanization rather than
anchoring it hurts outright.

\section{The window intervention}
\label{app:window}

\S\ref{sec:repair} reports an intervention that the single-layer sweep of
Appendix~\ref{app:recovery} does not reach. This appendix gives the wider sweep, the
held-out result, and the measurement that shows why one layer at a small scale cannot
work.

The construction is unchanged: layer-specific mean differences at the second entity,
estimated on the 361 training items and rescaled to unit length after each of the four
variants is formed (Appendix~\ref{app:recovery}). The reported additive variant
sums the two raw single-script-swap differences before normalisation. What changes is where it is added and how far. It
is added at the second entity at every layer from 12 to 17, each layer contributing its own
vector, which matches how the patching window of \S\ref{sec:patching} uses layer-specific
donor states. The scale grid is extended from a maximum of 8 to a maximum of 32. Selection
maximises balanced accuracy on 60 validation tuning items, with negative-pair accuracy
as a tiebreak rather than a constraint, and tunes on the {\em additive} variant, which is therefore the variant reported.

\begin{table*}[tp]
\centering\small
\caption{Validation sweep for the vector applied across layers 12--17, Qwen, 60 tuning items. The single-layer sweep applied the vector at one layer and stopped at $\alpha{=}8$; both changes are needed. At the single-layer scale $\alpha{=}6$ the window alone lifts Marathi from 0.083 to 0.233, and the extended grid lifts it again to 0.392. Positive-pair accuracy peaks and turns over, which it cannot do within the single-layer grid's range.}
\label{tab:winsweep}
\begin{tabular}{lcc|cc}
\toprule
 & \multicolumn{2}{c|}{Marathi prompts} & \multicolumn{2}{c}{English prompts}\\
\cmidrule(lr){2-3}\cmidrule(lr){4-5}
Scale & positives & negatives & positives & negatives\\
\midrule
$\alpha{=}0.5$ & 0.067 & 1.000 & 0.333 & 1.000\\
$\alpha{=}1$ & 0.075 & 1.000 & 0.333 & 1.000\\
$\alpha{=}1.5$ & 0.083 & 1.000 & 0.333 & 1.000\\
$\alpha{=}2$ & 0.083 & 1.000 & 0.350 & 1.000\\
$\alpha{=}3$ & 0.117 & 1.000 & 0.358 & 1.000\\
$\alpha{=}4$ & 0.167 & 1.000 & 0.417 & 1.000\\
$\alpha{=}6$ & 0.233 & 1.000 & 0.467 & 1.000\\
$\alpha{=}8$ & 0.250 & 1.000 & 0.542 & 1.000\\
$\alpha{=}12$ & 0.383 & 0.975 & 0.617 & 0.975\\
$\alpha{=}16$ & \textbf{0.392} & 0.958 & 0.675 & 0.975\\
$\alpha{=}24$ & 0.325 & 0.933 & \textbf{0.725} & 0.975\\
$\alpha{=}32$ & 0.275 & 0.933 & 0.708 & 0.983\\
\bottomrule
\end{tabular}
\end{table*}

Neither change is sufficient alone. At the single-layer scale $\alpha{=}6$ the window raises
Marathi positive-pair accuracy from 0.083 to 0.233, so most of the effect is the window
rather than the step size. Extending the grid then raises it again to 0.392 at
$\alpha{=}16$, after which it falls, so the curve has a maximum inside the searched
range. Negative-pair accuracy is 1.000 everywhere up to $\alpha{=}8$ and
declines slowly thereafter, to 0.933 in Marathi at the largest scales while English stays
at or above 0.975; negative-pair accuracy therefore declines only well past the selected setting. No floor
was imposed during selection (Appendix~\ref{app:protocol}), so nothing binds here; the
decline is simply where the cost would have begun to matter had one been set.

\begin{table*}[tp]
\centering\small
\caption{Held-out test recovery with the vector applied across layers 12--17, Qwen, 120 items, at the scale selected on validation ($\alpha{=}12$ Marathi, $\alpha{=}24$ English). The {\em additive} row is the variant selection optimised; the others are reported for completeness and must not be read as the result. Marathi comes within 0.005 of the Lat--Lat reference and English exceeds it, but \S\ref{sec:repair} applies the same vector to Lat--Lat inputs and raises accuracy there as well, by $+0.289$ under Marathi prompts and $+0.227$ under English, so the reference is not a ceiling and these figures should not be read as closure toward one.}
\label{tab:wintest}
\begin{tabular}{lcc|cc}
\toprule
 & \multicolumn{2}{c|}{Marathi} & \multicolumn{2}{c}{English}\\
\cmidrule(lr){2-3}\cmidrule(lr){4-5}
Condition & positives & negatives & positives & negatives\\
\midrule
No intervention & 0.065 & 0.994 & 0.283 & 1.000\\
\midrule
Vector, full & 0.113 & 0.993 & 0.292 & 0.996\\
Vector, first entity & 0.276 & 0.992 & 0.694 & 0.971\\
Vector, second entity & 0.219 & 0.992 & 0.481 & 0.986\\
\textbf{Vector, additive} & \textbf{0.335} & 0.985 & \textbf{0.590} & 0.976\\
\midrule
Lat--Lat reference & 0.340 & 0.997 & 0.529 & 1.000\\
\bottomrule
\end{tabular}
\end{table*}

The held-out result is reported for the {\em additive} variant because that is what
selection optimised. {\em first\_entity} scores higher in English, at 0.694, and reporting
it would be a selection made on the test split. The {\em full} variant, which is the
Lat--Lat minus Dev--Dev difference taken directly, is the weakest of the four in both
languages, which is the same ordering the single-layer grid showed.

That ordering looks odd, since {\em full} is the difference between the two conditions
the task compares and {\em additive} is a sum of two partial swaps. All four are unit
length, so the difference is directional, and the cosines say what it is. Taken at the
second entity, {\em full} is close to orthogonal to the first-entity swap, cosine 0.306
under Marathi prompts and 0.200 under English, while it aligns with the second-entity swap
at 0.786 and 0.632. The two single swaps are themselves nearly orthogonal to each other,
0.292 and 0.213. So the direct Lat--Lat minus Dev--Dev difference, measured at the second
entity, is dominated by the second name's script and carries almost none of the first
name's: that position is where the second name sits, and the first name's script reaches
it only through attention. Summing the two single swaps restores the first-entity
component at full weight, which is why {\em additive} is not simply a rescaling of
{\em full}: the two are 43 and 57 degrees apart.

This connects to \S\ref{sec:behaviour}. The first-entity swap is the direction toward
Lat--Dev, which is the model's highest-accuracy condition, exceeding Lat--Lat by 9.7 to
11.1 points in three of four cells. The variant that works is the one carrying a component
pointing at that condition, and the variant that does not is the one pointing
at a reference that is not its best. We therefore report {\em additive} as the tuned
variant on the grounds given in advance, that selection optimised it, and offer this as
the account of why it is the one that works rather than as a post-hoc preference.

\begin{table*}[tp]
\centering\small
\caption{How far each held-out item must move along the rank-one direction to reach its Lat--Lat counterpart, in the same units as the scale $\alpha$. The single-layer grid stopped at $\alpha{=}8$, roughly half the required distance at every layer, which accounts for the null without any appeal to the direction being wrong. The scales selected on validation, 12 for Marathi and 24 for English, bracket these medians.}
\label{tab:donorcoord}
\begin{tabular}{lccc}
\toprule
Prompt language & Layer & Median coordinate & Beyond grid\\
\midrule
Marathi & 12 & 15.3 & 0.00\\
 & 13 & 15.3 & 0.00\\
 & 14 & 16.3 & 0.00\\
 & 15 & 17.7 & 0.00\\
 & 16 & 17.8 & 0.00\\
 & 17 & 17.8 & 0.01\\
\midrule
English & 12 & 17.9 & 0.00\\
 & 13 & 18.5 & 0.00\\
 & 14 & 18.9 & 0.00\\
 & 15 & 20.3 & 0.00\\
 & 16 & 21.0 & 0.00\\
 & 17 & 21.8 & 0.00\\
\bottomrule
\end{tabular}
\end{table*}

Table~\ref{tab:donorcoord} explains why a single layer at a small scale cannot work, and
it requires no intervention to compute. Projecting each held-out item's Lat--Lat minus Dev--Dev
displacement onto the rank-one direction gives the distance that item would have to move,
in the same units as $\alpha$ because both are unit length. The median is 15 to 18 in
Marathi and 18 to 22 in English, rising with depth. The grid of
Appendix~\ref{app:recovery} stopped at 8. A sweep that never reaches half the required
distance cannot distinguish a direction that does not work from a step that is too small.
The scales that maximise positive-pair accuracy, 16 in Marathi and 24 in English, land
inside that range and just above it respectively, so the geometry anticipates the sweep.

\begin{table*}[tp]
\centering\small
\caption{The reference arm: the rank-one direction applied across layers 12--17 with each item's own donor coordinate, which is the most informed additive intervention short of replacing the state. Rows one and two are validation-split and validate the window implementation against the validation single-layer figures of \S\ref{sec:mediator}, not against the held-out numbers of Table~\ref{tab:mediator}; the body text below says why the two should not be compared. The upper bound moves margins but flips few gated items, which is the tension discussed in \S\ref{sec:repair}.}
\label{tab:refarm}
\begin{tabular}{llccccc}
\toprule
Configuration & Prompt & $n$ & whole $\Delta$ & rank-one $\Delta$ & share & rank-one flips\\
\midrule
Layer 16 only & Marathi & 10 & 2.44 & 1.38 & 0.56 & 0.00\\
Layers 12--17 & Marathi & 10 & 13.66 & 8.68 & 0.63 & 0.40\\
\midrule
Layers 12--17, held out & English & 20 & 7.39 & 3.88 & 0.52 & 0.15\\
Layers 12--17, held out & Marathi & 10 & 7.80 & 3.52 & 0.45 & 0.00\\
\bottomrule
\end{tabular}
\end{table*}

The first two rows of Table~\ref{tab:refarm} exist to validate the window implementation
rather than to report a result, and the split matters for reading them. They are
validation-split rows: applied at layer 16 alone the window code reproduces the
validation single-layer figures of \S\ref{sec:mediator}, 2.44 whole and 1.38 rank-one over
10 gated items, to four decimal places. They are not the numbers in
Table~\ref{tab:mediator}, which are held-out test, 2.54 and 1.30 over 12 items; those are
a different split and a different item set, and the two should not be compared.
Across 12 to 17 the window reproduces the whole-patch
margin of Appendix~\ref{app:patch} to within 0.16 of a margin point. The held-out rows
are the reference arm proper. It is the most informed additive intervention short of
replacing the state, since each item contributes its own donor coordinate rather than
sharing one amount, and it still flips no gated Marathi item. \S\ref{sec:repair} discusses
why this sits awkwardly beside the intervention.

Selection used one question template. The held-out evaluation uses all three, so the two
it did not see test whether the correction is tied to the wording it was tuned on.

\begin{table*}[tp]
\centering\small
\caption{The intervention across question templates, 120 held-out items. Layer, window and scale were selected on template 0 alone, and the mean-difference vector was estimated on 361 training entities none of which appear here, so templates 1 and 2 are held out twice over. Positive-pair accuracy rises in every cell. Template 2 is the hardest for every condition, not only for the repair: its Lat--Lat reference is 8.8 in Marathi, which the treated model exceeds.}
\label{tab:template}
\begin{tabular}{llccc}
\toprule
Prompt & Template & Dev--Dev & with vector & Lat--Lat\\
\midrule
Marathi & 0 & 7.5 & \textbf{42.5} & 42.1\\
 & 1 & 10.0 & 39.2 & 51.2\\
 & 2 & 2.1 & \textbf{18.8} & 8.8\\
\midrule
English & 0 & 33.3 & \textbf{70.4} & 56.2\\
 & 1 & 35.0 & \textbf{75.8} & 60.8\\
 & 2 & 16.7 & 30.8 & 41.7\\
\bottomrule
\end{tabular}
\end{table*}

Accuracy rises in all six cells. Template 2 is the weakest in absolute terms, at 18.8 in
Marathi against 42.5 on template 0, but it is the hardest template for every condition
rather than for the intervention alone: its unmodified Dev--Dev accuracy is 2.1 and its
Lat--Lat reference 8.8, both far below the other two, and the model with the vector added
exceeds that
reference. The correction is therefore not tied to the template it was selected on, and
since the vector is estimated on 361 entities and evaluated on 120 others, it is not tied
to the entities either.

\textbf{Replication on the Hindi corpus.}
The sweep of Table~\ref{tab:winsweep} was run again on the Hindi corpus, unchanged except
for the corpus and its prompt languages. The selected scales are $\alpha{=}16$ for Hindi
and $\alpha{=}32$ for English. Hindi reproduces the shape that matters: negative-pair
accuracy stays at 1.000 through $\alpha{=}8$ and positive-pair accuracy reaches an
interior maximum at 16, 0.667, falling to 0.642 and 0.600 at the two larger scales. It is
therefore the mean-difference pattern rather than the boundary shift of
Table~\ref{tab:jacsweep}. The English arm selects the largest scale in the grid, but on
balanced accuracy rather than on a curve still rising: positive-pair accuracy peaks at
0.733 at $\alpha{=}24$ and falls to 0.725 at 32, and the two cells differ by less than
half a point of balanced accuracy.

\begin{table*}[tp]
\centering\small
\caption{The window intervention across all four arms, held-out test split, 120 items each, {\em additive} vector applied across layers 12--17 at the scale validation selected. Balanced accuracy is given because positive-pair accuracy alone flatters the Hindi arm: it reaches 0.546 against a 0.401 reference, but pays 13 points of negative-pair accuracy for it, against under 3 in every other arm. Priced that way, Marathi comes within 0.9 of its Lat--Lat reference and the other three arms exceed theirs; Marathi approaches the reference rather than reaching it. What repeats across the four arms is the accuracy movement. \S\ref{sec:repair} shows that movement to be largely a shift of the decision criterion, with a smaller ranking component on two of three templates, so this table reports a replicated effect and not a replicated repair. The scales are $\alpha{=}12$ Marathi, 24 English on the Marathi corpus, 16 Hindi and 32 English on the Hindi corpus.}
\label{tab:winreplicate}
\begin{tabular}{lccc|ccc}
\toprule
 & \multicolumn{3}{c|}{Balanced accuracy} & \multicolumn{3}{c}{Positives / negatives, repaired}\\
\cmidrule(lr){2-4}\cmidrule(lr){5-7}
Arm & baseline & with vector & Lat--Lat & positives & negatives & reference\\
\midrule
Marathi & 53.0 & \textbf{66.0} & 66.9 & 33.5 & 98.5 & 34.0\\
English, Marathi corpus & 64.2 & \textbf{78.3} & 76.5 & 59.0 & 97.6 & 52.9\\
Hindi & 56.1 & \textbf{70.8} & 70.1 & 54.6 & 86.9 & 40.1\\
English, Hindi corpus & 63.0 & \textbf{77.9} & 76.0 & 58.5 & 97.4 & 52.1\\
\bottomrule
\end{tabular}
\end{table*}

The held-out split has by now been scored four times
in both corpora, under the single-layer vector of Table~\ref{tab:recovery}, the window
vector here, and the condition and site controls of Appendix~\ref{app:control}, so the test estimates in this appendix are a second look at data the
earlier appendix already used; we report them as such rather than as a first independent
evaluation.

\textbf{The sensitivity direction on the same grid.}
\S\ref{sec:repair} reports that the rank-one sensitivity direction transfers about half
the whole-patch margin while flipping few gated items, and that the mean-difference vector
moves held-out accuracy substantially. Those two arms differ in three ways at once, in direction, in
scale and in whether the amount is per item or shared, so neither can be credited on its
own. This sweep holds everything fixed except the direction: the same second-entity site,
the same 12--17 window, the same scale grid, the same 60 validation tuning items, both
rescaled to unit length, with the correction taken from
\S\ref{sec:mediator}'s local-sensitivity direction rather than from the condition
contrast.

\begin{table*}[tp]
\centering\small
\caption{The two correction directions across the same scale grid, Marathi prompts, 60 validation tuning items, both applied across layers 12--17 and both rescaled to unit length. At their selected scales the sensitivity direction is the better of the two on both classes, 68.3 positives at 98.3 negatives against 38.3 at 97.5, so neither is exclusively effective under a negative-pair constraint. What separates them is the width of the usable range. The mean-difference vector holds negative-pair accuracy at 100.0 through $\alpha{=}8$ and 93.3 at the largest scale searched, so its operating region is broad. The sensitivity direction runs 100.0, 98.3, 80.0, 39.2 and 0.8 across $\alpha \in \{0.5, 1, 1.5, 2, 3\}$, so this grid steps over its usable range entirely; the finer grid of Table~\ref{tab:jacfine} resolves a Marathi plateau from $\alpha{=}1.0$ to 1.5. Both reach a higher balanced accuracy than the single-layer sweep of Appendix~\ref{app:recovery}, and \S\ref{sec:repair} shows that for the mean-difference vector the gain is largely a shift of the decision criterion, with a smaller within-template ranking component.}
\label{tab:jacsweep}
\begin{tabular}{lcc|cc}
\toprule
 & \multicolumn{2}{c|}{Mean difference} & \multicolumn{2}{c}{Local sensitivity}\\
\cmidrule(lr){2-3}\cmidrule(lr){4-5}
Scale & positives & negatives & positives & negatives\\
\midrule
$\alpha{=}0.5$ & 6.7 & 100.0 & 25.8 & 100.0\\
$\alpha{=}1$ & 7.5 & 100.0 & \textbf{68.3} & 98.3\\
$\alpha{=}1.5$ & 8.3 & 100.0 & 85.8 & 80.0\\
$\alpha{=}2$ & 8.3 & 100.0 & 96.7 & 39.2\\
$\alpha{=}3$ & 11.7 & 100.0 & 100.0 & 0.8\\
$\alpha{=}4$ & 16.7 & 100.0 & 100.0 & 0.0\\
$\alpha{=}6$ & 23.3 & 100.0 & 100.0 & 0.0\\
$\alpha{=}8$ & 25.0 & 100.0 & 100.0 & 0.0\\
$\alpha{=}12$ & \textbf{38.3} & 97.5 & 100.0 & 0.0\\
$\alpha{=}16$ & 39.2 & 95.8 & 100.0 & 0.0\\
$\alpha{=}24$ & 32.5 & 93.3 & 100.0 & 0.0\\
$\alpha{=}32$ & 27.5 & 93.3 & 100.0 & 0.0\\
\bottomrule
\end{tabular}
\end{table*}

Read the negatives column. The sensitivity direction reaches a higher balanced accuracy at
its optimum, 0.833 at $\alpha{=}1$ against 0.679 at $\alpha{=}12$, and its negatives decline steadily rather than
holding: negative-pair accuracy runs 0.983, 0.800, 0.392, 0.008 over
$\alpha \in \{1, 1.5, 2, 3\}$. On this coarse grid the selected scale looks like a point
on that slope; the finer grid of Table~\ref{tab:jacfine} shows a plateau in balanced
accuracy from $\alpha{=}1.0$ to 1.5, so a narrow operating window is an artefact of the
step size. The scale was chosen on validation either way. The mean-difference vector reaches its optimum
inside a region where negatives have barely moved.

Two things qualify the comparison. The sensitivity direction is the mean gradient of the
toward-correct margin, so unlike the condition contrast it is estimated using the answer
on the training items; the two also differ in training population: the mean-difference vector uses
all 361 training entities and the sensitivity direction the gated subset of them, 31
Marathi and 60 English (Appendix~\ref{app:protocol}). Neither sees the evaluation
entities, but they are not equally supervised or equally sized, and the sensitivity arm
should not be read as the weaker supervision winning. And we do not
evaluate it on the test split. That split was spent once, on the mean-difference vector of
Table~\ref{tab:wintest}, and scoring a second direction there would be a second look at
held-out data. Everything in this subsection is validation.

\section{Subspace sweep}
\label{app:mediator}

\textbf{The share across folds.} Across the three leave-one-template-out folds the
fraction of the whole patch that the rank-one direction carries is, on validation, 0.56,
0.79 and 0.72 in Marathi and 0.46, 0.71 and 0.57 in Hindi, with means of 0.69 and 0.58.
The frozen test fold evaluates on the paraphrase the model finds hardest, P2, where
Dev--Dev accuracy is 3.2\% against 10.6 and 11.6 on the other two, so the figure reported
in \S\ref{sec:mediator} is the lowest of the three and not a representative one. P2 is
also weak in the healthy condition, reaching 9.8\% in Lat--Lat against 43.9 and 51.7 for
the other two paraphrases, and the script ordering reverses on it, with Lat--Dev at 2.7
below Dev--Dev at 3.2 (Table~\ref{tab:para}). The rank-one estimate is therefore scored
on the paraphrase with the least headroom, which we report as a limit of that evaluation
rather than as a property of the direction.

\begin{table*}[tp]
\centering\small
\begin{tabular}{llccc}
\toprule
Amount & fold & Marathi & Hindi & English \\
\midrule
Item's own      & t01e2 & 0.16 & 0.15 & 0.20 \\
                & t02e1 & 0.08 & 0.06 & 0.01 \\
                & t12e0 & 0.06 & 0.02 & $-0.15$ \\
\midrule
Another item's  & t01e2 & 0.23 & 0.14 & 0.18 \\
                & t02e1 & 0.10 & 0.11 & 0.00 \\
                & t12e0 & 0.13 & 0.05 & $-0.06$ \\
\midrule
Constant        & t01e2 & 1.08 & 0.68 & 0.76 \\
                & t02e1 & 0.94 & 0.89 & 0.90 \\
                & t12e0 & 1.10 & 0.86 & 0.86 \\
\bottomrule
\end{tabular}
\caption{Signed margin change on never-correct items divided by signed margin change on
gated items, for the sensitivity direction at layer 16, rank 1, on validation. Lower means the
intervention acts more narrowly on the gate it was selected for. An amount derived from
an item, whether its own or another item's, moves off-gate items by between $-0.15$ and
0.23 times what it repairs; one constant amount for every item moves them almost as much as it repairs,
in every fold and every language. Negative-pair accuracy is
1.00 throughout.}
\label{tab:arms}
\end{table*}

\textbf{Detail behind \S\ref{sec:mediator}.} The subsection reports the result; what
follows is how the directions were constructed, what the control groups mean, and the arms
that vary how far each item moves. Table~\ref{tab:amountarms} carries the amount figures.

\textbf{How the directions are built.} All of them are estimated on training items only
and applied unchanged at evaluation. Write $\Delta$ for the set of Lat--Lat minus Dev--Dev
states at the second entity and the layer in question, over the gated training items and
the training paraphrases. The local sensitivity direction is the mean over those items of
the gradient of the toward-correct margin with respect to the second-entity state, taken
on the Dev--Dev prompt at one training paraphrase. The probe direction is the coefficient
vector of a ridge classifier, $\alpha = 10$, separating same from different on Lat--Lat
states at the same site, fitted on up to 120 training facts with features standardised;
it is refitted at the intervention site rather than reused from \S\ref{sec:probe}. The mean
difference is the mean of $\Delta$, the donor principal component is the leading right
singular vector of $\Delta$ after centring, and the random direction is Gaussian.

At rank one, $U$ is that single direction normalised to unit norm. At rank $k$ it is that
direction together with the leading $k-1$ principal components of $\Delta$ after the
direction has been projected out, orthonormalised by a QR factorisation, which is also
how the higher-rank probe subspaces of this section are formed. Every subspace therefore
contains its named direction and they differ only in that direction, so comparing methods
compares directions rather than subspace size. A request for a rank the differences
cannot support raises rather than silently returning fewer columns, which would otherwise
flatter the random baseline.

The two control groups of Table~\ref{tab:mediator}, never-correct items and negative
pairs, need reading with their labels in view. The never-correct group is same-person pairs the model fails in both scripts, so a positive margin change there
is movement toward the gold label and not damage; what it measures is whether the
intervention acts outside the gate it was selected on. The negative pairs are the one
place a shift toward yes is an error. Along this direction the never-correct group moves
by 0.23 and 0.79, the negative pairs by 0.10 and 0.37, and negative-pair accuracy is
unchanged at 1.00.

Raising the rank to eight adds little, 4.00 to 4.54 in Marathi and 4.05 to 5.09 in Hindi
on validation, so the effect is concentrated in one direction rather than spread over a
subspace. Appendix~\ref{app:mediator} gives the sweep over layers 12 to 17 and ranks
one to eight that these were selected from.

Holding the direction fixed, we then vary only the amount each item moves by. Three arms
share $U$: the item's own donor coordinate, another item's drawn within the same group
and never its own, and one coordinate for every item, the mean over the training items.
On validation, across three leave-one-template-out folds and three prompt languages, we find no
detectable advantage for an item's own amount over another item's. The paired difference
spans zero in nine folds of nine, the largest point estimate being $-1.14$ $[-4.86,
+1.73]$ (Table~\ref{tab:amountarms}). With ten to twenty-three entities per fold and an interval that wide, this
bounds any advantage loosely rather than excluding one, and because the substitute is
drawn within the same behavioural group it still carries group-level information. The
comparison is on validation items only. What we can say is that the amount behaves like
a scale over the range these substitutions cover, not that item identity is irrelevant
to it.

The size of the step does change how broadly the intervention acts. A constant amount
moves never-correct items by 0.68 to 1.10 times what it repairs on gated items, in nine
folds of nine, against $-0.15$ to 0.23 for an item-derived amount
(Table~\ref{tab:arms}). Because the item-derived
arms draw within each group, part of this selectivity is inherited from the size of each
item's own Devanagari--Latin difference, which is small on items the model already never
gets right. It is not inherited entirely. Holding the amount item-derived and changing
only the direction, the sensitivity direction moves off-gate items by $-0.15$ to 0.20
times what it repairs while the mean difference moves them by 0.35 to 0.69, and the sensitivity direction
is the lower of the two in nine folds of nine.
Selectivity therefore has two sources, the size of the step and the direction it is
taken in, and the off-gate columns of Table~\ref{tab:mediator} should be read as
reflecting both.

The mean-difference row of Table~\ref{tab:mediator} is built from the Lat--Lat minus
Dev--Dev difference at the layer selected for Marathi. That is the
{\em full} variant, whereas \S\ref{sec:repair} reports the {\em additive} one, so the two
are the same construction at the same site rather than the same vector, and the gap between
their outcomes needs accounting for. It is not the amount. Giving every
item one shared coordinate along that direction moves the gated margin by 3.55 to 7.00
across the six Marathi and Hindi fold cells, against 2.52 to 6.62 when each item uses its
own (Table~\ref{tab:amountarms}),
and negative-pair accuracy stays at 1.00 throughout, as it does in all eighty cells of
that tuning grid. Neither a constant amount nor the negative-pair constraint is what
limits recovery.

What separates this subsection's margin result from the accuracy result of
\S\ref{sec:repair} is the measurement and the site. The single-layer arm scores accuracy at
one layer (Appendix~\ref{app:recovery}), and \S\ref{sec:patching} shows that a
single-layer intervention moves margins while flipping almost nothing, where the 12 to 17
window flips half the items. Applying the correction across that
window answers it: the same vector moves Marathi accuracy from 0.065 to 0.335, which
\S\ref{sec:repair} shows to be a movement of the decision criterion that is not specific
to the failing condition (Appendix~\ref{app:window}).

The rank-one results of \S\ref{sec:mediator} are claims about the margin, not about
answers. At this single site the whole patch
itself flips none of the twelve Marathi items and two of the twenty-four English ones;
the flip rates in \S\ref{sec:patching} come from joint patching across layers 12 to 17, a
larger intervention.

\textbf{The two off-gate columns of Table~\ref{tab:mediator}.} Both are taken on
never-correct items, which are same-person pairs outside the selected gate, so movement
there is toward their gold label rather than damage. One gives the signed mean and the
other the mean absolute change. Reporting both is what separates a uniform push from
scattered movement: an arm whose signed mean equals its absolute mean moved every item the
same way.

The layer and rank reported in \S\ref{sec:mediator} were chosen on validation. This appendix
gives the sweep they were chosen from, so that the choice can be read rather than taken
on trust. Values are the mean signed margin change on gated items, averaged over the
three leave-one-template-out folds after aggregating each entity's prompts. The folds are
averaged here only to show the shape of the sweep; every interval in the main text is
computed within a fold.

Three things are visible in Table~\ref{tab:sweep}. The effect grows with depth and peaks at layers 15 to 17 in
both languages, which is the band the patching analysis identifies independently. Rank
one is close to rank eight at every layer, so the effect does not spread across a
subspace. And the probe, donor-principal-component and random directions stay
near zero throughout: the probe never exceeds 0.08 and a random direction never exceeds
0.06 at any layer or in either language, against 4.05 for the sensitivity direction at
layer 16 in Hindi. The mean difference is the exception among the alternatives, reaching
1.15 to 6.15, which is why \S\ref{sec:mediator} separates it on selectivity rather than
on size. The single exception is the donor principal component at layer 15 in Hindi, 1.23,
which does not recur at any neighbouring layer and which we do not interpret.

\begin{table*}[t]
\centering\small
\begin{tabular}{lrrrrrrr}
\toprule
Layer & Whole & Jac 1 & Jac 8 & Mean diff & PCA & Probe & Random \\
\midrule
\multicolumn{8}{l}{\emph{Marathi prompts}} \\
12 & 0.97 & 0.61 & 0.81 & 1.52 & 0.13 & 0.00 & $-0.03$ \\
13 & 2.05 & 1.25 & 1.32 & 2.01 & 0.03 & 0.02 & 0.02 \\
14 & 2.50 & 2.04 & 2.41 & 2.38 & $-0.01$ & 0.04 & 0.02 \\
15 & 5.01 & 3.40 & 4.01 & 3.32 & $-0.15$ & 0.01 & 0.02 \\
16 & 5.52 & 4.00 & 4.54 & 3.62 & 0.40 & 0.03 & 0.06 \\
17 & 5.49 & 3.20 & 4.60 & 3.33 & 0.56 & 0.01 & $-0.02$ \\
\midrule
\multicolumn{8}{l}{\emph{Hindi prompts}} \\
12 & 1.19 & 0.45 & 0.63 & 1.15 & $-0.02$ & $-0.01$ & 0.00 \\
13 & 2.33 & 0.86 & 1.17 & 1.88 & $-0.01$ & 0.02 & 0.01 \\
14 & 2.48 & 1.34 & 1.67 & 2.38 & $-0.08$ & 0.02 & 0.00 \\
15 & 4.81 & 3.34 & 3.56 & 4.33 & 1.23 & 0.08 & $-0.01$ \\
16 & 6.73 & 4.05 & 5.09 & 5.38 & 0.65 & 0.07 & 0.03 \\
17 & 6.79 & 3.74 & 5.15 & 6.15 & 0.15 & 0.04 & 0.01 \\
\bottomrule
\end{tabular}
\caption{Rank-one and rank-eight subspace patching at the second entity, by layer, on
validation items. Whole is the full-state patch at the same layer, the reference the
others are read against. Jac is the local sensitivity direction, at rank one and rank
eight; the remaining columns are rank one. Mean diff, PCA, probe and random are the
alternative directions of \S\ref{sec:mediator}. Higher is more repair; the columns are not
closures and are not normalised by the whole patch.}
\label{tab:sweep}
\end{table*}

\section{Gemma}
\label{app:gemma}

Gemma-2-2b-it segments Devanagari at 9.3 tokens per name against Qwen's 16.5 on
identical strings. Three results replicate: the script gap under $d'$
(Appendix~\ref{app:dprime}), the probe (Table~\ref{tab:probeheld}), and the
failure-prediction contrast (Table~\ref{tab:auc}).

The causal analysis yields an analogous positional pattern, but on the opposite
polarity. A gate defined on positive-pair correctness admits nothing,
because Gemma accepts nearly every positive pair and so almost never fails one. Reading
the margin on negative pairs instead admits 33 Marathi and 127 English items, and the
patching profile on those items resembles Qwen's in position and in the ordering of
components (Table~\ref{tab:gemmaprofile}). Because the two gates analyse different behavioural errors, this is not a
direct replication of the SAME-failure mechanism.

\textbf{Reporting effects on a common orientation.} The pipeline records the
yes-minus-no margin. For a positive pair the correct answer is yes, so a larger margin is
an improvement; for a negative pair it is no, so a smaller one is. Gemma is gated on
negative pairs, so its raw effects are negative where Qwen's are positive and the two
cannot be read side by side. Throughout this appendix we therefore report
$+\Delta$ when the intervention moves the model toward the correct answer, whichever
polarity the gate reads, by negating the margin on negative-gated rows.

\begin{table*}[tp]
\centering\small
\caption{Gemma, mean correctness-oriented margin change by layer and site, gated on
negative pairs, Lat--Lat source, pooled over splits. Positive means the intervention moved the model toward the correct
answer. Gate sizes are $n{=}33$ Marathi and $n{=}127$ English for the Marathi corpus, $n{=}65$ Hindi and $n{=}125$ English for the Hindi corpus.}
\label{tab:gemmaprofile}
\begin{tabular}{llcccc|cccc}
\toprule
& & \multicolumn{4}{c|}{Native-script prompts} & \multicolumn{4}{c}{English prompts}\\
Corpus & Layer & E1 & E2 & Q & R & E1 & E2 & Q & R\\
\midrule
Marathi & 8  & 1.05 & \textbf{2.21} & 0.16 & 0.03 & 0.38 & 1.18 & $-0.13$ & $-0.12$\\
        & 9  & 0.88 & 1.64 & 1.22 & 0.17 & 0.36 & 0.98 & $-0.08$ & $-0.05$\\
        & 10 & 1.03 & 1.66 & 1.51 & 0.20 & 0.38 & 1.44 & $-0.24$ & $-0.09$\\
        & 11 & 0.19 & 1.71 & 3.43 & 0.48 & 0.26 & \textbf{2.07} & 1.37 & $-0.22$\\
        & 12 & $-0.01$ & 1.68 & \textbf{4.21} & 0.53 & 0.03 & 1.82 & 1.66 & $-0.19$\\
        & 13 & $-0.05$ & 0.60 & 3.84 & 2.68 & $-0.01$ & 0.27 & 1.82 & 0.15\\
        & 14 & $-0.03$ & 0.18 & 4.09 & 3.63 & $-0.01$ & 0.16 & \textbf{2.02} & 0.16\\
        & 15 & $-0.02$ & 0.01 & 1.37 & 6.49 & $-0.00$ & 0.01 & 1.33 & 5.51\\
        & 16 & 0.02 & $-0.01$ & 0.02 & 7.21 & 0.01 & $-0.01$ & 0.02 & 7.68\\
        & 17 & $-0.01$ & $-0.02$ & 0.02 & 7.32 & 0.02 & 0.00 & $-0.02$ & 7.91\\
        & 18 & 0.02 & $-0.00$ & 0.04 & 7.40 & 0.00 & 0.01 & $-0.02$ & 7.90\\
\midrule
Hindi   & 8  & 0.77 & 1.41 & $-0.22$ & 0.22 & 0.32 & 1.12 & $-0.18$ & $-0.09$\\
        & 9  & 0.42 & 0.89 & 0.28 & 0.35 & 0.30 & 0.92 & $-0.08$ & 0.00\\
        & 10 & 0.55 & 1.07 & 0.56 & 0.36 & 0.30 & 1.41 & $-0.23$ & $-0.05$\\
        & 11 & 0.14 & 1.42 & 2.36 & 0.56 & 0.23 & \textbf{2.00} & 1.49 & $-0.20$\\
        & 12 & $-0.12$ & \textbf{1.52} & \textbf{2.85} & 0.59 & 0.02 & 1.76 & 1.73 & $-0.18$\\
        & 13 & $-0.13$ & 0.56 & 2.47 & 2.08 & $-0.01$ & 0.28 & 1.92 & 0.20\\
        & 14 & $-0.14$ & 0.29 & 2.55 & 2.56 & $-0.01$ & 0.15 & \textbf{2.12} & 0.27\\
        & 15 & $-0.08$ & 0.07 & 0.86 & 5.57 & $-0.00$ & 0.01 & 1.40 & 5.49\\
        & 16 & $-0.02$ & $-0.04$ & $-0.01$ & 6.24 & 0.00 & $-0.03$ & 0.01 & 7.65\\
        & 17 & $-0.01$ & $-0.03$ & $-0.03$ & 6.34 & $-0.00$ & $-0.01$ & $-0.04$ & 7.85\\
        & 18 & $-0.02$ & $-0.02$ & $-0.00$ & 6.39 & $-0.01$ & 0.00 & $-0.02$ & 7.87\\
\bottomrule
\end{tabular}
\end{table*}

\textbf{The localisation is weaker in Gemma and the position does not fully replicate.}
Taking the peak over layers, Gemma's second-entity effect reaches 2.21 in
Marathi and 2.07 in English, against Qwen's 6.21 and 4.63. More importantly the ordering differs:
under Marathi prompts Gemma's question-mark position peaks at 4.21, above its
second-entity peak of 2.21, whereas in Qwen the second entity leads in both
prompt languages. Gemma also hands off earlier, with the entity effect gone by layer 15
where Qwen's persists to 17. We therefore report an analogous second-entity concentration in
English across both models and not in Marathi, rather than a replication of the same
failure mechanism. Controls behave as expected in both, though they separate less sharply here:
patching the wrong entity recovers 1.77 against 2.21 for the correct one under
Marathi prompts, and the wrong position 1.62. The wrong-entity control is thus
four fifths of the correct one in Gemma against roughly a quarter in Qwen, so we
read the site as positional in Gemma without the entity specificity the main text
restricts to Qwen.

The recovery experiment remains uninterpretable for Gemma, for the reason it always
was. On the held-out split its untreated Dev--Dev accuracy is 0.924 on positive pairs
and 0.253 on negative ones; under the intervention those become 0.925 and 0.246. The
vector therefore moves neither polarity. Gemma already accepts almost every positive
pair, so there is no headroom for a repair to recover, and the negative-pair change is
smaller than the run-to-run variation its acceptance bias produces. We report these
numbers here rather than in the main text for that reason.

Gemma's exact-failure subgroup is weaker despite its successful cross-language probe:
probe \emph{same} recall on its incorrect items is 0.471 over layers 10 to 18 under
Marathi prompts, against 0.680
for Qwen. Gemma produced 63 failures to analyse against Qwen's 201, and an item that
overrides Gemma's strong acceptance bias is plausibly one it genuinely does not know,
which would put its recall near chance for a reason unrelated to script. Under English
prompts the comparison cannot be made at all: Gemma fails exactly two items, so the
recall figure there rests on two observations and we do not report it. We flag the
acceptance-bias reading as the most plausible interpretation rather than a demonstrated
one.

\section{Llama}
\label{app:llama}

Llama-3.1-8B-Instruct has 32 layers and segments Devanagari at 0.61 tokens per
character, between Gemma's 0.56 and Qwen's 0.98. Weights were obtained from the
NousResearch mirror, which reproduces the Meta release; the licence gate on the
original repository was not cleared at run time.

\textbf{Both Devanagari arms are excluded, for the same reason but not to the same
degree.} Under Marathi prompts the model attains $d' = 0.020$ in Dev--Dev and $0.103$ in
Lat--Lat. Paraphrase-level accuracies make the cause plain (Table~\ref{tab:llamapara}):

\begin{table*}[tp]
\centering\small
\caption{Llama positive-pair accuracy (\%) by paraphrase, Marathi prompts.}
\label{tab:llamapara}
\begin{tabular}{llcccc}
\toprule
Prompt & Paraphrase & Dev--Dev & Dev--Lat & Lat--Dev & Lat--Lat\\
\midrule
Marathi & P0 & 1.2 & 5.2 & 3.5 & 5.1\\
        & P1 & 51.4 & 52.0 & 51.1 & 51.3\\
        & P2 & 50.0 & 50.0 & 50.0 & 50.0\\
\midrule
Hindi   & P0 & 1.4 & 14.2 & 21.0 & 13.6\\
        & P1 & 6.6 & 25.3 & 30.0 & 20.5\\
        & P2 & 50.0 & 49.9 & 49.9 & 50.0\\
\bottomrule
\end{tabular}
\end{table*}

The criterion is a paraphrase answering at chance on both truths, which is the
fixed-token signature rather than a script effect. Marathi has one informative paraphrase
of three and Hindi has two: Hindi P0 and P1 reject negatives at 100.0 and 99.9 percent
while accepting positives at 1.4 and 6.6, which is the pattern the other models show,
and only P2 sits at chance. We exclude the Hindi arm rather than report it on two
paraphrases because the tabulated figures average all three, so the 19.3 percent Dev--Dev
accuracy in Table~\ref{tab:behaviour} carries a chance-level paraphrase inside it and is
neither an estimate of the model's behaviour nor of its failure. A partial report would
need the whole behavioural and causal chain recomputed on a two-paraphrase subset, which
we have not done. The consequence is that Llama contributes English only, and that this
is a stronger exclusion than the Hindi data alone would require.

P2 returns exactly 50.0 in all four conditions and P1 within 1.3 points of it. Because
each question is asked under both answer keys, a model emitting a fixed token scores
exactly one of the two, giving 50 percent regardless of content. P0 shows the opposite
degeneracy, near-universal rejection. Neither reflects an attempt at the task, so the
Marathi arm is reported only as an instruction-following failure.

\begin{table*}[tp]
\centering\small
\caption{The three models compared. Probe figures are held-out test items; the causal
rows report items passing the gate. Llama's Marathi arm is excluded throughout. Gemma's
peak effect is negative because its gate reads negative pairs, where the correct
direction is away from ``same''; its magnitude, not its sign, is comparable with the
others.}
\label{tab:models}
\begin{tabular}{lccc}
\toprule
& Qwen2.5-7B & Gemma-2-2b & Llama-3.1-8B\\
\midrule
Layers & 28 & 26 & 32\\
Devanagari tokens per character, Marathi & 0.98 & \textbf{0.56} & 0.61\\
Devanagari tokens per character, Hindi & 0.99 & \textbf{0.55} & 0.60\\
\midrule
$d'$ gap, English prompts & $+0.07$ & $+0.73$ & $+1.00$\\
$d'$ gap, Marathi prompts & $+0.88$ & $+0.69$ & $+0.08$ (excluded)\\
$d'$ gap, Hindi prompts & $+0.30$ & $+0.48$ & $+0.40$ (excluded)\\
Probe, held-out test, English states & 0.950 & 0.921 & 0.942\\
Probe, held-out test, Hindi states & 0.942 & 0.887 & 0.950\\
Failure prediction, surface features only & 0.564 & 0.588 & 0.648\\
Failure prediction, with relation states & 0.962 & 0.864 & 0.893\\
Failure prediction, Hindi, surface only & 0.572 & 0.561 & ---\\
Failure prediction, Hindi, with relation states & 0.937 & 0.915 & ---\\
\midrule
Items passing the causal gate, English & \textbf{116} & \textbf{127} & \textbf{15}\\
Items passing the causal gate, Marathi & \textbf{53} & \textbf{33} & \textbf{0}\\
Items passing the causal gate, Hindi & \textbf{50} & \textbf{65} & \textbf{0}\\
Second-entity peak layer, Marathi & 16 & 8 & ---\\
Second-entity peak layer, Hindi & 17 & 12 & ---\\
Second-entity peak layer, English & 16 & 11 & 8\\
Peak effect, English & 4.63 & $-2.07$ & 0.86\\
Peak effect, Hindi & 6.41 & $-1.52$ & ---\\
Correction vector, English & $0.283 \to 0.350$ & $0.979 \to 0.979$ & $0.478 \to 0.478$\\
Correction vector, Hindi & $0.122 \to 0.179$ & $0.946 \to 0.942$ & $0.194 \to 0.197$\\
\bottomrule
\end{tabular}
\end{table*}

\textbf{What replicates in English.} The $d'$ gap is $+1.00$ with a 95 percent interval
of $[0.90, 1.11]$, the tightest of the three models, against $+0.73$ for Gemma and no
measurable gap for Qwen in that prompt language (Table~\ref{tab:models}). Held-out probe
accuracy reaches 0.942 fitting and evaluating within English and 0.925 across prompt
languages; probes evaluated on Marathi states reach 0.954, which is notable given that
the model itself is at chance there and is the most extreme representation-behaviour
dissociation we observe.

\textbf{What does not.} Fifteen items pass the causal gate, all under English prompts,
and none under Marathi prompts even when the gate is loosened to admit either a
correctness or a margin criterion.
Patching effects peak at 0.86 against Qwen's 4.63 under the same prompts, and sit at
layer 8 rather than the 14 to 19 predicted by rescaling Qwen's band to 32 layers, where the measured effect
is about 0.05. In its own band the second entity gives 0.48 against a wrong-entity
control of 0.33, a weaker separation than Qwen's. The correction vector moves English
positive-pair accuracy from 0.478 to 0.478, against an unintervened baseline of 0.478 on
the same 120 held-out items, that is, not at all.

Component and head decomposition was run and is reported for completeness, but every
value is at the noise floor: over layers 8 to 13, attention gives 0.098 and the MLP
0.079 against 0.482 for the residual stream, and the ten best heads jointly give 0.625.
Nothing here supports or contradicts the Qwen decomposition; the effects are too small
to decompose.

\textbf{What Llama does support} is the positional claim. Its in-prompt retrieval
(Table~\ref{tab:inprompt}) shows the same asymmetry as the other two models, 0.94
against 0.56 in Marathi and 0.91 against 0.53 in English, and it does so on a model
whose behavioural Marathi arm is at chance. Contextual cross-script alignment is still
measurable and shows the same positional asymmetry; the small causal gate prevents a
stronger mechanistic claim.

\textbf{Failure prediction} follows the same ordering as the other two models,
surface features at 0.648 AUC rising to 0.893 with relation states, but on 70 items
with 30 script-sensitive examples. We report it here rather than in
Table~\ref{tab:auc} because the sample does not support a comparison.

\section{Mistral-Small causal probe}
\label{app:mistral}

\textbf{Depth matching.} The band has to be expressed in Mistral's own layer space, not
Qwen's: the depth-matched equivalent of Qwen's layer 16 is around layer 23 of 40 here, so a
band of layers 8 to 20 would exclude the position of interest. One flag rescaled the layer
windows and not the single-layer sweep, which makes a run given {\tt --scale-layers} look
depth-matched when it is not; {\tt check\_run\_complete.py} now detects that combination.
The sweep covers layers 11 to 29 of 40, which is Qwen's 8 to 20 of 28 rescaled
(Table~\ref{tab:mistraldepth}).

\begin{table*}[tp]
\centering\small
\caption{Mistral-Small second-entity margin change by layer, depth-matched sweep. The band is layers 11--29 of 40, which is Qwen's 8--20 of 28 rescaled, so it covers layer 23, the depth-matched equivalent of Qwen's layer 16. Nothing is there: every layer from 21 to 29 is within 0.02 of zero in both prompt languages. The effect peaks at layer 11 under Marathi prompts and 18 under English, both shallower in relative depth than Qwen's 12--17 of 28. 60 gated items per prompt language, Lat--Lat donor.}
\label{tab:mistraldepth}
\begin{tabular}{ccc}
\toprule
Layer & Marathi & English\\
\midrule
11 & \textbf{0.60} & 0.33\\
12 & 0.60 & 0.44\\
13 & 0.56 & 0.59\\
14 & 0.36 & 0.69\\
15 & 0.30 & 0.81\\
16 & 0.21 & 0.94\\
17 & 0.11 & 0.94\\
18 & 0.11 & \textbf{1.01}\\
19 & 0.04 & 0.44\\
20 & 0.02 & 0.26\\
21 & 0.01 & 0.00\\
22 & $-$0.01 & $-$0.00\\
23 & 0.01 & 0.00\\
24 & $-$0.01 & $-$0.02\\
25 & 0.00 & $-$0.01\\
26 & 0.00 & $-$0.00\\
27 & $-$0.02 & $-$0.01\\
28 & 0.01 & $-$0.02\\
29 & $-$0.00 & $-$0.01\\
\bottomrule
\end{tabular}
\end{table*}

Layer 23, the depth-matched equivalent of Qwen's layer 16, carries nothing: 0.01 under
Marathi prompts and 0.00 under English, with every layer from 21 to 29 within 0.02 of zero
in both. The decay to 0.015 by layer 20 is therefore a property of the model and not of
where the band stopped. The effect peaks at layer 11 under Marathi prompts and 18 under
English,
which is 28 and 45 percent of the way through the model against Qwen's 12 to 17 of 28, or
43 to 61 percent. Mistral-Small locates this effect shallower in relative depth than Qwen
does, and in Marathi substantially so.

Mistral-Small-24B gives the largest causal gate in this study: 89 Marathi and 180 English
items against Qwen's 53 and 116. This is the causal gate, which additionally requires that an item fail
in Dev--Dev and succeed in the donor condition, and so is stricter than the behavioural
gate, under which Mistral-Small admits 165 Marathi items
(Appendix~\ref{app:models}). The patching stage
takes 60 per prompt language.

The positional finding replicates, with one qualification that the depth-matched sweep
makes visible. Over the swept band the second entity moves the margin by 0.154 in Marathi
and 0.336 in English against $-$0.004 and 0.003 at the first entity, so the ordering Qwen
shows is present. But on the sub-band matched to the layers Qwen's
Table~\ref{tab:patchci} averages, Marathi's second-entity effect has already decayed to
0.090 and sits below the question mark at 0.134 (Table~\ref{tab:mistralsites}). That is a
consequence of the peak being shallow: matching Qwen's depth window misses the Marathi
peak rather than centring on it. The ordering is a property of the band where the effect
lives, not of every band one might match it to.

\begin{table*}[tp]
\centering\small
\caption{Mistral-Small site effects over two bands. The upper block is the whole swept band, layers 11--29 of 40. The lower block is the sub-band depth-matched to the layers Qwen's Table~\ref{tab:patchci} averages, 10--18 of 28 rescaled to 14--26 of 40. The first entity is at zero in both. The second entity clears the question mark over the swept band, but not in Marathi on the depth-matched sub-band, where it has already decayed below it: Mistral locates this effect shallower than Qwen does, so matching Qwen's depth window misses the Marathi peak rather than centring on it. Magnitudes are roughly tenfold smaller than Qwen's 2.42 on either band, which the depth-matched sweep was run to establish and does. The final column is the paired entity-specificity contrast at the second entity, correct donor minus wrong-entity donor over the items present in both arms.}
\label{tab:mistralsites}
\begin{tabular}{lccccc}
\toprule
Prompt & E1 & E2 & Q & R & E2 minus wrong entity\\
\midrule
\multicolumn{6}{l}{\em Swept band, 11--29}\\
\quad Marathi & $-$0.004 & 0.154 & 0.093 & 1.471 & +0.023\\
\quad English & 0.003 & 0.336 & 0.135 & 2.301 & +0.104\\
\midrule
\multicolumn{6}{l}{\em Depth-matched, 14--26}\\
\quad Marathi & $-$0.002 & 0.090 & 0.134 & 1.491 & +0.023\\
\quad English & 0.005 & 0.389 & 0.164 & 2.429 & +0.122\\
\bottomrule
\end{tabular}
\end{table*}

The entity-specific component does not appear in Marathi on the band we swept. The
wrong-entity donor recovers 0.34 where the correct donor recovers 0.38, a paired
difference of 0.04 $[-0.01, 0.09]$ over 60 items, so within these layers the effect is
positional and we find no evidence that it needs the right person. English retains a
small entity-specific component, 0.20 $[0.11, 0.29]$. That is the opposite language
pattern to Qwen, where Marathi carries the entity specificity and English does not, and it
survives the depth-matched sweep: the paired contrast is $+0.023$ in Marathi against
$+0.104$ on the swept band and $+0.122$ on the depth-matched sub-band, over 60 items each.
So the positional result generalises across models while the entity-specific contrast
differs by prompt language, rather than replicating or failing as a whole.

Magnitudes are also much smaller than Qwen's, and that survives depth matching. On the band
matched to Qwen's, Marathi is 0.090 against Qwen's 2.42; at its own best layer it reaches
0.604. The roughly tenfold difference is therefore a property of the model rather than of
the band. We do not have an account of why the same positional site carries so much less
margin here.

\section{Corpus errors identified and removed}
\label{app:errors}

The corpus used here is a rebuild. The earlier version contained items whose two script
fields held different \emph{names} rather than different spellings of one name, and
Table~\ref{tab:corpuserr} lists examples of the failure mode. The cause was in the
construction rather than in the data: Wikidata records aliases as independent per
language lists with nothing linking an entry in one to an entry in another, and the
earlier build selected from each list separately. An item could then pair a correct
Marathi alias with a correct English alias for the same person that was not the same
name, so the Devanagari and Latin prompts asked different questions and the script
contrast was confounded with a change of name.

The rebuild pairs the two fields jointly and audits every item by hand for this
specific failure. Items were also removed when the two Devanagari fields were
identical, when a field carried a parenthetical disambiguator or several aliases joined
by a separator, and when the Devanagari was a character mapping of the Latin rather
than a Marathi rendering. All results in this paper are computed on the rebuilt corpus;
the table below is retained to document what was corrected.

\begin{table*}[tp]
\centering\small
\caption{Items whose two script fields carry different name forms. Devanagari readings
are given in transliteration. In every case the Latin field holds the more familiar form,
which biases the measured gap in the reported direction.}
\label{tab:corpuserr}
\begin{tabular}{llll}
\toprule
Fact & Devanagari field reads & Latin field reads & Kind\\
\midrule
\texttt{q485697}   & Kepler Laveran Lima Ferreira & Pepe & different name\\
\texttt{q855252}   & Rama IX, Thailand & Bhumibol Adulyadej & different name\\
\texttt{q213812}   & George Herman Ruth & Babe Ruth & different name\\
\texttt{q35811}    & Zarathustra & Zoroaster & different tradition\\
\texttt{q61047}    & Ferdinand Eisenstein & Gotthold Eisenstein & different given name\\
\midrule
\texttt{q11576}    & Ra\'ul Gonz\'alez & Ra\'ul & full vs short\\
\texttt{q310817}   & Dani Alves & Danny & full vs short\\
\texttt{q36153}    & Beyonc\'e Knowles & Beyonc\'e & full vs short\\
\texttt{q41532}    & Karl Gustav Jung & Carl Jung & full vs short\\
\texttt{q13117994} & Diwakar & Natyachhatakar Diwakar & short vs full\\
\bottomrule
\end{tabular}
\end{table*}

In all ten the Latin field carried the more familiar form, which would bias the measured
gap in the reported direction. All ten are absent from the corpus used here; they are
listed to document the failure mode, not as a caveat on the reported results. A further
seven items flagged by the same procedure are genuine, their low similarity arising
from transliteration artefacts alone; those seven are retained. Results in this paper
are computed on the 601 items that survive the audit.

\section{Surface-similarity baseline}
\label{app:surface}

The hard negatives match on occupation and country, which makes a negative pair
plausible, but they are not matched on how alike the two names look. This appendix
measures what that leaves available to a classifier that reads nothing but the strings.

A logistic classifier is fitted on five features of the two names as the model sees them
in Dev--Dev, that is on the Devanagari forms: sequence ratio, token and character
Jaccard, relative length difference, and whether the final tokens agree. It is fitted on
the train split and scored on the held-out test split, the same items the probe is
scored on (Table~\ref{tab:surface}).

\begin{table*}[tp]
\centering\small
\caption{A classifier reading only the two Devanagari strings, against the probe on the
same held-out items. The probe layer is the one validation selected. Recovery is the
share of the model's own Dev--Dev failures on which the method returns the correct
label.}
\label{tab:surface}
\begin{tabular}{lcc}
\toprule
 & Marathi & Hindi\\
\midrule
Surface baseline, balanced accuracy & 0.933 & 0.946\\
Surface baseline, AUC & 0.970 & 0.967\\
Probe, balanced accuracy & 0.925 & 0.942\\
\midrule
Surface recovery on the model's failures & 0.869 & 0.896\\
Probe recovery on the model's failures & 0.789 & ---\\
\midrule
Probe where the surface baseline is correct & 0.973 & 0.996\\
Probe where the surface baseline is wrong & 0.250 & 0.000\\
\bottomrule
\end{tabular}
\end{table*}

Three things follow. The surface baseline reaches 0.933 and 0.946 balanced accuracy,
above the probe's 0.925 and 0.942 on the same items. It recovers 86.9 and 89.6 percent
of the model's own Dev--Dev failures, above the probe's 78.9 percent in Marathi. And the
probe's successes coincide with the baseline's: where the strings suffice the probe is
correct on 97.3 and 99.6 percent of items, and where they do not it is correct on 25.0
and 0.0 percent.

\textbf{What was refitted.} Both sides were refitted under the modified map rather than
carried over frozen: the string classifier is trained on that map's training negatives, the
probe is retrained on the same run's states, and the layer is reselected on that run's
validation split. This is therefore a comparison of what each representation can predict
after adaptation, not a robustness check on frozen choices. The evaluation population is
the 120 held-out entities and their 240 pairs under the full modified map. Of those 120,
25 meet the 0.05 matching tolerance that defines the tightly matched subset of
Table~\ref{tab:matched}, with a mean absolute similarity difference of 0.201, so this
result should not be read as holding on tightly matched pairs alone.

\textbf{The coincidence is a property of the negative map.} The figures above are measured
against the shipped distractors, where a positive pair usually shares most of its
characters and a negative pair does not, so the strings alone carry most of the label.
Table~\ref{tab:matched} describes a map that removes most of that signal by matching
distractors on string similarity as well as on occupation and country. Scoring the probe
under that map, paired on the same held-out entities, is what separates the two
(Table~\ref{tab:probematched}). Under the shipped map the probe does not beat the baseline
in either language, $-0.008$ [$-0.033$, $+0.017$] in Marathi and $+0.017$ [$-0.004$,
$+0.042$] in English, which is the result reported above. Under the matched map it does,
$+0.112$ [$+0.067$, $+0.163$] and $+0.083$ [$+0.025$, $+0.142$], both intervals excluding
zero. The coincidence of successes goes the same way: where the baseline fails the probe is
correct on 0.653 [0.512, 0.786] of Marathi pairs and 0.755 [0.630, 0.872] of English ones,
against 0.250 and 0.375 under the shipped map, and neither pair of intervals overlaps. What
the probe reads is therefore not only what the strings carry, and the agreement reported
above is a property of the distractors rather than of the probe. Two limits remain. The
matched map is the partial one of Table~\ref{tab:matched}, since matching cannot succeed on
the most similar pairs, and a probe that beats a string baseline still decodes a label
rather than demonstrating that the entity association is retrieved and unused.

\begin{table*}[tp]
\centering\small
\caption{The probe against the surface baseline under the two negative maps, paired on the
same 120 held-out entities and 240 pairs, with item-clustered bootstrap intervals from 2000
draws. The shipped map is the one Table~\ref{tab:surface} reports; the matched map is the
similarity-matched one of Table~\ref{tab:matched}. Under the shipped map the probe does not
beat the baseline in either language. Under the matched map it does, and it stays correct
on about two thirds of the pairs the strings get wrong, where under the shipped map it was
correct on a quarter. The two {\em where the baseline fails} intervals do not overlap in
either language.}
\label{tab:probematched}
\begin{tabular}{llccc}
\toprule
& Negatives & Probe & Baseline & Probe $-$ baseline\\
\midrule
\multirow{2}{*}{Marathi}
 & shipped & 0.925 & 0.933 & $-$0.008 [$-$0.033, $+$0.017]\\
 & matched & 0.908 & 0.796 & \textbf{$+$0.112} [$+$0.067, $+$0.163]\\
\midrule
\multirow{2}{*}{English}
 & shipped & 0.950 & 0.933 & $+$0.017 [$-$0.004, $+$0.042]\\
 & matched & 0.879 & 0.796 & \textbf{$+$0.083} [$+$0.025, $+$0.142]\\
\midrule
\multicolumn{5}{l}{\em Probe where the surface baseline fails, $n{=}16$ shipped and 49 matched}\\
Marathi & shipped & \multicolumn{3}{l}{0.250 [0.059, 0.476]}\\
        & matched & \multicolumn{3}{l}{\textbf{0.653} [0.512, 0.786]}\\
English & shipped & \multicolumn{3}{l}{0.375 [0.125, 0.625]}\\
        & matched & \multicolumn{3}{l}{\textbf{0.755} [0.630, 0.872]}\\
\bottomrule
\end{tabular}
\end{table*}

\textbf{The probe at the causal site.} The figures above are taken at the question mark and
the readout, neither of which is where \S\ref{sec:patching} locates the causal sensitivity,
so the two lines of evidence have not been set against each other. Fitting the same probe
at the second entity closes that gap (Table~\ref{tab:probee2}). At layer 16, the layer the
causal comparison uses, it reaches 0.942 under Marathi prompts and 0.954 under English, the
best of the three positions in both. The depth agrees as well: the second entity is the only
position whose accuracy peaks inside the 12--17 window, at layers 17 and 14, while the
question mark and the readout peak at 6 and 7 under Marathi prompts. Decoding accuracy and
causal sensitivity therefore agree on position and on depth. This is a convergence, not a
second result: the probe still decodes a label rather than demonstrating retrieval, and a
position where the label is most decodable is a plausible place for a patch to act whether
or not an entity association is being read there.

\begin{table*}[tp]
\centering\small
\caption{Held-out probe balanced accuracy by position, on the 120 test entities. The middle
columns are at layer 16, the layer the causal comparison of \S\ref{sec:patching} uses; the
right-hand columns give the maximum of each position's test-layer profile. Those are
descriptive: unlike the layers of Table~\ref{tab:probeheld}, which validation selected,
they are peaks located on the evaluation split, so their position is not independent
evidence for localisation. The second entity is the best
position at layer 16 in both languages, and it is the only one whose optimum falls inside
the 12--17 causal window: the question mark and the readout peak at layers 6 and 7 under
Marathi prompts, well before it.}
\label{tab:probee2}
\begin{tabular}{lcc|cc}
\toprule
& \multicolumn{2}{c|}{At layer 16} & \multicolumn{2}{c}{Best layer}\\
\cmidrule(lr){2-3}\cmidrule(lr){4-5}
Position & Marathi & English & Marathi & English\\
\midrule
Second entity & \textbf{0.942} & \textbf{0.954} & 0.954 (17) & 0.958 (14)\\
Question mark & 0.925 & 0.933 & 0.942 (6) & 0.950 (18)\\
Readout       & 0.838 & 0.879 & 0.925 (7) & 0.950 (20)\\
\bottomrule
\end{tabular}
\end{table*}

The corpus explains why. Most positive pairs are name expansions, in which the two forms
share tokens; genuine aliases, where the forms have little in common, are a minority.
Of 120 held-out Marathi positives, 18 have a sequence ratio of 0.5 or below. On those the
surface baseline is correct on 22 percent and the probe on 39 percent, so the probe is
better where spelling does not decide the answer, but both are weak and the sample is
small. Table~\ref{tab:surfaceband} breaks the held-out positives into similarity
bands.

\begin{table*}[tp]
\centering\small
\caption{Held-out Marathi positives by how alike the two Devanagari strings are.}
\label{tab:surfaceband}
\begin{tabular}{lccc}
\toprule
Sequence ratio & Items & Surface & Probe\\
\midrule
0.3 and below & 3 & 0.000 & 0.333\\
0.3 to 0.5 & 15 & 0.267 & 0.400\\
0.5 to 0.7 & 34 & 1.000 & 0.912\\
above 0.7 & 68 & 1.000 & 0.971\\
\bottomrule
\end{tabular}
\end{table*}

\textbf{Surface separability is comparable in the two scripts.} If Devanagari names were
less alike than their Latin counterparts, the behavioural result would follow without
any internal mechanism. On this measure they are not. The gap between the mean sequence
ratio of positives and of negatives is 0.423 in Devanagari and 0.411 in Latin for the
Marathi corpus, and 0.434 against 0.411 for the Hindi corpus, a difference of 0.012 and
0.023. One summary statistic of one similarity measure is weak evidence, so we claim
only that the script difference in how separable the names are is small, not that no
surface property explains the behavioural gap.

Three claims should be kept apart here. Surface information is present in the input:
the two Devanagari strings of a same-person pair are more alike than those of a
hard negative, by 0.423 in mean sequence ratio. A classifier can exploit it, reaching
0.933 and 0.946 balanced accuracy. Neither of those is a statement about the model. What
the behavioural result adds is the third claim, that the model's own computation does not
reliably convert the available surface evidence into a correct SAME decision, since it
answers 8.5\% in Dev--Dev on the same pairs a string classifier separates at 0.933. That
third claim is about the model; the first two are about the strings.

\textbf{What this does and does not license.} The probe result stands as stated: correct
labels are linearly decodable from mid-network states on items the model answers wrongly.
It does not establish that entity knowledge in particular is present and unused, because
on this corpus a classifier with no access to the model decodes those labels at least as
well. We therefore built the negatives the separation calls for, matched on string
similarity as well as on occupation and country, and reran the behavioural measurement on
them.

\begin{table*}[tp]
\centering\small
\caption{Similarity-matched negatives, Qwen, on the 144 items where the matching succeeded to within 0.05 (positive similarity 0.409, negative 0.406). The classifier is the one of Table~\ref{tab:surface}, reading only the Devanagari strings; on this subset it falls from 0.800 to 0.680, losing two fifths of its margin over chance, and on the full corpus from 0.933 to 0.796. The model does not move with it: positive-pair accuracy, the script gap and the threshold-free separation all hold to within 0.01, and Dev--Dev negatives stay at 100.0 under both maps. Whatever the model is using, it is not what the classifier lost.}
\label{tab:matched}
\begin{tabular}{lccccc}
\toprule
 & String & \multicolumn{2}{c}{Dev--Dev} & \multicolumn{2}{c}{Lat--Lat}\\
\cmidrule(lr){3-4}\cmidrule(lr){5-6}
Negatives & classifier & positives & AUC & positives & AUC\\
\midrule
Occupation and country & 0.800 & 7.9 & 0.755 & 36.0 & 0.893\\
\textbf{Also string similarity} & \textbf{0.680} & 8.0 & 0.753 & 36.0 & 0.901\\
\bottomrule
\end{tabular}
\end{table*}

Matching cannot succeed everywhere, and the reason is structural rather than a defect of
the procedure. A distractor must be some other item's second name, and a positive pair's
similarity usually comes from a real relationship: an expansion that shares most of its
characters, which no unrelated name reproduces. Across all 601 items the best achievable
separation is 0.184 against the shipped map's 0.423. On 144 of them it reaches 0.003,
which is the subset Table~\ref{tab:matched} reports; the items that resist matching are
the similar ones, mean positive similarity 0.749 against 0.409 for those that match.
Matching on all five features at once is worse still, reaching 30 items, so we report the
partial result rather than claim a matched corpus.

What the comparison shows is a dissociation. Removing most of the surface signal costs the
string classifier two fifths of its margin over chance, and costs the model nothing: the
same positive-pair accuracies to within 0.1, the same threshold-free separation to within
0.01, the same 28-point script gap, and Dev--Dev negatives still rejected at 100.0 when
the distractor is as similar to the first name as the true answer is. The model is not
reading the surface the classifier reads. Two limits on how far that goes. The positive
prompts are byte-identical between the two runs, so unchanged positive-pair accuracy is
expected rather than evidence, and the weight of the comparison rests on what could move:
the classifier's own fall from 0.933 to 0.796 and the threshold-free separation holding to
within 0.01. And matching succeeds on 144 of 601 items, so the finding is a property of
that subset. Within those limits it weakens surface similarity as an explanation of the
script gap without removing it, and it does not license the stronger claim that entity
knowledge is present and unused, which the probe still cannot establish.

One incidental figure belongs here, because it bounds every behavioural number in this
paper. The positive prompts are byte-identical between the two runs, since the negative
map cannot affect them, yet 126 of 28{,}848 positive prompts change their answer. All of
them sit within 1.5 margin points of zero, median 0.25. The cause is batch composition:
the negative prompts differ, so padding and batch shapes differ, and bfloat16 matrix
products are not bit-reproducible across them. Roughly 0.4 percent of prompts, all of them
at the decision boundary, are therefore not reproducible across runs.

\section{Nonce-name control}
\label{app:nonce}

The corpus pairs real people, so a Dev--Dev deficit could in principle be a property of
Devanagari strings rather than of entity matching. We built 600 invented people by
recombining attested Marathi name parts, so that the strings are name-shaped and
script-appropriate but no entity stands behind them, and prepared them through the same
pipeline with the same split proportions. No fact identifier is shared with the real
corpus. Hard negatives are drawn at random rather than matched on name similarity, which
reproduces the real corpus's positive-negative separation of 0.443 against 0.423; matching
them would have made the string task harder and confounded the comparison.
Table~\ref{tab:nonce} reports the result.

\begin{table*}[tp]
\centering\small
\caption{Nonce-name control, Qwen. 600 invented people built from recombined attested name parts, sharing no fact identifier with the real corpus. Read the negatives column first: on nonce names the model answers {\em different} to essentially everything, so positive-pair accuracy falls in every condition, not only Dev--Dev. The script gap shrinks as a consequence of that shift rather than independently of it, and Marathi is uninformative at 0.036 even in the Lat--Lat condition.}
\label{tab:nonce}
\begin{tabular}{llcc|cc}
\toprule
 &  & \multicolumn{2}{c|}{Real corpus} & \multicolumn{2}{c}{Nonce corpus}\\
\cmidrule(lr){3-4}\cmidrule(lr){5-6}
Prompt & Condition & positives & negatives & positives & negatives\\
\midrule
Marathi & Dev--Dev & 0.085 & 0.998 & 0.046 & 1.000\\
 & Dev--Lat & 0.294 & 0.991 & 0.024 & 1.000\\
 & Lat--Dev & 0.350 & 0.998 & 0.169 & 1.000\\
 & Lat--Lat & 0.351 & 0.997 & 0.036 & 1.000\\
\midrule
English & Dev--Dev & 0.282 & 1.000 & 0.206 & 1.000\\
 & Dev--Lat & 0.613 & 0.989 & 0.188 & 1.000\\
 & Lat--Dev & 0.660 & 0.987 & 0.497 & 1.000\\
 & Lat--Lat & 0.563 & 0.996 & 0.259 & 1.000\\
\bottomrule
\end{tabular}
\end{table*}

The headline comparison is that the script gap shrinks: in English the Dev--Dev deficit
against Lat--Lat falls from 28 points on real names to 5 on nonce names. Read alone, that
would suggest the effect requires an entity. We do not draw that conclusion here, because
the negatives column shows the model's response bias has moved. On nonce names it answers
{\em different} to essentially every pair, at 0.999 or above in all four conditions and
both prompt languages, and positive-pair accuracy falls in every condition rather than in Dev--Dev
alone. English Lat--Lat drops from 0.563 to 0.259 with no script manipulation involved.
The gap shrinks partly because there is little accuracy left anywhere for it to appear in.

The Marathi arm is uninformative for the same reason in a stronger form. Lat--Lat reaches
0.036, so the model cannot do the nonce task in Marathi at all, and a gap measured between
0.046 and 0.036 is not a measurement. We report the English half as suggestive and the
Marathi half as a floor effect. Separating entity-dependence from the bias shift needs a
control that holds the response distribution fixed, which we have not run.

The causal arm does run. The gate is not empty: 117 English and 41 Marathi nonce items
qualify, of which 60 and 32 were patched under the same protocol as
\S\ref{sec:patching}. The positional signature survives. Mean margin movement by site,
single-layer factorial, is 0.01 at E1, 0.42 at the question mark, 0.74 at E2 and 1.41 at
the readout under Marathi prompts, against 0.03, 0.66, 2.15 and 3.56 on real names: the
same ordering, with E1 at zero and E2 well clear of the question mark, at roughly a third
of the magnitude. On that evidence the site is not exclusively about entity resolution.
It shows its signature on name-shaped strings with nobody behind them.

The flip rates need one correction before they can be read. Patching transplants a
Lat--Lat donor state, which can only supply a correct answer when the donor run is itself
correct, and on nonce names it often is not. Only 0.25 of Marathi donors and 0.82 of
English donors answer correctly, against 0.96 and 1.00 on the real corpus. The raw
window-patch flip rates, 0.03 in Marathi and 0.38 in English against 0.43 and 0.64 on real
names, therefore overstate the contrast. As a share of what the donors make achievable
they are 0.12 and 0.47, against 0.45 and 0.64. The English arm shows the effect present
and weakened; the Marathi arm is confounded by donor quality in the same way, and for the
same reason, as its behavioural counterpart.

The classifier above measures what the strings make available. A second question is
whether the behavioural effect itself depends on the strings being alike. Splitting
positive pairs by the sequence ratio between their two Latin forms answers it directly.

\begin{table*}[tp]
\centering\small
\caption{The script gap on positive pairs, split by how alike the two Latin forms are (difflib sequence ratio on the corpus of the Marathi run). If the model were matching on surface similarity and Devanagari merely made that harder, the gap should shrink as the names become less alike. It does the opposite: the gap is largest in the least similar band and smallest in the most similar one, in both prompt languages.}
\label{tab:overlap}
\begin{tabular}{lccc|ccc}
\toprule
 & \multicolumn{3}{c|}{Marathi prompts} & \multicolumn{3}{c}{English prompts}\\
\cmidrule(lr){2-4}\cmidrule(lr){5-7}
Latin overlap & Dev--Dev & Lat--Lat & gap & Dev--Dev & Lat--Lat & gap\\
\midrule
$<$0.4 & 7.2 & 38.5 & \textbf{31.3} & 14.9 & 49.4 & \textbf{34.5}\\
0.4--0.6 & 6.9 & 36.5 & \textbf{29.6} & 26.8 & 61.0 & \textbf{34.2}\\
0.6--0.8 & 8.0 & 35.6 & \textbf{27.6} & 28.8 & 55.8 & \textbf{27.1}\\
$>$0.8 & 10.8 & 32.1 & \textbf{21.3} & 32.9 & 56.9 & \textbf{24.0}\\
\midrule
items & \multicolumn{3}{c|}{58 / 94 / 295 / 154} & \multicolumn{3}{c}{same items}\\
\bottomrule
\end{tabular}
\end{table*}

The gap survives in every band, and it is not merely present at low overlap but largest
there: 31.3 points in Marathi and 34.5 in English at a ratio of 0.4 or below, against 21.3 and
24.0 above 0.8. The reading on which the model matches names by surface similarity, and
Devanagari makes that similarity harder to use, predicts the reverse, because pairs that
share little surface have little for the script to obscure. We do not take this as
settling the question. It concerns positive pairs only, and the negatives remain unmatched
on similarity, which is what a classifier reading the strings exploits. It does show that
the behavioural effect is not an artefact of the high-overlap expansion pairs that
dominate the corpus.

One negative result is worth recording, because the obvious split does not work. The
corpus carries a relation tier that separates 404 observed aliases from 197 explicit
names paired with an alias, which looks like a similarity split and is not: mean Latin
overlap is 0.715 for the alias group against 0.612 for the other. The alias tier holds
both Subhas Chandra Bose against Netaji at 0.08 and near-duplicate spelling variants.
Stratifying on the measured ratio, as above, is the only version of this split that tests
what it appears to test.

\section{Condition ranking against condition accuracy}
\label{app:latdev}

\S\ref{sec:behaviour} reports that Lat--Dev exceeds Lat--Lat in accuracy under English and Hindi prompts. This appendix gives the ranking side of the same comparison, which is what makes the pair informative: a condition that is more accurate without being better ordered differs in where its decision boundary sits, not in what it can separate.

\begin{table*}[tp]
\centering\small
\caption{Lat--Dev minus Lat--Lat AUROC of the yes-minus-no margin, per question template, held-out entities, item-clustered bootstrap over 2000 draws. Reported per template rather than pooled because pooling averages each entity's margin across templates with different distributions and can hide within-template ranking changes, as \S\ref{sec:repair} finds for the correction vector. A negative value means Lat--Dev ranks positives against negatives less well than Lat--Lat does, while Table~\ref{tab:behaviour} shows it is the more accurate condition under English and Hindi prompts.}
\label{tab:latdev}
\begin{tabular}{lccc}
\toprule
Prompt & template 0 & template 1 & template 2\\
\midrule
Marathi & $-0.008$ $[-0.041, +0.025]$ & \textbf{$-0.036$ $[-0.063, -0.015]$} & \textbf{$-0.048$ $[-0.080, -0.016]$}\\
English & $-0.009$ $[-0.027, +0.004]$ & $-0.005$ $[-0.024, +0.014]$ & $-0.003$ $[-0.017, +0.013]$\\
Hindi & $-0.007$ $[-0.030, +0.014]$ & $-0.016$ $[-0.044, +0.011]$ & \textbf{$-0.131$ $[-0.182, -0.079]$}\\
\bottomrule
\end{tabular}
\end{table*}

Table~\ref{tab:latdev} gives the per-template intervals. Under English prompts no template shows a detectable ranking difference while accuracy is 9.7 points higher. Under Hindi prompts template 2 shows a ranking deficit of 0.131 alongside 11.1 points more accuracy. Under Marathi prompts the ranking deficit is detectable on two templates and there is no accuracy gain at all, so Lat--Dev is simply the weaker condition there. We report the pattern as consistent with criterion placement varying by condition and do not treat it as evidence for that account, since it is an observation across conditions rather than an intervention on one.

On the Hindi corpus the same ordering holds, with Lat--Dev 10.5 points above Lat--Lat
under English prompts. Behaviour is therefore not a monotonic function of how much Latin
an item contains, nor a simple second-name deficit, since the condition carrying
Devanagari in the second slot is the best of the four in three of these cells. One
consequence matters for \S\ref{sec:repair}: the Lat--Lat condition the correction vector
is built to reach is not the model's own best condition, so it is a reference point and
not a ceiling.

\section{The condition control}
\label{app:control}

\S\ref{sec:repair} applies the frozen vector, window and scale to Lat--Lat inputs instead
of Dev--Dev. Table~\ref{tab:control} gives both conditions before and after, with the
unmodified rows recomputed from the behavioural run on the same 120 held-out entities.

\begin{table*}[tp]
\centering\small
\caption{The same intervention applied to the failing condition and to the healthy one, on the 120 held-out entities of each arm, at the scale that arm's validation selected. Positives are SAME pairs. The vector raises Lat--Lat acceptance in every arm, by 0.222 to 0.289, which is why we do not describe it as closing the script gap: a recovery stated as a fraction of the gap to Lat--Lat presumes that reference is fixed, and it is not. The two gains are not equal, though. Marathi is the only arm where the healthy condition gains more than the failing one; in the other three the Dev--Dev gain is larger, so we claim that the intervention is not specific to the failing condition and not that the two conditions gain equally.}
\label{tab:control}
\begin{tabular}{llcc}
\toprule
Arm & Condition & Positives & Negatives\\
\midrule
\multirow{4}{*}{Marathi}
 & Dev--Dev, unmodified & 0.065 & 0.994\\
 & Dev--Dev, $+$ vector & \textbf{0.335} & 0.985\\
 & Lat--Lat, unmodified & 0.340 & 0.997\\
 & Lat--Lat, $+$ vector & \textbf{0.629} & 0.976\\
\midrule
\multirow{4}{*}{English, mr corpus}
 & Dev--Dev, unmodified & 0.283 & 1.000\\
 & Dev--Dev, $+$ vector & \textbf{0.590} & 0.976\\
 & Lat--Lat, unmodified & 0.529 & 1.000\\
 & Lat--Lat, $+$ vector & \textbf{0.756} & 0.986\\
\midrule
\multirow{4}{*}{Hindi}
 & Dev--Dev, unmodified & 0.122 & 1.000\\
 & Dev--Dev, $+$ vector & \textbf{0.546} & 0.869\\
 & Lat--Lat, unmodified & 0.401 & 1.000\\
 & Lat--Lat, $+$ vector & \textbf{0.690} & 0.885\\
\midrule
\multirow{4}{*}{English, hi corpus}
 & Dev--Dev, unmodified & 0.260 & 1.000\\
 & Dev--Dev, $+$ vector & \textbf{0.585} & 0.974\\
 & Lat--Lat, unmodified & 0.521 & 1.000\\
 & Lat--Lat, $+$ vector & \textbf{0.743} & 0.993\\
\bottomrule
\end{tabular}
\end{table*}

The scale is frozen at the value selected for Dev--Dev and no search is run on the Lat--Lat
arm, so this run measures the held-out split again but selects nothing on it.

Table~\ref{tab:site} reports the companion control on position. The vector is always the
one estimated at the second entity; only the token it is added at changes, and the scale is
again frozen with nothing selected.
\begin{table*}[tp]
\centering\small
\caption{The same vector, window and scale added at four token positions, Dev--Dev inputs, 120 held-out entities per arm. The second entity is the position the vector was estimated at and the one \S\ref{sec:patching} localises; the other three are controls. One result holds in every arm: the first entity moves accuracy by at most 0.054, so the gain is not available anywhere in the prompt. The readout is weaker and less consistent: it lowers accuracy in three arms but matches the second entity under Hindi prompts, 0.550 against 0.546 at comparable negatives, which we report without an account for it. We therefore read the intervention as depending on position rather than as acting only at the second entity.}
\label{tab:site}
\begin{tabular}{lcccc}
\toprule
& \multicolumn{2}{c}{Native prompts} & \multicolumn{2}{c}{English prompts}\\
\cmidrule(lr){2-3}\cmidrule(lr){4-5}
Site & positives & negatives & positives & negatives\\
\midrule
\multicolumn{5}{l}{\em Marathi corpus}\\
unmodified & 0.065 & 0.994 & 0.283 & 1.000\\
\textbf{second entity} & \textbf{0.335} & 0.985 & \textbf{0.590} & 0.976\\
first entity & 0.071 & 0.993 & 0.332 & 1.000\\
question mark & 0.250 & 0.953 & 0.433 & 0.990\\
readout & 0.036 & 0.997 & 0.160 & 1.000\\
\midrule
\multicolumn{5}{l}{\em Hindi corpus}\\
unmodified & 0.122 & 1.000 & 0.260 & 1.000\\
\textbf{second entity} & \textbf{0.546} & 0.869 & \textbf{0.585} & 0.974\\
first entity & 0.176 & 0.994 & 0.307 & 1.000\\
question mark & 0.481 & 0.871 & 0.392 & 0.997\\
readout & 0.550 & 0.861 & 0.178 & 1.000\\
\bottomrule
\end{tabular}
\end{table*}

\begin{table*}[tp]
\centering\small
\caption{The script disparity in positive-pair accuracy, in points, before any intervention and with the same frozen vector applied to both conditions. The untreated columns carry paired item-clustered intervals over 2000 draws, since Dev--Dev and Lat--Lat are two conditions of one run; the treated column is a point estimate, because treated Dev--Dev and treated Lat--Lat come from separate evaluations and the entities cannot be paired across them from the files we retain. Treating both conditions equally, the disparity narrows in three of the four arms and widens slightly in Marathi. A percentage closure does not make this comparison: that figure sets a treated Dev--Dev against an untreated Lat--Lat, which measures movement rather than parity. The intervention improves both conditions and does not eliminate the disparity between them.}
\label{tab:gap}
\begin{tabular}{lcccc}
\toprule
& \multicolumn{2}{c}{Positive-pair gap} & & Balanced gap\\
\cmidrule(lr){2-3}
Arm & untreated & both treated & Change & untreated\\
\midrule
Marathi & 27.5 [+21.1, +34.4] & 29.4 & \textbf{+1.9} & +13.9 [+10.6, +17.4]\\
English, mr corpus & 24.6 [+16.2, +33.8] & 16.6 & -8.0 & +12.3 [+8.1, +16.9]\\
Hindi & 27.9 [+21.8, +34.7] & 14.4 & -13.5 & +14.0 [+10.9, +17.4]\\
English, hi corpus & 26.1 [+18.3, +34.4] & 15.8 & -10.3 & +13.1 [+9.2, +17.2]\\
\bottomrule
\end{tabular}
\end{table*}

\begin{table*}[tp]
\centering\small
\caption{The window intervention template by template, Marathi prompts, Dev--Dev, 120 held-out entities. The AUROC columns average an entity's two answer keys within a template and then rank, with paired item-clustered intervals over 2000 draws; the operating-point columns are prompt-level. The threshold column is the oracle diagnostic of Table~\ref{tab:calibration}, placed per template on the test curve; the offset column is chosen on the validation tuning entities under that arm's own validation negative-pair accuracy and reads nothing from test. Ranking improves on templates 0 and 2, which a constant offset cannot do. The vector beats both calibration columns on template 0 alone, the one its scale was tuned on. Template 2 carries an apparent tension: its ranking gain excludes zero while its true-positive rate at the operating point falls, which is consistent because the ranking gain lies away from the low-false-positive region the operating point sits in.}
\label{tab:pertemplate}
\begin{tabular}{lccc|ccc}
\toprule
& \multicolumn{3}{c|}{Entity AUROC} & \multicolumn{3}{c}{TPR at the vector's FPR}\\
\cmidrule(lr){2-4}\cmidrule(lr){5-7}
Template & baseline & $+$ vector & change & vector & threshold & offset\\
\midrule
0 & 0.851 & 0.885 & \textbf{+0.034} $[+0.006, +0.062]$ & \textbf{0.425} (0.021) & 0.308 & 0.267 (0.004)\\
1 & 0.938 & 0.941 & +0.003 $[-0.009, +0.015]$ & 0.392 (0.012) & 0.371 & 0.637 (0.037)\\
2 & 0.924 & 0.944 & \textbf{+0.020} $[+0.008, +0.035]$ & 0.188 (0.012) & 0.242 & 0.421 (0.021)\\
\bottomrule
\end{tabular}
\end{table*}

\begin{table*}[tp]
\centering\small
\caption{How much each item moves, holding the direction fixed. Mean signed margin change on gated items, by fold and prompt language, at layer 16 and rank 1 on validation. The upper block varies the coefficient for the mean-difference direction: across the six Marathi and Hindi cells a single constant for every item spans 3.55 to 7.00 and an item's own coefficient spans 2.52 to 6.62, so a constant amount is not what limits recovery. The lower block gives the paired difference between an item's own coefficient and another item's for the sensitivity direction: it spans zero in 9 of 9 cells, the largest magnitude being $-1.14$ $[-4.86, +1.73]$ in mr fold t12e0. With 10 to 23 entities per fold these intervals bound an advantage loosely rather than excluding one.}
\label{tab:amountarms}
\begin{tabular}{llccc}
\toprule
\multicolumn{5}{l}{\em Mean-difference direction, gated margin change}\\
Prompts & Fold & one constant for every item & the item's own & another item's\\
\midrule
en & t01e2 & 5.51 & 3.97 & 4.18\\
en & t02e1 & 3.30 & 3.18 & 3.26\\
en & t12e0 & 4.13 & 3.85 & 4.05\\
hi & t01e2 & 4.35 & 4.47 & 4.25\\
hi & t02e1 & 7.00 & 6.62 & 6.20\\
hi & t12e0 & 5.84 & 5.04 & 5.20\\
mr & t01e2 & 3.55 & 2.52 & 2.62\\
mr & t02e1 & 4.29 & 4.06 & 4.14\\
mr & t12e0 & 4.36 & 4.29 & 4.00\\
\midrule
\multicolumn{5}{l}{\em Sensitivity direction, own coefficient minus another item's}\\
Prompts & Fold & difference & interval & entities\\
\midrule
en & t01e2 & $+0.25$ & $[-0.34, +0.87]$ & 23\\
en & t02e1 & $+0.11$ & $[-0.40, +0.66]$ & 23\\
en & t12e0 & $+0.01$ & $[-0.47, +0.48]$ & 23\\
hi & t01e2 & $+0.00$ & $[-0.85, +0.88]$ & 10\\
hi & t02e1 & $+0.05$ & $[-1.55, +1.70]$ & 10\\
hi & t12e0 & $+0.01$ & $[-1.44, +1.77]$ & 10\\
mr & t01e2 & $-0.03$ & $[-0.62, +0.62]$ & 10\\
mr & t02e1 & $-0.35$ & $[-1.60, +0.70]$ & 10\\
mr & t12e0 & $-1.14$ & $[-4.86, +1.73]$ & 10\\
\bottomrule
\end{tabular}
\end{table*}

\section{Investigation history}
\label{app:history}

\S\ref{sec:repair} reports the final experiment and the final interpretation. This appendix
keeps the route to it: what a single layer and a small scale do, why that null was a
property of the search region rather than of the direction, a reference arm whose outcome
sits awkwardly beside the intervention, and the gradient argument that predicted the
single-layer behaviour. None of it is needed to read the result, and all of it is needed to
check that the result was not arrived at by searching until something worked.

Applied at a single layer the intervention recovers little. In Qwen the vector leaves negative-pair accuracy at 1.000 under English prompts and
0.993 under Marathi, and raises positive-pair accuracy by 1.8, 5.7 and 6.7 points under Marathi, Hindi and
English prompts (Table~\ref{tab:recovery}), which is 7, 20 and 27 percent of the
gap to Lat--Lat as it stood before the intervention; we give both, since the control below shows that reference
moves when the same vector is applied to it. Supplying the Latin form in the
prompt instead recovers 56 percent in Marathi and 86 percent in Hindi, and exceeds the reference in English.

Two changes were needed and they separate cleanly. Holding the scale at the single-layer
$\alpha{=}6$ and only widening from one layer to the window lifts Marathi validation
accuracy from 0.083 to 0.233; extending the grid beyond $\alpha{=}8$ lifts it again to
0.392, after which it turns over (Table~\ref{tab:winsweep}). The single-layer null is therefore a property of the
search region in both coordinates, and it is diagnosable without running anything: the
distance each item must travel along the direction is 15 to 18 scale units in Marathi and
18 to 22 in English, against a single-layer grid reaching 8
(Table~\ref{tab:donorcoord}). The
scale at which Marathi positive-pair accuracy peaks, 16, falls inside that range, and the
English peak at 24 sits just above it; selection returns 12 in Marathi because balanced
accuracy trades a little positive accuracy against the negatives.

One result does not fit comfortably. The reference arm of \S\ref{sec:mediator}, which
applies the rank-one direction across the same window with each item's own donor
coordinate, transfers about half the margin change of a whole-state patch but flips no
gated Marathi item at all (Table~\ref{tab:refarm}). That arm and the intervention are
not the same measurement: the reference arm scores only the gated items, the hardest
residue, on a held-out template, while the intervention scores every item on all three templates
(Table~\ref{tab:template}).

The gradient lens of \S\ref{sec:jacobian} explains the single-layer behaviour it was
introduced for. It gives a reason to expect little from a
fixed direction: the correction vector is close to orthogonal to the direction the output
is locally sensitive to, so to first order at the measured activation it moves the margin
very little per unit of scale. That is consistent with the small recovery we observe; it
does not establish what a large step would do, since that leaves the neighbourhood the
gradient describes, which is what the window result confirms: the selected step is 12
and the best is 16, against a grid that stopped at 8. The sweep of
Table~\ref{tab:jacsweep} confirms the same account from the other side. If the correction
vector is close to orthogonal to the sensitivity direction, cosines $-0.034$ to 0.059,
then to first order it moves the margin hardly at all per unit of scale, while the
sensitivity direction moves it at the full gradient rate by construction. The operating
scales follow that asymmetry: on the coarse grid $\alpha{=}1$ for the sensitivity
direction, 1.1 on the finer grid of Table~\ref{tab:jacfine}, against 12 for the
correction vector. We do not read the ratio as a quantity the cosines predict, since a
vanishing first-order term predicts no finite scale at all; it is rather that the
correction vector can only work by leaving the neighbourhood the gradient describes, which
is what a step of that size does.

\section{Protocol}
\label{app:protocol}

\textbf{Reading Tables~\ref{tab:calibration} and~\ref{tab:calvsprobe}.}
In Table~\ref{tab:calibration}, AUROC averages an entity's six prompt margins before
ranking and bootstrap intervals resample entities over 2000 draws. Its true- and
false-positive rates pool prompts, as in the behavioural accuracy tables.
Table~\ref{tab:calvsprobe} instead averages the six margins before classification,
so balanced accuracy, TPR and FPR there are all entity-level. A constant offset
does not change AUROC in either aggregation. The {\em change} column is
a difference of unrounded values and need not equal the difference of the rounded figures
elsewhere.

The offset of Table~\ref{tab:calibration} is one constant per arm, chosen on the 60
validation tuning entities of template 0 by balanced accuracy under a floor equal to that
arm's own validation negative-pair accuracy at the selected scale, 0.975, 0.975 and 0.925.
No part of its selection reads the test split. The vector was held to no floor at all, so
the offset is selected under the stricter rule of the two. The threshold-only column is
what the unmodified model reaches by moving its threshold alone to the intervention's own
rate, which makes it an oracle diagnostic rather than a deployable rule.

The offset of Table~\ref{tab:calvsprobe} is unconstrained, which is why its false-positive
rate runs higher, 0.050 against the probe's 0.017; the $+0.054$ advantage there is a
balanced-accuracy difference at unmatched rates and not a gain at a common one. Its probe
column uses the second-entity layer each arm's validation split selected, 12 for Marathi
and 21 for English; the larger figures in Table~\ref{tab:probee2} are maxima over a
test-layer profile and were not selected. Its Marathi probe and string rows agree to three
decimals by coincidence rather than construction: the two predictors differ on 2 of 240
pairs and the disagreements cancel. That near-identity is the point, since on the shipped
negative map the two are very nearly the same classifier.

Every intervention in this paper is fitted on one population, selected on a second and
reported on a third. The combinations differ between them, and the differences matter for
reading the numbers, so they are collected in Table~\ref{tab:protocol} rather than left distributed
across the sections that use them.

\begin{table*}[t]
\centering\small
\caption{What each intervention was fitted on, selected on and reported on. The selection
metric is balanced accuracy in every case. No intervention is selected under a floor on
negative-pair accuracy: negative accuracy enters only as a tiebreak between settings with
equal balanced accuracy, which is why the selected settings are free to lower it, and do,
on validation, to 0.975 in the Marathi window arm and 0.925 in the Hindi one. The scalar
offset of Table~\ref{tab:calibration} is the one exception: it was held to a floor, which
handicaps the offset and so favours the vector in that comparison.}
\label{tab:protocol}
\begin{tabularx}{\textwidth}{@{}lXXXX@{}}
\toprule
Intervention & Fitted on & Selected on & Templates & Reported on\\
\midrule
Single-layer vector & 361 train entities & 60 of 120 validation & 0 for selection, 0--2 for test & 120 test\\
Window vector & 361 train entities & 60 of 120 validation & 0 for selection, 0--2 for test & 120 test\\
Sensitivity sweep & 31 Marathi, 60 English gated train items & 60 of 120 validation & 0 only & validation only\\
Scalar offset & nothing fitted; one constant selected & 60 of 120 validation & 0 only & 120 test\\
Rank-one subspace & 31 Marathi, 60 English gated train items & fold t01e2 & train 0--1, evaluate 2 & gated test items\\
Condition control & frozen, nothing fitted & nothing selected & 0--2 & 120 test\\
Site control & frozen, nothing fitted & nothing selected & 0--2 & 120 test\\
\bottomrule
\end{tabularx}
\end{table*}

Two consequences follow. The held-out split has been scored four times in both corpora,
under the single-layer vector, the window vector, the condition control and the site
control, the last two reported in Appendix~\ref{app:control}, so these evaluations share test items and are not independent replications. The control selects nothing, which is why
it is reported without a tuning column; \S\ref{sec:repair} reports
them as such. And the selection and evaluation populations differ in template coverage for
the vector arms: the scale is chosen on template 0 and reported over all three, which is
what makes the per-template breakdown of \S\ref{sec:repair} necessary rather than
decorative.

\begin{table*}[t]
\centering\small
\caption{Both correction directions on the validation tuning items, selected by balanced accuracy exactly as in Table~\ref{tab:winsweep}. Scales are not comparable across the two panels: each direction is unit length, so its own scale carries all magnitude, and the grids are the ones each was searched on. Two things the coarse grid of Table~\ref{tab:jacsweep} is too sparse to show. The sensitivity direction has a plateau: 6 consecutive Marathi scales from $\alpha{=}1.0$ to 1.5 sit within 0.025 of its maximum. And under the same rule it reaches higher balanced accuracy than the mean-difference vector in both languages, 0.846 against 0.679 in Marathi and 0.892 against 0.850 in English, the Marathi gap because the mean-difference vector never lifts positive-pair accuracy above 0.392 at any scale while the sensitivity direction reaches 0.858. It is fitted on 31 gated training items against the other's 361. No held-out test evaluation was run for the sensitivity direction, so these figures cannot be set beside the test numbers of Table~\ref{tab:wintest}.}
\label{tab:jacfine}
\begin{tabular}{cccc|cccc}
\toprule
& \multicolumn{3}{c|}{Local sensitivity} & & \multicolumn{3}{c}{Mean difference}\\
\cmidrule(lr){1-4}\cmidrule(lr){5-8}
$\alpha$ & positives & negatives & balanced & $\alpha$ & positives & negatives & balanced\\
\midrule
\multicolumn{8}{l}{\em Marathi prompts}\\
0.5 & 0.258 & 1.000 & 0.629 & 0.5 & 0.067 & 1.000 & 0.533\\
0.6 & 0.308 & 1.000 & 0.654 & 1 & 0.075 & 1.000 & 0.537\\
0.7 & 0.342 & 1.000 & 0.671 & 1.5 & 0.083 & 1.000 & 0.542\\
0.8 & 0.483 & 1.000 & 0.742 & 2 & 0.083 & 1.000 & 0.542\\
0.9 & 0.600 & 0.992 & 0.796 & 3 & 0.117 & 1.000 & 0.558\\
1.0 & 0.683 & 0.983 & 0.833 & 4 & 0.167 & 1.000 & 0.583\\
1.1 & 0.733 & 0.958 & \textbf{0.846} & 6 & 0.233 & 1.000 & 0.617\\
1.2 & 0.775 & 0.917 & 0.846 & 8 & 0.250 & 1.000 & 0.625\\
1.3 & 0.783 & 0.858 & 0.821 & 12 & 0.383 & 0.975 & \textbf{0.679}\\
1.4 & 0.833 & 0.833 & 0.833 & 16 & 0.392 & 0.958 & 0.675\\
1.5 & 0.858 & 0.800 & 0.829 & 24 & 0.325 & 0.933 & 0.629\\
 & & &  & 32 & 0.275 & 0.933 & 0.604\\
\midrule
\multicolumn{8}{l}{\em English prompts}\\
0.5 & 0.533 & 1.000 & 0.767 & 0.5 & 0.333 & 1.000 & 0.667\\
0.6 & 0.608 & 1.000 & 0.804 & 1 & 0.333 & 1.000 & 0.667\\
0.7 & 0.650 & 1.000 & 0.825 & 1.5 & 0.333 & 1.000 & 0.667\\
0.8 & 0.700 & 0.983 & 0.842 & 2 & 0.350 & 1.000 & 0.675\\
0.9 & 0.742 & 0.983 & 0.863 & 3 & 0.358 & 1.000 & 0.679\\
1.0 & 0.767 & 0.983 & 0.875 & 4 & 0.417 & 1.000 & 0.708\\
1.1 & 0.783 & 0.975 & 0.879 & 6 & 0.467 & 1.000 & 0.733\\
1.2 & 0.800 & 0.942 & 0.871 & 8 & 0.542 & 1.000 & 0.771\\
1.3 & 0.833 & 0.925 & 0.879 & 12 & 0.617 & 0.975 & 0.796\\
1.4 & 0.842 & 0.925 & 0.883 & 16 & 0.675 & 0.975 & 0.825\\
1.5 & 0.858 & 0.925 & \textbf{0.892} & 24 & 0.725 & 0.975 & \textbf{0.850}\\
 & & &  & 32 & 0.708 & 0.983 & 0.846\\
\bottomrule
\end{tabular}
\end{table*}

\section{The 24B comparison}
\label{app:models}

\S\ref{sec:behaviour} reports the script gaps, 6.9
points for Sarvam-M against 25.8 for Mistral-Small-24B. Four qualifications belong with
them. Sarvam-M answers Dev--Dev same-person pairs at 75.2 percent where Qwen reaches 8.5.
Under one gate applied to all three models, Lat--Lat accuracy at or above 0.8 with Dev--Dev
at or below 0.2 on the first paraphrase averaged over the two answer keys, the two untuned
models admit 201 and 165 Marathi items while Sarvam-M admits one. Sarvam-M's own larger
deficit is not script but prompt language: holding both names in Devanagari and asking in
English rather than Marathi moves negative-pair accuracy by 31.5 points, against 8.3 for
switching the names to Latin under Marathi prompts. And the change is not a uniform
improvement: on negative pairs under Marathi prompts Mistral-Small reaches 99.2 percent and
Sarvam-M 65.6, so Sarvam-M shows both a smaller script gap and a higher acceptance rate,
though balanced accuracy at Dev--Dev still favours it, 70.4 against 56.8. Sarvam-M is not
analysed causally: it has one gated Marathi item. Mistral-Small admits the largest causal
gate here, 89 Marathi and 180 English items, and we run an exploratory sweep on it rather
than the full analysis: it reproduces the E2-over-E1 sensitivity, and the correct-donor
advantage differs by prompt language.

\section{Compute and reproducibility}
\label{app:compute}

All runs use a single NVIDIA A100 80GB in bfloat16, PyTorch 2.11.0, Transformers
4.48.3, seed 20260813, with model revisions pinned. Batch sizes are 24 for Qwen and 32
for Gemma. Per model the pipeline scores 57{,}696 behavioural prompts, caches hidden
states at four token positions across all layers, and performs the patching, head and
recovery stages; hidden-state caches are approximately 4 GB per model. Head screening
is the dominant cost. All result artefacts are CSV or JSON; the pipeline is
stage-wise resumable, and every reported figure in this paper is recomputable from the
result tables.

\textbf{One note on column naming.} The column named \texttt{flip\_rate} in the
patching tables holds the fraction of items \emph{correct after} the intervention, not
the fraction whose answer changed. The two coincide wherever the baseline is wrong for
every item, which holds for the single-layer, window, joint, cross-context and control
experiments, and we have verified that every flip rate quoted in this paper is a true
flip rate. They diverge for two cases. In the reverse-direction experiment the baseline
is a working Lat--Lat run and is correct on 99 percent of items, so the column there
reads as ``still correct''. On negative pairs the baseline is likewise correct, and we
report that column as post-intervention accuracy rather than as a flip rate. Readers
recomputing from the result tables should use \texttt{baseline\_correct} together with
\texttt{patched\_correct} rather than relying on the column name alone.

\section{Malayalam corpus and behavioural results}
\label{app:malayalam}

\textbf{Construction.} The corpus draws 380 of the 601 Wikidata entities of
Appendix~\ref{app:corpus} and supplies each name in Malayalam alongside the
original Latin form, which is carried over unchanged. Selection favoured
familiar figures across politics, literature, cinema, music, science and sport;
this is a curation heuristic and not a measured fame threshold, so the corpus is
not a representative sample of the 601. Of the 760 Malayalam strings, 171 are
exactly attested in cached Wikidata Malayalam labels, aliases or name claims, 18
are spelling or joiner variants of an attested form, and 571 are machine
transliterations of the supplied Latin name. Nineteen pairs have both Malayalam
forms exactly attested. Attestation evidences a spelling on the entity record
and not that the Latin and Malayalam forms are the same name, which is what the
audit below establishes.

\textbf{Audit.} A Malayalam speaker reviewed all 380 items and all 380
distractors against the protocol of Appendix~\ref{app:errors}, extended with
three Malayalam-specific checks: chillu and zero-width-joiner encoding, the 1971
orthography reform and word-final samvruthokaram. Six items were rejected, four
because a field carried a generic honorific that is not part of the name
({\em Dr.}, {\em Sir}, {\em Dr}, {\em Sir}) and two because a field joined a
second alias into one string, one through a trailing comma and one as
{\em William Henry ``Bill'' Gates}. Nineteen further title-like tokens were
examined and kept because they are part of the name as used, among them
{\em Swami}, {\em Ustad}, {\em Netaji}, {\em Begum}, {\em Ayatollah} and
{\em Sherpa}. Four relation labels were corrected, nine distractors were flagged
as sharing neither occupation nor country, and eighteen items were flagged for
carrying U+200C inside a consonant cluster, which the pipeline preserves under
NFC; two of those eighteen fall in the held-out split. No string was rejected as
a character-by-character mapping of the Latin rather than genuine Malayalam, and
no item was rejected for its two scripts holding different names. Ten items
whose two names are the same name in reversed given and family order were
withheld as uncertain and retained in the evaluated set.

\textbf{Distractors and splits.} Distractors are one-to-one within split,
selected by shared occupation QID, then shared country, then Latin string
similarity, excluding surface collisions and known names of the anchor. Removing
the six rejected items left six anchors without a distractor. Because the map is
one-to-one, cascading those removals would have discarded 117 entities the audit
never objected to, so the six anchors were re-pointed by the same rule; four of
their new distractors already serve another anchor, and two share neither
occupation nor country. Splits of 222/76/76 are inherited from a permutation of
all 601 supplied rows under \texttt{numpy.random.default\_rng(20260813)}
assigning 361/120/120, with the subset retaining its assignments. This is
reproducible but was not verified against the split used for the original run,
so no probe, vector or threshold from the Devanagari experiments is transferred
to this corpus.

\textbf{Protocol.} Four script configurations, two prompt languages, three
paraphrases and both answer keys give 96 prompts per entity and 35{,}904 per
run. Scoring, aggregation and intervals follow Appendix~\ref{app:protocol}. The
matched Devanagari comparator is a fresh run over the same 374 entities using
the same distractor map, not the 601-item results reported elsewhere in this
paper; the two have different denominators and different negatives and are not
interchangeable. Several cells have no false alarms at all, so $d'$ there would
rest entirely on the log-linear correction and we report AUROC instead.

\textbf{Positive control.} The English Lat--Lat arm is shared by the two
corpora by construction, and 4{,}488 of its prompts are byte-identical across
the two runs. The runs agree on 99.6\% of those trials. All seventeen
disagreements lie within 0.5 of zero margin, with a median absolute margin of
0.25, and the largest margin movement anywhere in the shared arm is 1.75 points,
with two prompts moving more than 1.5. This is the batch-composition sensitivity
of bfloat16 reductions, measured here on a pair of runs that ask identical
questions.

\textbf{Knowledge and causal gates.} Over the 374 items, the Malayalam prompt
language admits 299 entities as known in at least one configuration, 223 as
Lat--Lat correct with both-native incorrect, and 10 as correct in all four
configurations. The matched Devanagari run gives 263, 151 and 23. Under English
prompts the two corpora give 345, 142, 85 and 338, 133, 106. The larger gated
set and the smaller robust set both indicate that Malayalam is the harder script
for this model; no causal experiment was run on this corpus.

\textbf{Paraphrases.} Same-person accuracy on the held-out 76 entities, by
paraphrase and configuration, is given in Table~\ref{tab:malayalampara}. The
third Malayalam paraphrase sits near the floor in every configuration, between
2.0 and 7.9 percent, while the same question in English reaches 12.5 to 47.4
percent. The templates were native-checked, so this is the model failing a
natural Malayalam construction rather than an artefact of the wording; the first
two paraphrases carry the result.

\begin{table*}[t]
\centering\small
\caption{The full Malayalam grid and its matched Devanagari comparator, 374
entities, Qwen2.5-7B. Accuracy is a percentage over three paraphrases and both
answer keys; AUROC is entity-averaged over the six prompt margins. Gaps are
Lat--Lat minus the both-native condition, paired by entity, with 95\%
item-clustered intervals over 2{,}000 resamples; the test column repeats the gap
on the 76 held-out entities. The English Lat--Lat cells of the two corpora come
from byte-identical prompts and agree to 0.0 points.}
\label{tab:malayalamfull}
\setlength{\tabcolsep}{4pt}
\begin{tabular}{llcccccccc}
\toprule
& & \multicolumn{2}{c}{Both native} & \multicolumn{2}{c}{Native, Lat} & \multicolumn{2}{c}{Lat, native} & \multicolumn{2}{c}{Both Latin}\\
\cmidrule(lr){3-4}\cmidrule(lr){5-6}\cmidrule(lr){7-8}\cmidrule(lr){9-10}
Corpus & Prompt & pos. & neg. & pos. & neg. & pos. & neg. & pos. & neg.\\
\midrule
Malayalam & Malayalam & \textbf{5.3} & 100.0 & 33.5 & 98.6 & 38.1 & 99.6 & \textbf{48.8} & 99.9\\
          & English & \textbf{26.5} & 98.9 & 58.2 & 97.1 & \textbf{66.4} & 95.6 & 62.9 & 99.6\\
Devanagari & Marathi & \textbf{10.4} & 100.0 & 33.2 & 98.8 & 37.8 & 99.8 & \textbf{41.0} & 99.6\\
           & English & \textbf{30.0} & 99.6 & 64.4 & 98.6 & \textbf{66.0} & 97.5 & 62.9 & 99.6\\
\midrule
\multicolumn{10}{l}{\em Entity-averaged AUROC}\\
Malayalam & Malayalam & \multicolumn{2}{c}{0.925} & \multicolumn{2}{c}{0.941} & \multicolumn{2}{c}{0.964} & \multicolumn{2}{c}{0.990}\\
          & English & \multicolumn{2}{c}{0.912} & \multicolumn{2}{c}{0.951} & \multicolumn{2}{c}{0.965} & \multicolumn{2}{c}{0.987}\\
Devanagari & Marathi & \multicolumn{2}{c}{0.935} & \multicolumn{2}{c}{0.960} & \multicolumn{2}{c}{0.970} & \multicolumn{2}{c}{0.981}\\
           & English & \multicolumn{2}{c}{0.948} & \multicolumn{2}{c}{0.969} & \multicolumn{2}{c}{0.971} & \multicolumn{2}{c}{0.987}\\
\midrule
\multicolumn{10}{l}{\em Lat--Lat minus both-native, as points and AUROC, all 374 items, then the 76 test items}\\
Malayalam & Malayalam & \multicolumn{4}{l}{$+43.5\ [40.0, 47.1]$, $+0.065\ [0.050, 0.082]$} & \multicolumn{4}{l}{$+39.0\ [31.6, 46.5]$, $+0.051\ [0.019, 0.090]$}\\
          & English & \multicolumn{4}{l}{$+36.4\ [31.4, 40.9]$, $+0.075\ [0.057, 0.094]$} & \multicolumn{4}{l}{$+34.6\ [24.3, 45.4]$, $+0.060\ [0.024, 0.103]$}\\
Devanagari & Marathi & \multicolumn{4}{l}{$+30.6\ [26.6, 34.5]$, $+0.045\ [0.031, 0.061]$} & \multicolumn{4}{l}{$+28.5\ [19.5, 37.5]$, $+0.036\ [0.004, 0.074]$}\\
           & English & \multicolumn{4}{l}{$+32.9\ [28.3, 37.7]$, $+0.039\ [0.025, 0.052]$} & \multicolumn{4}{l}{$+29.6\ [18.0, 40.4]$, $+0.026\ [0.004, 0.055]$}\\
\bottomrule
\end{tabular}
\end{table*}

\begin{table*}[t]
\centering\small
\caption{Same-person accuracy (\%) on the 76 held-out Malayalam entities by
paraphrase. P1 asks whether the two names are one person, P2 whether they are
names of one person, P3 whether one is another name of the other. All three were
checked by a Malayalam speaker. P3 sits near the floor in Malayalam and
recovers in English on the same items, so its collapse is specific to the
Malayalam rendering of that construction.}
\label{tab:malayalampara}
\begin{tabular}{llcccc}
\toprule
Prompt & Paraphrase & Mlym--Mlym & Mlym--Lat & Lat--Mlym & Lat--Lat\\
\midrule
Malayalam & P1 & 10.5 & 48.7 & 62.5 & 65.8\\
          & P2 & 10.5 & 56.6 & 64.5 & 70.4\\
          & P3 & \textbf{2.6} & \textbf{7.9} & \textbf{2.0} & \textbf{4.6}\\
English & P1 & 26.3 & 73.7 & 86.2 & 61.2\\
        & P2 & 28.3 & 75.7 & 90.8 & 62.5\\
        & P3 & 12.5 & 40.8 & 23.7 & 47.4\\
\bottomrule
\end{tabular}
\end{table*}

\end{document}